\documentclass[11pt,a4paper]{article}
\usepackage{times,latexsym}
\usepackage[dvipsnames]{xcolor}
\usepackage[T1]{fontenc}
\usepackage{subcaption}
\usepackage{graphicx}
\usepackage{linguex}
\usepackage{booktabs}
\usepackage{longtable}
\usepackage{enumitem}
\usepackage{censor}
\usepackage{xurl}

\usepackage[acceptedWithA]{tacl2018v2}

\usepackage{scalerel}
\def\raven{\scalerel*{\includegraphics[width=4em]{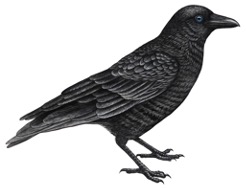}}{X}}

\usepackage{xspace,mfirstuc,tabulary}

\newif\iftaclinstructions
\taclinstructionsfalse 
\iftaclinstructions
\renewcommand{\confidential}{}
\renewcommand{\anonsubtext}{(No author info supplied here, for consistency with
TACL-submission anonymization requirements)}
\newcommand{\instr}
\fi

\newcommand{\lm}[1]{\textsc{PLM}s}

\iftaclpubformat 

\else

\fi

\title{How much do language models copy from their training data? \\Evaluating linguistic novelty in text generation using RAVEN \raven{}}

\author{
 R. Thomas McCoy,\Thanks{Work partially done while at Microsoft Research.}$^{\ \ 1}$ Paul Smolensky,$^{2,1}$ Tal Linzen,$^{3}$ Jianfeng Gao,$^{2}$ Asli Celikyilmaz\footnotemark[1]$^{\ \ 4}$ \\
 \textsuperscript{1}Johns Hopkins University \textsuperscript{2}Microsoft Research
 \textsuperscript{3}New York University \textsuperscript{4}Facebook AI Research\\
  \texttt{tom.mccoy@jhu.edu, psmo@microsoft.com, linzen@nyu.edu,} \\
  \texttt{jfgao@microsoft.com, aslic@fb.com} \\
}

\date{}

\newif\ifparens\parensfalse

\usepackage{cleveref}
\crefname{ExNo}{}{}
\crefname{SubExNo}{}{}

\creflabelformat{SubExNo}{(#2#1#3)}
\creflabelformat{ExNo}{(#2#1#3)}
\crefrangelabelformat{SubExNo}{(#3#1#4--#5\crefstripprefix{#1}{#2}#6)}
\crefrangelabelformat{ExNo}{(#3#1#4)--(#5#2#6)}

\begin{document}
\maketitle
\begin{abstract}
  Current language models can generate high-quality text.
  Are they simply copying text they have seen before, or have they learned generalizable linguistic abstractions?
  To tease apart these possibilities, we introduce RAVEN, a suite of analyses for assessing the novelty of generated text, focusing on 
  sequential structure ($n$-grams) and syntactic structure. We apply these analyses to four neural language models (an LSTM, a Transformer, Transformer-XL, and GPT-2). For local structure---e.g., individual dependencies---model-generated text is substantially less novel than our baseline of human-generated text from each model's test set. 
  For larger-scale structure---e.g., overall sentence structure---model-generated text is as novel or even more novel than the human-generated baseline, but models still sometimes copy substantially, in some cases duplicating passages over 1,000 words long from the training set. 
  We also perform extensive manual analysis showing that GPT-2's novel text is usually well-formed morphologically and syntactically but has reasonably frequent semantic issues (e.g., being self-contradictory). 
\end{abstract}

\setlength{\Extopsep}{4pt}
\setlength{\Exlabelwidth}{0.7em}
\setlength{\SubExleftmargin}{1.35em}
\renewcommand{\firstrefdash}{} 

\section{Introduction}

How deep is deep learning? Are neural networks ``discovering intricate structures" that support sophisticated generalization \cite{lecun2015deep}, or are they ``stochastic parrots" that simply memorize seen examples and recombine them in shallow ways \cite{stochasticparrots}?

We focus on this question in the area of open-ended text generation. 
Neural network language models (LMs) can generate grammatical, coherent 
text (\citeauthor{see2019massively}, \citeyear{see2019massively}; \citeauthor{brown2020language}, \citeyear{brown2020language}, section 3.9.4),
but the text alone cannot tell us 
if it was constructed by the model or copied from the training set.
We argue that it is important to disentangle these possibilities. That is, in addition to evaluating the \textit{quality} of generated text, as is already standard \cite{gatt2018survey,celikyilmaz2020evaluation}, we should also evaluate its \textit{novelty}.

Novelty is important for several reasons. 
From a linguistic perspective, one core component of knowing a language is the ability to combine familiar parts in novel ways \cite{chomsky1957syntactic,hockett1963problem}. 
From a machine learning perspective, models are meant to learn the training \textit{distribution}, not just memorize the training \textit{set}  \cite{dietterich1995overfitting}. 
Finally, on the more practical side, models that copy training data 
might leak sensitive information \cite{carlini2020extracting} or repeat hate speech \cite{stochasticparrots}.

In this work, to assess the novelty of generated text, we introduce a suite of analyses called RAVEN (\textbf{RA}ting \textbf{VE}rbal \textbf{N}ovelty).\footnote{GitHub code will be released soon.}\textsuperscript{,}\footnote{ \textit{Verbal} here uses its broad definition of ``linguistic" rather than the narrow definition of ``verb-related.'' This acronym refers to ``The Raven" by Edgar Allan Poe, in which the narrator encounters a mysterious raven which repeatedly cries out, ``Nevermore!" The narrator cannot tell if the raven is simply repeating something that it heard a human say, or if it is constructing its own utterances (perhaps by combining \textit{never} and \textit{more})---the same basic ambiguity that our paper addresses. This acronym is also a nod to  \citeauthor{stochasticparrots}'s (\citeyear{stochasticparrots}) comparison of LMs to another utterance-memorizing bird, the parrot.
} 
These analyses cover both sequential structure ($n$-grams) and syntactic structure. 
We apply these analyses to text generated by an LSTM, a Transformer, Transformer-XL, and all 4 sizes of 
GPT-2 (the largest LM for which we had access to the training data). 
Because there are many ways to generate text from LMs, we test 12 generation methods and 4 prompt lengths. 
As a baseline, we also analyze human-generated text from each model's test set.

We find that models display novelty for all aspects of structure that we analyze: they generate novel $n$-grams, novel morphological combinations, and novel syntactic structures. For instance, GPT-2 coins several types of novel words, including 
inflections (e.g., \textit{Swissified}) and derivations (e.g., \textit{IKEA-ness}), and 74\% of sentences generated by Transformer-XL have a syntactic structure that no training sentence has. 
Thus, \textbf{neural language models do not simply memorize; instead they use productive processes that allow them to combine familiar parts in novel ways}. 
Nonetheless, when considering small $n$-grams, these models are less novel than the baseline. For example, for each model, the baseline human-generated text has 1.4 to 3.3 times as many novel bigrams as the model-generated text does. For $n$-grams larger than 5-grams, models are \textit{more} novel than the baseline, but they still occasionally copy extensively: GPT-2 sometimes duplicates training passages that are \textbf{over 1,000 words long}.
Overall, by evaluating novelty, we gain a new
window into how models 
have or have not succeeded
at generalizing beyond their experience.

\section{Background}

\paragraph{Memorization and copying:} 
Neural networks are capable of extensive memorization: they can memorize randomly-labeled examples \cite{zhang2021understanding} and can reveal training data when subjected to adversarial attacks \cite{shokri2017membership,carlini2019secret,carlini2020extracting}.
We study copying in text generated under standard, non-adversarial conditions, a topic which four other works have touched on. 
\citet[][Section 8.2]{brown2020language} study copying of 8-grams by GPT-2, and \citet{lee2021deduplicating} study copying of 50-grams by Transformer LMs, while \citet{chen2021evaluating} and \citet{ziegler2021rote} look at copying of large $n$-grams in the code-generating model Codex. 
We perform a more comprehensive analysis of duplication:
We look across the full range of $n$-gram sizes and analyze a range of architectures and generation methods. Beyond $n$-grams, we also evaluate copying of other linguistic structures (e.g., dependency arcs). Thus, we focus on linguistic generalization, while past work 
focused more on data privacy.

\paragraph{NLG evaluation:} Careful evaluation of natural language generation (NLG) is crucial 
because of
the ELIZA effect \cite{weizenbaum1966eliza}, ``the susceptibility of people to read far more understanding than is warranted into strings of symbols---especially words---strung together by computers'' \cite{hofstadter1995fluid}. 
Because generated text has such power to guide people's views of AI, it is important to understand what capacities actually underlie the generation of that text in order to give a balanced view of the model.

Unfortunately, NLG evaluation is challenging because many NLG tasks are open-ended. For example, a dialogue system can generate multiple plausible responses for the same user input. 
The prevalent evaluation methods 
quantify the \textit{quality} of generated text, via a single holistic score \cite{Zhang2020bertscore} or via scores that focus on specific properties \cite{dou2021scarecrow} such as fluency \cite{mutton2007gleu}, 
coherence \cite{lapata2005automatic},
or factual accuracy \cite{kryscinski2020evaluating}.
We argue that evaluation of open-ended NLG should emphasize not only quality but also \textit{novelty}: \textbf{is the generated text novel, or does it simply duplicate part of the training set?} Novelty is important because, without novelty, quality does not reveal much about the model's abilities. 
For example, suppose that a model is being evaluated for coherence. If the model simply copies a paragraph from its training set, it will produce highly coherent text, but only because it has learned how to copy---not because it has learned how to be coherent.

The previously-studied attribute that is most similar to novelty is \textit{diversity} \cite{zhu2018texygen,hashimoto2019unifying}: can a model generate a diverse range of 
output sentences? Like novelty, diversity is rooted in differences between pieces of text. Despite this superficial similarity, novelty and diversity are distinct. Novelty covers how the generated text differs from the training set, while diversity covers how the generated text is different from other generated text. A model could be diverse but not novel (by copying a diverse set of training sentences), or novel but not diverse (by repeatedly generating the same novel sentence). 

Much discussion about evaluating LMs focuses on whether they \textit{understand} language \cite{bender2020climbing,marcus2020next}, whereas we assess the novelty of surface text. Thus, our main analyses only 
test 
whether models have abstractions governing form (e.g., syntax), 
not meaning.

\section{Motivation and approach}

\paragraph{Motivation:} The analyses in RAVEN are inspired by a scientific question: To what extent do NLG models have generalizable linguistic abilities? This question motivates our focus on novelty because only novel text can illustrate linguistic generalization. 
There may be some practical use cases for which novelty is not important---but for answering our scientific question, and for working toward general-purpose LMs that can handle unfamiliar situations, novelty is crucial.

\paragraph{Approach:} 
We generate many samples of text from LMs, and then evaluate 
how novel the 
text is. 
We assess novelty for two types of structure: $n$-grams and syntactic structure.
We count a generated structure as duplicated if it appears in the training set or the context (the concatenation of the prompt and the text that the LM has already generated based on the prompt); otherwise, it is novel. 

Copying is not necessarily undesirable \cite{Khandelwal2020Generalization}. For instance, some long $n$-grams might reasonably be duplicated from the training set, such as the title of a book. 
To contextualize a model's degree of duplication,
we compare the model-generated text to human-generated text from the model's (in-distribution) test set, which gives a baseline for how much duplication can be expected within the model's training domain. 
If the model is at least as novel as the baseline, we conclude that it is not copying excessively.
Two prior papers \cite{pannitto2020recurrent,meister2021language} have also analyzed models' linguistic abilities by comparing model-generated text to human-generated text, but neither of these focused on novelty.

\section{Experimental details}

\paragraph{Models:} To perform a controlled comparison across architectures, we used three models trained on the same dataset, namely Wikitext-103 \cite{merity2016pointer}. Wikitext-103 is a collection of high-quality Wikipedia articles tokenized at the word level. Its training set contains 103 million words.
Holding this training set constant, we compare the LSTM \cite{hochreiter1997long}, Transformer \cite{vaswani2017attention}, and Transformer-XL \cite[TXL;][]{dai2019transformer} architectures, chosen because they give examples of the two main types of processing prevalent in language modeling: recurrence (used in the LSTM) and self-attention (used in the Transformer), with TXL using both mechanisms.

In addition to these systematic analyses, we also analyzed GPT-2 \cite{radford2019language} as an example of a larger-scale Transformer LM 
(GPT-2 was the model with the largest training set that we could gain access to).
Unlike our other models, GPT-2 is trained on the WebText corpus, which is constructed from webpages linked to on Reddit. GPT-2 also differs from our other models in its tokenization scheme: All our other models use word-level tokenization (in which each token is a full word), but GPT-2 uses a subword tokenization scheme \cite{sennrich2016neural}. The WebText training corpus contains 7.7 billion words, making it much larger than Wikitext-103.
For more details about each model, see Appendix \ref{app:model_details}.

\paragraph{Prompts:} To generate text from a model, we input a prompt drawn from that model's test set, 
which comes 
from the same distribution as its training set. For Wikitext-103, we use 1000 prompts of length 512 words and have models generate 1000 words following the prompt. For WebText, we use 1000 prompts of length 564 subword tokens, and have models generate 1100 subword tokens; these numbers are 1.1 times the corresponding Wikitext-103 numbers because there are approximately 1.1 subword tokens per word in WebText. 
As our baseline human-generated text,
we use the text that follows the prompt in the corpus. 
For tokenization details, see Appendix \ref{app:tokenization}.

\paragraph{Decoding method: top-40 sampling:} As its prediction about which word will appear next, a language model outputs a probability distribution over the vocabulary. There are many ways to select a word to generate from this distribution, which are called \textit{decoding methods}. 

When evaluating a model's novelty, an important consideration is that novelty is not always positive: a model that generates random nonsense would be highly novel. Thus, we want to choose a decoding method that gives high-quality text, because high novelty is only positive when accompanied by high quality.
To this end, the decoding scheme that we use is top-$k$ sampling with $k=40$, where the model's distribution is truncated to the 40 highest-ranked words then renormalized and sampled from. 
We chose top-40 sampling because it is what \citet{radford2019language} used for GPT-2 and what \citet{dai2019transformer} used for TXL; because this method was selected by the creators of these models, we can be reasonably confident that it produces high-quality text from these models. For consistency, we use this same decoding scheme for our LSTM and Transformer, for which there is no established decoding scheme. 
For experiments with other decoding methods, see Section \ref{sec:decoding_scheme_experiment}.

\section{N-gram novelty}

We first investigate novelty at the level of $n$-grams, where an $n$-gram is a sequence of $n$ words.

\subsection{How often are generated $n$-grams novel for various values of $n$?}

\paragraph{Findings:} For small values of $n$, $n$-grams generated by LMs are rarely novel. For larger values ($n$ > 6), generated $n$-grams are almost always novel. 

\begin{figure}[t]
    \begin{subfigure}{\columnwidth}
        \centering
        \includegraphics[width=\textwidth]{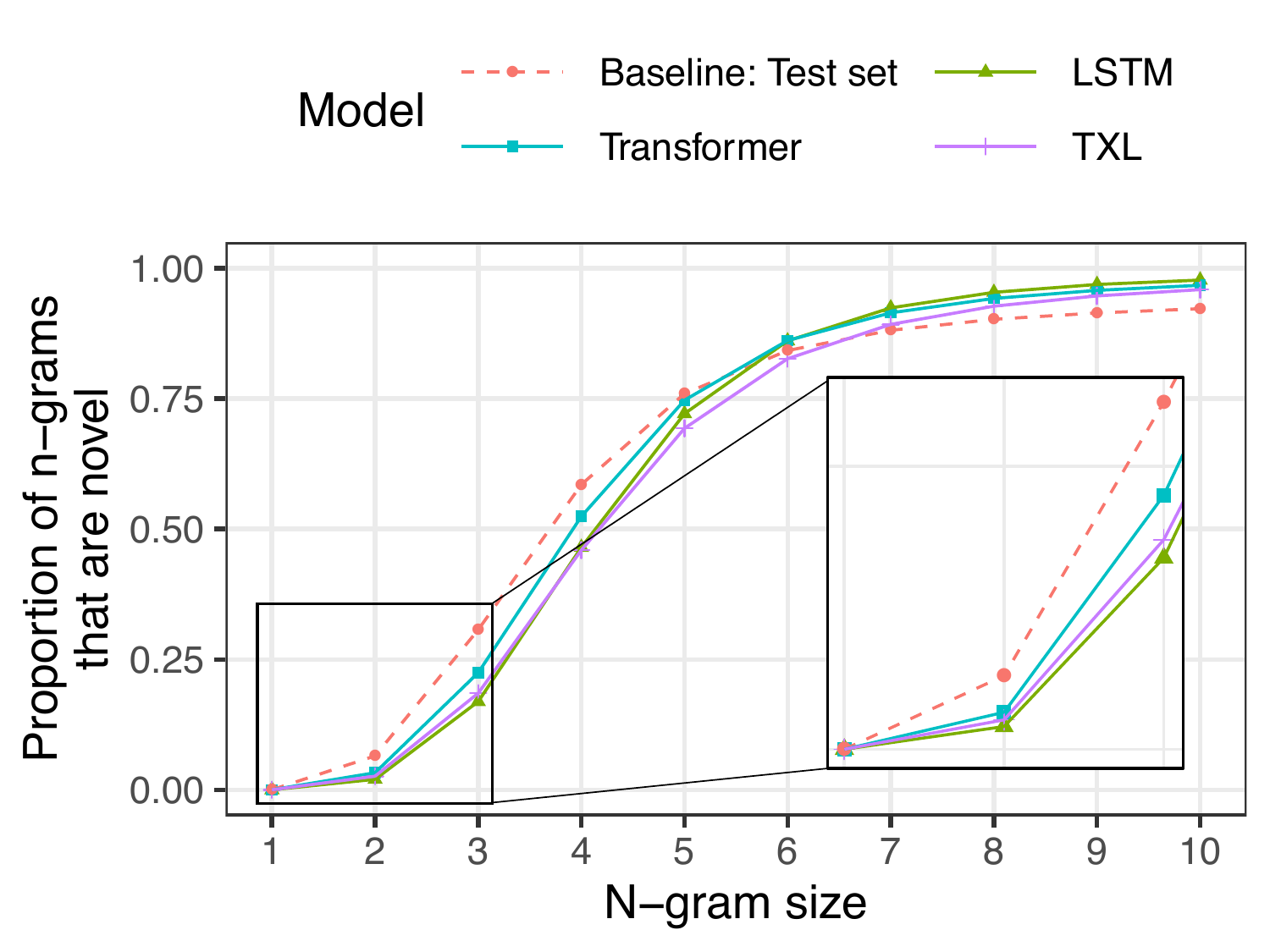}
        \subcaption{Models trained on Wikitext-103}\label{fig:ngram_overlap_wikitext}
    \end{subfigure}%
    
    \begin{subfigure}{\columnwidth}
        \centering
        \includegraphics[width=\textwidth]{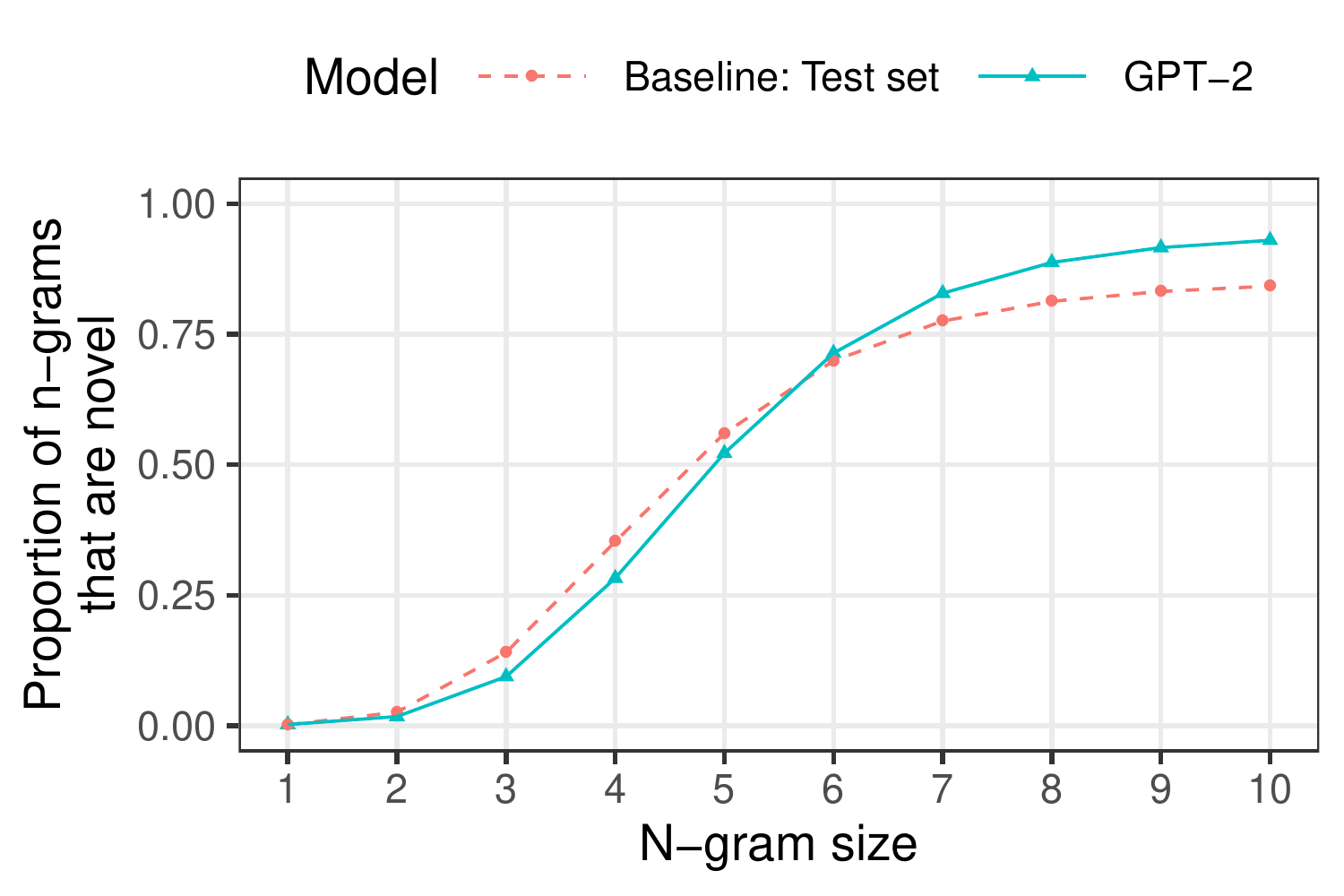}
        \subcaption{Models trained on WebText}\label{fig:ngram_overlap_webtext}
    \end{subfigure}
    \caption{Novelty by $n$-gram size. As baselines, we use text drawn from models' test sets.}
    \label{fig:ngram_overlap}
\end{figure}

\paragraph{Details:} Figure~\ref{fig:ngram_overlap} shows the proportion of generated $n$-grams that are novel, for values of $n$ from 1 to 10. We first note that the models are not merely copying: for all models, for $n$-grams of size 5 or larger, the majority of $n$-grams are novel.

We can obtain a more nuanced view by comparing the models to the baseline of text drawn from each model's test set.
For small $n$-grams ($n < 6$), models are less novel than the baseline. 
For instance, with Wikitext-103, the  baseline has 6\% of its bigrams being novel, while the models have 2\% to 3\% novelty; for trigrams, the  baseline has 31\% novelty while models have 17\% to 22\% (Figure \ref{fig:ngram_overlap_wikitext}). Thus, models are conservative at the small scale, rarely deviating from bigrams and trigrams they have seen before (though they do occasionally generate novel bigrams: see Appendix \ref{app:bigrams}). 
However, for larger $n$-grams ($n > 6$), the models are \textit{more} novel than the baseline.
Thus, at a larger scale, models cannot be described as mainly copying $n$-grams they have seen before.

Comparing the models to each other in the inset of Figure \ref{fig:ngram_overlap_wikitext}, we see that the LSTM is the least novel for small $n$-grams, while the Transformer is the most novel, and TXL falls in between. We conjecture the following explanation: Recurrence creates a recency bias \cite{ravfogel2019studying} which makes models that use recurrence likely to condition their predictions heavily on immediately preceding tokens, biasing them to memorize bigrams and trigrams. This explains why the LSTM duplicates so much: it operates entirely via recurrence. The Transformer
duplicates the least because it is driven entirely by self-attention (no recurrence), allowing it to condition on recent and faraway tokens more evenly. TXL
uses both recurrence and self-attention, placing it between the other two.

\subsection{Do models ever duplicate large $n$-grams?}

\textbf{Finding:} All models occasionally duplicate training set passages that are 100 words long or longer.

\paragraph{Details:} Models rarely duplicate $n$-grams larger than 10 tokens. However, there are occasional exceptions where models duplicate extremely long sequences. For instance, in our GPT-2 generated text, there are several cases where an entire generated passage (over 1,000 words long) appears in the training set.
To refer to these extreme cases, we use the term \textit{supercopying}, which we define as
the duplication of an $n$-gram of size 100 or larger. 
See Appendix \ref{app:supercopying} for examples of supercopied text.

\paragraph{What causes supercopying?} 
We
hypothesize that models supercopy passages that appear multiple times in the training set. For instance, the Wikitext-103 training set contains 159 articles about different instances of The Boat Race, a rowing competition: ``The Boat Race 1861," ``The Boat Race 2002," etc.
These articles are formulaic, with many sentences repeated across articles; e.g., the 100-gram in Appendix \ref{app:supercopying} that was generated by all 3 Wikitext-103 models occurs 56 times in the training set.
As evidence supporting this hypothesis, Figure \ref{fig:supercopycounts} (Appendix \ref{app:supercopying}) shows that
supercopied 100-grams appear far more times in the training set on average than randomly-selected 100-grams. 
This is consistent with the findings of \citet{lee2021deduplicating} and \citet{ziegler2021rote} that duplicated text tends to be common. \citet{carlini2020extracting} found that text can be extracted even if it only occurred once, but they used an adversarial method that deliberately tries to extract training data, instead of freely generating text.

\subsection{How is novelty related to the decoding scheme and the generated text's quality?}\label{sec:decoding_scheme_experiment}

\paragraph{Findings:} Changing decoding parameters can substantially alter a model's novelty: the novelty can be increased by increasing $p$ in top-$p$ sampling, $k$ in top-$k$ sampling, or the temperature. However, all modifications that increase the novelty of generated text also decrease the quality.

\paragraph{Details:} To get a single number that summarizes novelty, we use a new metric called the \textit{pointwise duplication score}: Each generated token gets a score quantifying the extent to which it duplicates previously-seen text. This score is equal to the size of the smallest novel $n$-gram that ends with this word. For example, if the word is the end of a novel 4-gram (e.g., \textit{these \underline{rules will not \textbf{be}}}), but all of the smaller $n$-grams ending with the word were duplicated (\textit{will not \textbf{be}}, \textit{not \textbf{be}}, and \textit{\textbf{be}}), then the pointwise duplication score is 4. To get the overall score, we average across the
tokens.
A downside of this basic score is that it can be heavily influenced by the extremely large duplication scores that arise from supercopying.
To address this factor, we
truncate each token's score at 5 before averaging
(see Appendix \ref{app:decoding_graphs} for untruncated results).

Using this score, we investigated a range of decoding methods.
Figure \ref{fig:decoding} (in Appendix \ref{app:decoding}) shows the effects of varying commonly-used decoding parameters. With top-$k$ sampling (truncating the distribution to the $k$ most probable tokens before sampling), increasing $k$ also increases novelty. With top-$p$ sampling (truncating the distribution to the the top $p$ probability mass before sampling; \citeauthor{holtzman2020curious}, \citeyear{holtzman2020curious}), increasing $p$ increases novelty. When using a temperature (which scales words' scores before taking the softmax), increasing the temperature increases novelty. 
All of these trends make intuitive sense: 
a small $k$, $p$, or temperature upweights the head of the model's distribution, and it makes sense that statistical learners would assign higher probability to things they have seen than things they have not, which would lead to the head of a model's distribution being less novel than the tail.

Could we make models perfectly novel just by changing the decoding scheme? 
Unfortunately, the decoding methods that increase novelty also decrease quality.
Measuring quality is challenging; ideally we would use human evaluations, but that is beyond the scope of this project because we have 336 conditions to evaluate (7 models with 4 prompt lengths and 12 decoding schemes). Instead, we use perplexity as a proxy for quality, under the assumption that high-quality text should have a low perplexity.
This assumption is certainly imperfect: text can have a low perplexity for degenerate reasons such as being repetitive \cite{holtzman2020curious}. 
Nonetheless, it can still give us a rough initial sense of general trends. 
We use GPT-2 to measure the perplexity of text generated by the LSTM, Transformer, and TXL; we use TXL to measure the perplexity of GPT-2 text. See Appendix \ref{app:perplexity} for discussion of these decisions.

\begin{figure}[t]
    \centering
    \includegraphics[width=\columnwidth]{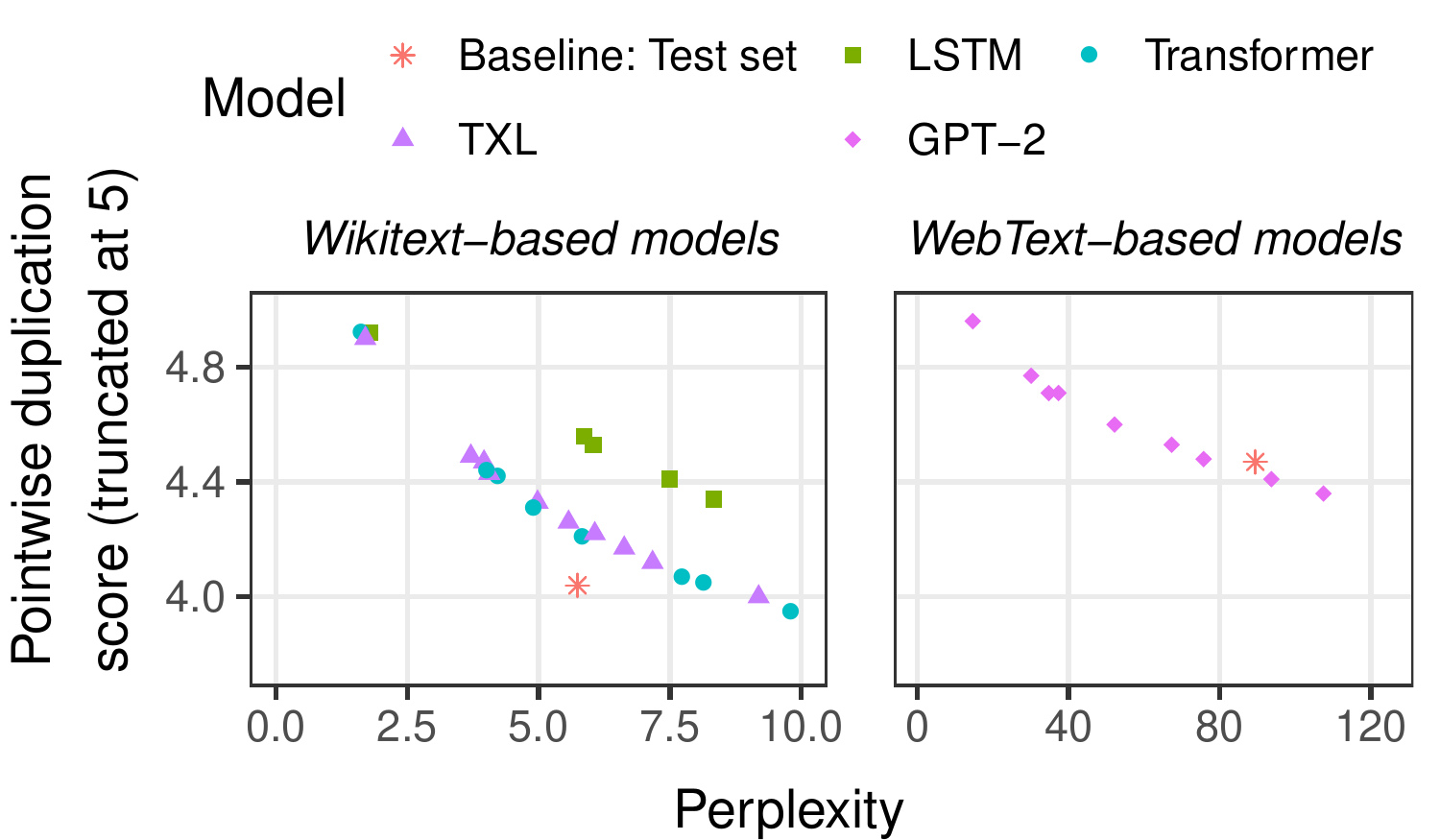}
    \caption{Manipulations to the decoding scheme that result in higher-quality text (i.e., lower perplexity; $x$-axis) also result in decreased novelty (i.e., a greater degree of duplication; $y$-axis). Each point shows a different decoding scheme.}
    \label{fig:quality}
\end{figure}

Figure \ref{fig:quality}
shows a clear tradeoff between novelty and quality.
None of models trained on Wikitext do as well as the baseline at managing 
this
tradeoff.
However, a model's perplexity does not entirely determine its level of novelty: Both Transformer architectures do better at this tradeoff than LSTMs, showing that it is possible to improve on this tradeoff using architectural innovations.

In contrast to the Wikitext-103 models, 
GPT-2 performs similarly to the baseline at the quality-novelty tradeoff.
The GPT-2 decoding scheme that comes closest to the baseline is top-$p$ decoding with $p = 0.95$; this achieves a perplexity of 93.7 (baseline: 89.4) and a truncated pointwise duplication score of 4.41 (baseline: 4.47).
Why does GPT-2 (with the right decoding scheme) outperform the Wikitext-103 models at matching the quality and novelty of its baseline? It is unlikely that the model architecture is the reason because GPT-2 is similar in architecture to the Wikitext-103 Transformer. In addition, although GPT-2 is our largest model, we doubt that model size is the explanation: GPT-2 Small shows similar results even though it is smaller than TXL.
It may then be that training set size is the key factor, as WebText is much larger than Wikitext-103. Alternately, the WebText baseline might be easier to meet than the Wikitext baseline, because the generic Internet text in WebText is generally lower-quality than the curated Wikipedia text in Wikitext-103.

\subsection{Other $n$-gram analyses}

Additional analyses are detailed in the appendices. We find that model size (Appendix \ref{app:size}) and prompt length (Appendix \ref{app:prompt_length}) do not have a clear effect on novelty; novelty is influenced by position within the generated text for some models, but the effect is small (Appendix \ref{app:position}); and our novelty results do not change much if we only consider duplication from the training set rather than duplication from the context and/or training set (Appendix \ref{app:duplication_training_context}).

\section{Syntactic novelty}

\paragraph{Findings:} At the level of global sentence structure, models show a high degree of syntactic novelty, with the majority of generated sentences having an overall syntactic structure that no training sentence has. Models also display some novelty for more local structure (e.g., individual dependency arcs), but they have much less novelty for local structure than the baselines do.

\paragraph{Details:} We have seen that models display some novelty. How deeply does their novelty extend? Are they just inserting words into memorized templates, or performing deeper syntactic composition?  
To investigate this question, we parsed 
our generated text and 
our models' 
training data using state-of-the-art constituency \cite{kitaev2018constituency} and dependency \cite{zhang2020efficient} parsers.
We then evaluated novelty for 7 aspects of syntax.

Though current parsers perform well, they are not perfect, so we cannot completely trust the parsers' output.
This is particularly a problem because the cases that are important to us (novel ones) are especially likely to confuse parsers.
To address this issue, we manually analyzed the examples identified as novel to estimate the parsers' error rates  
(details in Appendix \ref{app:syntax_vetting}). We concluded that 4 of the 7 attributes that we analyzed were handled accurately enough by the parsers for us to report numerical results, which are in Figure~\ref{tab:syntactic_novelty}. Here is a description of these attributes:

\vspace{-6pt}

\begin{itemize}[noitemsep]
    \item \textbf{POS sequence:} the sequence of part-of-speech tags for the words in the sentence.
    \item \textbf{Parse structure:} the sentence's constituency tree minus the leaves (the words).
    \item \textbf{Dependency arc:} a 3-tuple of a dependency relation (e.g., \textit{nsubj}) and the two words that hold that relation.
    \item \textbf{Dependency role:} a 3-tuple of a word, a dependency relation that the word is part of, and the word's position in that relation; e.g., ``\textit{watch} as the head of an \textit{nsubj} relation"
\end{itemize}

\vspace{-6pt}

\noindent
For POS sequences and parse structures, there is a high degree of novelty: across all models and baselines, the majority of sentences have an overall structure that no training sentence has. In addition, there is little difference between the models and the baselines. For the more local structure of dependency arcs and dependency relations, 
the baselines are far more novel than the models. This is similar at a high level to our $n$-gram results: Models tend to be less novel than the baseline for local structure (small $n$-grams), but they are more novel than the baseline for more global structure (large $n$-grams). See Appendix \ref{app:mismatch} for an example of how such a local/global mismatch is possible, and see Appendix \ref{app:syntactic_novelty_examples} for specific examples of syntactic generalization (e.g., nouns that were used as direct objects in generated text when they have never appeared as direct objects in training).

\begin{figure}
    \centering
    \begin{tabular}{lp{0.7cm}p{0.8cm}p{0.7cm}p{1cm}} \toprule
         & POS seq. & Parse struct. & Dep. arcs & Dep. roles  \\ \midrule
        Wiki baseline & 0.75 & 0.76 & 0.13 & 0.0050 \\
        LSTM & 0.77 & 0.78 & 0.07 & 0.0015 \\
        Transformer & 0.75 & 0.75 & 0.08 & 0.0024 \\
        TXL & 0.74 & 0.74 & 0.07 & 0.0021 \\ \midrule \midrule
        Web baseline & 0.59 & 0.61 & 0.05 & 0.0015 \\
        GPT-2 & 0.62 & 0.64 & 0.03 & 0.0007  \\ \bottomrule
    \end{tabular}
    \caption{Syntactic novelty. Abbreviations: \textit{seq}=sequence; \textit{dep}=dependency; \textit{struct}=structure.}
    \label{tab:syntactic_novelty}
\end{figure}

\section{Analysis}\label{sec:analysis}

We finish with some manual analysis of novel generated text. 
Such analysis is labor-intensive; to use this labor most effectively,
we focus exclusively on GPT-2 because it is the strongest-performing 
model.
For this initial analysis, 
we study only the novel unigrams that GPT-2 generates; GPT-2 uses subword tokenization, so it 
can generate novel words by combining seen subwords in novel ways. 
See Appendices \ref{app:morphology_categorization} and \ref{app:additional_examples} for a detailed taxonomy of the novel words that GPT-2 generates. Here in the main paper, we focus only on 4 targeted questions about these novel words. 
Throughout this section, any word in boldface is novel.

\subsection{When GPT-2 generates novel words, are they morphologically well-formed?}\label{sec:analysis_morph}

\paragraph{Finding:} The vast majority of GPT-2's novel words (96\%) are well-formed (Figure \ref{tab:syntactic_well_formedness}); this is, however, lower than the baseline (99\%).

\paragraph{Specific categories:} Forming English plurals requires a choice between two orthographic forms, \textit{-s} and \textit{-es}. In 72 of the 74 novel plurals, GPT-2 made the correct choice (e.g., \textit{Brazilianisms}, \textit{Fowleses}). The two incorrect examples were \textit{1099es} and \textit{SQLes}. Similarly, forming English possessives requires a choice between \textit{-'s} and \textit{-'}. Here, GPT-2 makes the correct choice in 135 out of 136 novel possessives (e.g., \textit{Flexagons'}, \textit{Runerealm's}), with the only error being \textit{watchmakers's}.

Acronyms provide another case for which we can easily quantify well-formedness. Our GPT-2-generated text contains 75 examples of novel acronyms that appear along with the full version of what the acronym stands for. In 72\% of cases, the acronym is not a suitable abbreviation (well-formed example in \ref{ex:acronymgood}, ill-formed example in \ref{ex:acronymbad}). 
There are valid reasons why an acronym might not match its expansion; e.g., sometimes English-language publications will translate a non-English phrase but not the
acronym derived from it, giving results such as 
\textit{Doctors Without Borders (MSF)}. However, in our baseline text, 
17 of the 21 acronyms that appeared with expansions were suitable,
so GPT-2 is still not suitable nearly as often as the baseline (28\% vs. 81\%).

\ex. West of England Cricket and Athletics Club (\textbf{WECAC})\label{ex:acronymgood}

\ex. Extremely Large Interactive Neutrino Experiment (\textbf{ELIGO})\label{ex:acronymbad}

\noindent
Some additional examples of success involve suffixes that require the stem to change spelling, with GPT-2 successfully making the change (\ref{ex:spellign_change}). Some additional mistakes are the use of a plural noun as the first component of a compound (\ref{ex:plural_compound}) and overregularization, namely using the regular suffix \textit{-th} instead of the exceptional suffix \textit{-nd} (\ref{ex:overreg}).

\ex.\label{ex:spellign_change}
\a. by `` \textbf{cookying} " certain searches on the internet
\b. \textbf{Summission} base camp
\d. the \textbf{ridiculousities} of war

\ex. \label{ex:plural_compound} 
The...rivers had their \textbf{headswaters} in a larger basin

\ex. \label{ex:overreg} 
the \textbf{752th} year\label{ex:752th}

\subsection{When GPT-2 generates novel words, do they fit within their syntactic context?}

\begin{figure}
    \centering
    \resizebox{\columnwidth}{!}{
    \begin{tabular}{ccccc} \toprule
        & \multicolumn{2}{c}{Morphology} & \multicolumn{2}{c}{Syntax} \\ \cmidrule(lr){2-3} \cmidrule(lr){4-5}
         & Baseline & GPT-2 & Baseline & GPT-2 \\ \midrule
        Correct & 0.99 & 0.96 & 0.97 & 0.94 \\
        Incorrect & 0.01 & 0.02 & 0.00 & 0.01 \\
        Unclear & 0.00 & 0.02 & 0.03 & 0.05 \\ \bottomrule
    \end{tabular}
    }
    \caption{Syntactic and morphological usage of novel words}
    \label{tab:syntactic_well_formedness}
\end{figure}

\paragraph{Finding:} The vast majority of GPT-2's novel words (94\%) are used in grammatically-correct contexts (Figure \ref{tab:syntactic_well_formedness}), but it does make more errors than we see in the baseline (e.g., \ref{ex:gpt2_syntax_errors}).

\ex. \label{ex:gpt2_syntax_errors}
\a. the manicure that I did for \textbf{Sally-themed} a year ago \label{ex:syntax_mistake_main_1}
\b. Slicex \textbf{load-samples} provides a single button\label{ex:syntax_mistake_main_2}

\paragraph{Agreement:} Despite these errors the vast majority of cases have proper syntax. Some particularly impressive cases involve novel plural words. First, (despite the one mistake in \ref{ex:syntax_mistake_main_2}), GPT-2 generally does well at providing plural verbs (underlined) to agree with novel plural nouns, whether the verb appears after the noun (\ref{ex:subjverb_basic}) or before the noun in the context of a question (\ref{ex:subjverb_inversion}). In (\ref{ex:subjverb_rc}), it correctly uses a plural verb for both verbs that agree with the novel plural subject---a verb within the relative clause, and a verb after it. The correct agreement with the verb after the relative clause is especially impressive because, in both sentences, there are 3 singular ``distractors" (italicized) between the subject and the verb. See \citet{haley2020bert} for similar observations but with BERT instead of GPT-2.

\ex. \label{ex:subjverb_basic}
\a. We know that \textbf{M-Sinks} \underline{need} a target
\b. \textbf{Torpexes} \underline{are} small hardpoints

\ex. \label{ex:subjverb_inversion}
Why \underline{do} \textbf{SQLes} have to change

\ex. \label{ex:subjverb_rc}
\a. The \textbf{Huamangas} , who \underline{are} descendants of indigenous people who lived on the \textit{Isthmus} of \textit{Tehuantepec} before it was covered by \textit{farmland} , \underline{have} been demanding that the federal government address the issue of climate change .
\b. \textbf{FOIA-requesters} who \underline{think} an \textit{agency} has a good \textit{reason} for withholding \textit{information} \underline{are} not always given a second opportunity to press their case .

\paragraph{Other plural-relevant syntax:} Beyond agreement, syntactic consequences of plurality are observed in a few other places as well: in using the plural possessive form that is just an apostrophe instead of the singular form of \textit{-'s} (\ref{ex:plural_possessive}); in having the pronouns that are coreferential with the noun be plural as well (\ref{ex:plural_pronoun}); and in following determiners that require a plural noun (\ref{ex:plural_determiner}).

\ex. \label{ex:plural_possessive}
The \textbf{Fowleses} \underline{'} lawyer 

\ex. \label{ex:plural_pronoun}
\a. I love \textbf{Klymits} , but it has been nearly impossible for us to find \underline{them} in stores .
\b. The \textbf{Sarrats} were lucky to have her as part of \underline{their} lives

\ex. \label{ex:plural_determiner}
\a. \underline{these} small \textbf{townites} 
\b. so \underline{many} \textbf{Brazilianisms} 

\paragraph{Incrementing/ordering:} Another type of inter-word relation that GPT-2 appears to have learned is  incrementing/ordering, with examples in the Appendix. In (\ref{ex:increment_numbers}), GPT-2 increments numbers from \textit{Firstly} to \textit{Fourteenthly}, with \textit{Thirteenthly} and \textit{Fourteenthly} being novel. In (\ref{ex:increment_variables}), it increments the letters at the ends of variable names in computer code, going from \textit{multiplyx} to \textit{multiplyy} to \textit{multiplyz}. Finally, in (\ref{ex:increment_alphabet}), the prompt ends with an alphabetical list of companies, and GPT-2 continues this list,
staying mostly in alphabetical order and including many novel words along the way.

\paragraph{Quotation marks:} A final aspect of sentence structure that we analyze is putting words within quotation marks. In human-generated text, there is an association between novel words and quotation marks: words are much more likely to appear inside quotation marks if they are novel, and they are much more likely to be novel if they appear inside quotation marks. This association is also present in GPT-2's generated text (Figure \ref{tab:quotations}), e.g.:

\begin{figure}
    \centering
    \begin{tabular}{ccc} \toprule
         & Baseline & GPT-2 \\ \midrule
        p(novel) & 0.0022 & 0.0022 \\
        p(novel | in quotes) & 0.023 & 0.028 \\
        p(in quotes) & 0.0016 & 0.0015 \\
        p(in quotes | novel) & 0.016 & 0.019 \\ \bottomrule
    \end{tabular}
    \caption{Quotation mark statistics. Computed over all word-level (not subword-level) unigrams.}
    \label{tab:quotations}
\end{figure}

\ex. 
\a. The `` \textbf{proto-poetry} " of modern times
\b. the `` \textbf{un-competition} " that is happening 

These results suggest that GPT-2 might encode some version of the concept ``novel word" which it can accesses when determining whether to include quotation marks.

\subsection{When GPT-2 generates novel words, do they result in reasonable meanings?}

\paragraph{Finding:} GPT-2 does less well in this area than in morphology and syntax, consistent with claims \cite{bender2020climbing} that language models only learn form, not meaning (Figure \ref{tab:semantic_well_formedness}).

\begin{figure}
    \centering
    \begin{tabular}{lcc} \toprule
         & Baseline & GPT-2 \\ \midrule
        Clearly suitable & 0.327 & 0.209 \\
        Potentially suitable & 0.643 & 0.587  \\
        Probably not suitable & 0.002 & 0.044 \\
        Clearly unsuitable & 0.001 & 0.072 \\
        Unclear & 0.028 & 0.089 \\ \bottomrule
    \end{tabular}
    \caption{How semantically suitable novel words are for their contexts.}
    \label{tab:semantic_well_formedness}
\end{figure}

\paragraph{Examples:} There are some generated examples for which there is clear evidence that the meaning is incorrect (\ref{ex:semantic_errors}). One frequent source of mistakes is numbers, revealing a general lack of understanding of the quantities that these numbers represent. Numerical errors include incorrect conversions (\ref{ex:lbkg}), physical impossibilities (\ref{ex:conversion_subtraction}), and inconsistent exchange rates (\ref{ex:kes1}): 

\ex. \label{ex:semantic_errors}
\a. An old school English term is a \textbf{Brazilianism} .
\e. ...adding an optional `` \textbf{no-knockout} " version ... so you can actually be knocked out 

\ex. \label{ex:number_conversion} 
\a.  a \textbf{1,240-lb} . ( \textbf{735-kg} ) device \label{ex:lbkg}
\b. the ... 4ml tank holds \textbf{10.4ml} of e juice .\label{ex:conversion_subtraction}
\b. \textbf{KES50} ( £ 3.50 )\label{ex:kes1} ... \textbf{KES100} ( £ 4.00 )\label{ex:kes2} ... \textbf{KES300} ( £ 4.50 )\label{ex:kes3} ... \textbf{KES200} ( £ 2.50 )\label{ex:kes4}

\noindent
Nonetheless, there are also some positive examples where GPT-2 essentially provides a clear and accurate definition of the novel word or otherwise makes use of all aspects of the word:

\ex.
\a. ... the process of \textbf{re-nitrification} that gives them a new supply of nitrogen
\b. the concept of ` \textbf{co-causation} ' , in which effects are thought to be caused by causes that act in parallel
\c. the `` \textbf{bondbreaking} enchantment " , which...permanently breaks any binding .

\subsection{What does GPT-2 generalize from?}\label{sec:genfrom}

We have seen that GPT-2 generates some novel words.
What types of generalization does GPT-2 use to create these words? There are two basic types of generalization that might be employed 
\cite{prasada1993generalisation,albright2003rules,dasgupta2021distinguishing}. 
First, a novel word could be created by a compositional rule that builds up word parts (\ref{ex:composition}). Alternatively, a novel word could be created via a similarity-based analogy, with similar word parts replacing each other, such as swapping \textit{giraffe} and \textit{elephant} (\ref{ex:analogy}). 

\ex.
\a. \textit{elephant} + \textit{-s} = \textit{elephants}\label{ex:composition}
\b. \textit{giraffes} - \textit{giraffe} + \textit{elephant} = \textit{elephants}\label{ex:analogy}

\noindent
As these examples show, a given word (e.g., \textit{elephants}) could have been formed in either of these ways, so we can never be certain about which approach GPT-2 is using. However, based on some examples which are reasonably clear, we suspect that GPT-2 employs both types of generalization.

\paragraph{Generalization by composition:}
In a few cases, GPT-2 generates a novel word whose stem never appears in training but does appear in the context (the prompt plus the previously-generated words): see (\ref{ex:gen_from_context_plural}). We believe that these examples are best explained by composition: analogy requires some notion of similarity between the two word parts being swapped, and it is unlikely that the model would have such similarity notions for a word stem it has never seen before. Thus, we think these examples are better understood as the model adding a prefix or suffix to a word from its context, without direct reference to another word that has that prefix or suffix---a form of composition.

\ex. \label{ex:gen_from_context_plural}
\a. using the \textbf{LHAW} to take out other \textbf{LHAWs} 
\b. Pelagic \textbf{epineopterygoid} ... \textbf{Sub-epineopterygoid} , N. scapulatus

\paragraph{Generalization by analogy:}
Appendix \ref{app:torero} contains one piece of generated text which we believe provides clear evidence for analogy.
The prompt for this generation contains the real English word \textit{torero} (borrowed from Spanish), which means ``bullfighter." The generation then contains several alternate forms of this word (some with plural inflection): \textit{\textbf{tear}ro}, \textit{\textbf{torn}ro}, \textit{\textbf{tearing}ros}, and \textit{\textbf{tears}ros} (e.g., in the sentence \textit{tearingros are taught to avoid the horns}). It appears, then, that GPT-2 has taken the word \textit{\textbf{tore}ro} and replaced the first 4 letters (\textit{tore}) with other forms of the verb \textit{tear}: \textit{tear}, \textit{torn}, \textit{tearing}, and \textit{tears}.
There is no morphological process in English that  adds \textit{-ro} to verbs, so it is unlikely that these words were generated via composition; instead, it seems more likely that they were generated via analogy.

\section{Discussion}

Using our analysis suite RAVEN, we have found that models generated many types of novelty---novel $n$-grams of all sizes, novel syntactic structures, and novel morphological combinations. However, they also show many signs of copying: for local structure, they are substantially less novel than the baseline; and we see occasional large-scale copying, such as duplicating passages from the training set that are over 1,000 words long. 

\noindent
\paragraph{Compositionality:}
Compositional generalization (combining familiar parts in novel ways) is often discussed in the context of out-of-distribution generalization \cite{hadley1994systematicity,hupkes2020compositionality,keysers2020measuring,li2021compositional}, typically relying on synthetic datasets to test models' compositional abilities \cite{lake2018generalization,kim2020cogs,mccoy2020does}. Our baseline results in Figure~\ref{tab:syntactic_novelty} show that compositional generalization is important even for in-distribution test sets drawn from large-scale natural corpora.
Most notably, the majority of test sentences had a sentence-level syntactic structure that had never appeared in the training set.

Turning to the model results in Figure \ref{tab:syntactic_novelty}, all models displayed nonzero rates of compositional generalization,
giving an existence proof that they can perform these types of generalization. Nonetheless, the models' scores are lower than the baseline,
so their 
generalization
might be limited to particular subcases, instead of being as 
general as human generalization. 
In the opposite direction, however, we also found examples where GPT-2 generalized too freely, such as generating the word \textit{752th} (Section \ref{sec:analysis_morph}). We conclude that it may not be enough to simply encourage models to be systematic, because language is not completely systematic. Instead, we need models that can both figure out linguistic rules and recognize exceptions to those rules  
\cite{o2015productivity,yang2016price}.

\paragraph{Evaluating novelty:}
Our core message is that novelty has not received the attention it deserves in NLG evaluation. For generated text to truly illustrate a model's generative capabilities, that text must be novel---otherwise, it may only illustrate the model's ability to copy but not other abilities (e.g., the ability to be coherent).
We recommend using the level of novelty found in an in-distribution test set as a baseline: if the model is at least as novel as this baseline, we can rule out the possibility that it is copying excessively.

Recent increases in training set sizes make it especially critical to check for novelty because the magnitude of these training sets can break our intuitions about what can be expected to occur naturally. For instance, some notable work in language acquisition \cite[e.g.][]{kuczaj1977acquisition,marcus1992overregularization} relies on the assumption that regular past tense forms of irregular verbs (e.g., \textit{becomed}, \textit{teached}) do not appear in a learner's experience, so if a learner produces such words, they must be novel to the learner. However, it turns out that, for all 92 basic irregular verbs in English, the incorrect regular form appears in GPT-2's training set; details are in Appendix \ref{app:overregularization_cvc}, along with results for another category often assumed to be novel in human experiments, namely nonsense words such as \textit{wug} \cite{berko1958child}. 
Thus, when we are using models trained on such large-scale datasets, it is not safe to assume that something is absent from the training set; we must explicitly check.

\paragraph{Improving novelty:}
One straightforward approach for increasing novelty
would be to modify the sampling procedure to suppress highly-copied outputs, similar to penalties used to prevent repetition \cite{keskar2019ctrl}.
Another approach would be to implement more nuanced forms of deduplication during training: We found that supercopying mainly arises when there is repetition in the training set, so eliminating such repetition might improve models' novelty. Indeed, concurrent work \cite{lee2021deduplicating} has shown that deduplication can substantially decrease copying of 50-grams from the training set. 

Ideally, however, we would find ways to decrease copying that are deeper, without requiring post-hoc modifications to the training data and sampling procedure. In humans, novelty has long been attributed to the usage of symbolic, compositional rules. Thus, greater novelty might be achieved through models that build in compositional mechanisms, such as RNNGs \cite{dyer2016recurrent} and TP-Transformers \cite{schlag2019enhancing}.

Alternatively, one major difference between text generation in humans and neural LMs is that humans usually have a meaning that they want to express that guides their text generation, whereas most neural text generation involves no explicit plan. This difference may partly explain the ways in which models are less novel than humans: since models mainly manipulate text alone, they fall back to repeating text they have seen before.
Thus, novelty may be improved by incorporating more explicit semantic planning \cite{rashkin2020plotmachines}.

\section{Conclusion}

In machine learning, it is critical to evaluate models on a withheld test set. Due to the open-ended nature of text generation, a model's generated text might be copied from the training set, in which case it is not withheld---so using that data to evaluate the model (e.g., for coherence or grammaticality) is not valid. Thus, it is important to consider novelty when evaluating text generation. We have introduced RAVEN, an analysis suite covering both sequential structure and syntactic structure, and have applied it to several models, showing that models are rarely novel for local structure but are often novel for larger-scale structure; however, they occasionally copy even very long passages. 
Beyond text generation, we hope that our work will motivate more careful consideration of maintaining withheld splits between training sets and evaluation sets across NLP.

\section*{Acknowledgments}

We thank OpenAI for providing access to the WebText dataset.
For helpful comments and discussion, we are grateful to Suhas Arehalli, Saadia Gabriel, Coleman Haley, Yichen Jiang, Nebojsa Jojic, Najoung Kim, G\'{e}raldine Legendre, Grusha Prasad, Eric Rosen, Sebastian Schuster, Paul Soulos, Shiyue Zhang, the Deep Learning Group at Microsoft Research Redmond, the Johns Hopkins Neurosymbolic Computation Lab, the NYU Computation and Psycholinguistics Lab, and the NYU Machine Learning for Language Group. 
Any errors are our own.
For technical assistance with Hugging Face, we thank Teven Le Scao and Patrick von Platen.
We are also grateful to the Maryland Advanced Research Computing Center (MARCC) for providing the computing resources used in our experiments. 
The raven image used in our title comes from Pixabay user Nika\_Akin.\footnote{\url{https://pixabay.com/illustrations/crow-raven-black-dark-bird-ink-4779560/}}

Portions of this research were supported by the National Science Foundation Graduate Research Fellowship Program under Grant No. 1746891. Any opinions, findings, and conclusions or recommendations expressed in this material are those of the authors and do not necessarily reflect the views of the National Science Foundation.

\bibliography{tacl2018}
\bibliographystyle{acl_natbib}

\appendix

\section{Model details}\label{app:model_details}

In all cases, we generated text using a beam size of 1. Except where otherwise mentioned, the decoding scheme was top-40 sampling.

\paragraph{LSTM:} The LSTM architecure was introduced by \citet{hochreiter1997long}. We used the implementation from \citet{gulordava2018colorless} to train a 2-layer LSTM language model with a hidden state size of 1024 and dropout of 0.2.\footnote{We used the hidden size of 1024 following \citet{grave2017improving}. For the other hyperparameters, we searched over two numbers of layers (1 layer or 2 layers) and four dropouts (0.1, 0.2, 0.5, and 0.65) and chose the combination that gave the lowest perplexity on the validation set.} We trained this model using stochastic gradient descent with an initial learning rate of 20. The learning rate was divided by 4 whenever the validation loss did not improve from the previous epoch, and we halted training when validation loss failed to improve for two consecutive epochs. This model achieved a perplexity of 45.7 on the test set. The best-performing LSTM trained on Wikitext-103 that we are aware of from past literature is in \citet{grave2017improving}, and its test-set perplexity was 48.7, so we conclude that our newly-trained model is a strong LSTM. The fact that our model outperforms the previous one is likely due to its number of layers: we used 2 layers, while \citeauthor{grave2017improving} do not mention the number of layers used, which we believe most likely means that they used 1 layer.

\paragraph{Transformer:} As a basic Transformer language model, we use the trained model from \citet{Khandelwal2020Generalization} (note that we only use the base Transformer, without the datastore that \citeauthor{Khandelwal2020Generalization} introduce). This model uses the Transformer decoder architecture from \citet{vaswani2017attention} trained in fairseq \cite{ott2019fairseq} using tied weights \cite{press2017tied}, adaptive inputs \cite{baevski2018adaptive}, and an adaptive softmax \cite{grave2017efficient}. Its perplexity on the test set is 18.65.

\paragraph{Transformer-XL:} We use the HuggingFace \cite{wolf2020transformers} version of Transformer-XL \cite{dai2019transformer}. Transformer-XL is a Transformer modified to include a recurrence mechanism and a different positional encoding scheme, which are intended to help it process long-distance dependencies. Its perplexity on the test set is 18.30.

\paragraph{GPT-2:} We use the HuggingFace \cite{wolf2020transformers} version of GPT-2 \cite{radford2019language}, which is a decoder-only Transformer architecture \cite{vaswani2017attention} trained on the WebText corpus, which is constructed from webpages linked to on Reddit which have received at least 3 karma. 
There are four sizes of GPT-2 available from HuggingFace (\texttt{gpt2}, \texttt{gpt2-medium}, \texttt{gpt2-large}, and \texttt{gpt2-xl}). The largest of these, \texttt{gpt2-xl}, is the one that \citet{radford2019language} refer to as \textit{GPT-2} and is the one we use for our experiments except where otherwise indicated.

\section{Tokenization and other text preprocessing}\label{app:tokenization}

\paragraph{Prompts:} For the Wikitext prompts, we used prompts of length 0, 16, 128, and 512; most of our experiments used only the length-512 prompts. 
For the WebText prompts, the prompt lengths were 0, 18, 141, and 564 subword tokens; however, we extended the prompt past that length as was necessary to ensure that the prompt ended with a complete word. If this required adding more than 10 additional tokens, we discarded the prompt and sampled a new one. The WebText prompt lengths were chosen to be approximately 1.1 times the Wikitext prompt lengths because there are approximately 1.1 WebText subword tokens for every word.

As our baseline text, we used the words that followed the prompt in the test set.
Due to the small size of the Wikitext-103 test set (it contains approximately 245,000 tokens, while the continuations following the prompts total 1,000,000 tokens), some of the Wikitext-103 continuations that were used to make the baseline text necessarily have parts that overlap with parts of other continuations, but no two continuations are identical.
For the Webtext baseline, there was no such overlap because the dataset was large enough to avoid it.

\paragraph{$N$-gram novelty:} For computing $n$-gram novelty, we did not perform any processing of Wikitext text or the text generated by Wikitext models; thus, this text uses the tokenization from the Wikitext-103 dataset, which is a slightly modified version of the Moses tokenizer \cite{koehn2007moses}. For WebText text and text generated by GPT-2, we converted GPT-2's subword IDs into text using the GPT2Tokenizer from the HuggingFace Transformers library (version 2.11.0). We then replaced each newline with the token \textit{\&NEWLINE;} (which never occurs in WebText), to be consistent with Wikitext, in which each newline is a token. Wherever there were multiple spaces in a row, we replaced them with a single space. We then tokenized this text using the Moses tokenizer \cite{koehn2007moses} and used the resulting tokens to compute $n$-gram novelty.

\paragraph{Syntactic novelty:} Many of our syntactic analyses operate at the level of sentences. Thus, we first sentence-tokenized our text using the NLTK sentence tokenizer\footnote{\url{nltk.org}} and then parsed them using a constituency parser \cite{kitaev2018constituency} and dependency parser \cite{zhang2020efficient}. These parsers perform their own tokenization, so we wanted to provide them with untokenized text. For WebText baselines and GPT-2 text, this was straightforwardly accomplished by not performing word-level tokenization before passing text to the sentence tokenizer and the parser. For Wikitext baselines and text generated by Wikitext-based models, we first detokenized the text using the Moses detokenizer \cite{koehn2007moses} and then passed the detokenized text to the parsers.

\paragraph{Analyses:} For identifying novel unigrams for our analyses in Section \ref{sec:analysis}, we used the GPT-2 generated text as it was tokenized for the $n$-gram novelty evaluations, and we then used the $n$-gram novelty annotations to determine which unigrams were novel.

\section{Novel bigrams}\label{app:bigrams}

Figure \ref{fig:novel_bigrams} shows examples of novel bigrams generated by each of our models.

\begin{figure*}[t]
    \centering
    \begin{tabular}{lc} \toprule
     Generation method & Examples of novel bigrams in context  \\ \midrule
     LSTM & ... surrounded by a \textbf{tall brick} , square with a flat facade ... \\
     & ... including the placement of \textbf{some prosimians} , with both ...  \\
     &  ... the first legislative assembly of the newly \textbf{created territory} ... \\ \midrule
     Transformer & ... it is revealed to be the \textbf{poet Gidding} , now living ... \\
     & Archaeological findings from La \textbf{Venta indicate} that ... \\
     &  ... proposed by \textbf{Edward Armour} in his wooden Ironclads . \\ \midrule
     Transformer-XL & ... under the 1906 – \textbf{07 shipbuilding} program . \\
     & ... the same \textbf{raw warmth} that was present in other ...\\
     &  It is assumed \textbf{that Suetonius} used information gathered ... \\ \midrule \midrule
     GPT-2 & ... \textbf{Scope Names} with Closures ... \\
     &  ... bidding \$ 6 against the \textbf{AppNexus SSP} and just ...  \\
     & ... a space ( e.g. , a \textbf{ceiling overhang} ) ... \\
     \bottomrule
    \end{tabular}
    \caption{Randomly-selected examples of novel bigrams generated by models. Most of the novel bigrams are used well in context.}
    \label{fig:novel_bigrams}
\end{figure*}

\section{Supercopying examples and statistics}\label{app:supercopying}

\begin{figure}[t]
    \centering
    \includegraphics[width=0.9\columnwidth]{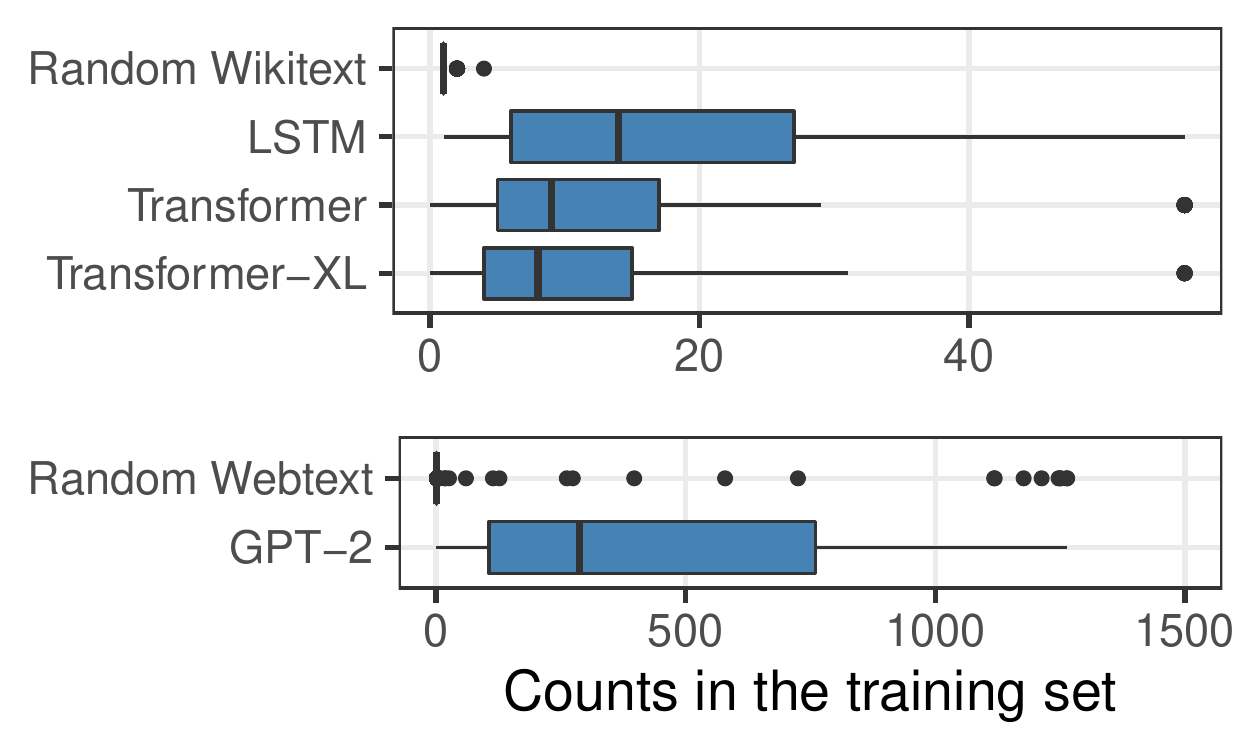}
    \caption{Counts of how often 100-grams supercopied by each model appear in that model's training set, compared to counts of random 100-grams from the training sets. For legibility, some GPT-2 outliers have been removed. The biggest outlier was a supercopied passage that occurred 176,424 times in GPT-2's training set.}
    \label{fig:supercopycounts}
\end{figure}

Figure \ref{fig:supercopy_examples} gives examples of generated text, as well as text from our models' test sets, that we classify as supercopying: duplicating a passage that is 100 words long or longer from the training set. Figure \ref{fig:supercopycounts} gives statistics about how many times supercopied 100-grams appeared in the relevant model's training set.

\begin{figure*}
    \centering
    \begin{tabular}{p{2cm}p{13cm}} \toprule
     Generation method & Example of supercopying  \\ \midrule
     Wikitext test set & <eos> <eos> = = Themes = = <eos> <eos> The Hustler is fundamentally a story of what it means to be a human being , couched within the context of winning and losing . Describing the film , Robert Rossen said : " My protagonist , Fast Eddie , wants to become a great pool player , but the film is really about the obstacles he encounters in attempting to fulfill himself as a human being . He attains self @-@ awareness only after a terrible personal tragedy which he has caused — and then he wins his pool game . " Roger Ebert concurs with this assessment \\
     LSTM, Transformer, Transformer-XL & . <eos> <eos> = = Background = = <eos> <eos> The Boat Race is a side @-@ by @-@ side rowing competition between the University of Oxford ( sometimes referred to as the " Dark Blues " ) and the University of Cambridge ( sometimes referred to as the " Light Blues " ) . The race was first held in 1829 , and since 1845 has taken place on the 4 @.@ 2 @-@ mile ( 6 @.@ 8 km ) Championship Course on the River Thames in southwest London . The rivalry is a major point of honour \\ \midrule
     WebText test set & All crafted armor is Adaptive . NEWLINE NEWLINE Nearly all schematics require a new Premium ( green ) crafting material called a Component . This new Component is made by combining two existing crafting materials along with two vendor purchased materials , all of which are of the same Grade as the component . A unique Component exists for each skill and each Grade . Example : An Armormech crafter wants to craft a Grade 6 Chest Piece ( Level 46 ) . This is a Prototype ( blue ) quality armor and requires 4 Ciridium and 4 Durasteel Armor Assembly Components . The player will need 2 Durasteel , 2 Zal Alloy , and 2 Thermoplast Flux to craft each component . Once they have the 4 Components and 4 Ciridum from Underworld Trading , they will be able to craft their chest piece . If a player chose to , they could level up exclusively by crafting Components ! NEWLINE NEWLINE In most cases , Crafting times have been greatly reduced . This will enable crafters to craft more items to coincide with the faster leveling .  \\
     GPT-2 & your bank account . NEWLINE NEWLINE Your billing zip code needs to be 5 digits . NEWLINE NEWLINE Please double check your CEP info . The CEP format should be something like 12345-678 . NEWLINE NEWLINE Please double check your tax identifier . NEWLINE NEWLINE There was a problem saving your address . NEWLINE NEWLINE There was a problem saving your card info . NEWLINE NEWLINE There was a problem saving your personal information . NEWLINE NEWLINE McAfee Secure sites help keep you safe from identity theft , card fraud , spyware , spam , viruses and online scams . NEWLINE NEWLINE Copying Prohibited by Law - McAfee Secure is a Trademark of McAfee , Inc . NEWLINE NEWLINE Unknown card type . NEWLINE NEWLINE No card number provided . NEWLINE NEWLINE card number is in invalid format . NEWLINE NEWLINE Wrong card type or card number is invalid . NEWLINE NEWLINE card number has an inappropriate number of digits . NEWLINE NEWLINE Please enter numbers here . NEWLINE NEWLINE Please enter an integer value . NEWLINE NEWLINE Numbers must be less or equal to \$ \$ \$ \$ NEWLINE NEWLINE All the required fields have not been filled out . Click OK to proceed without all the required information , or click Cancel to finish entering the missing data .   \\ \bottomrule
    \end{tabular}
    \caption{Examples of supercopying. The passage in the second row appeared in the generated text for all 3 of our Wikitext models (LSTM, Transformer, and Transformer-XL).}
    \label{fig:supercopy_examples}
\end{figure*}

\section{Untruncated pointwise duplication scores}\label{app:decoding_graphs}

\begin{figure}[t]
    \centering
    \includegraphics[width=\columnwidth]{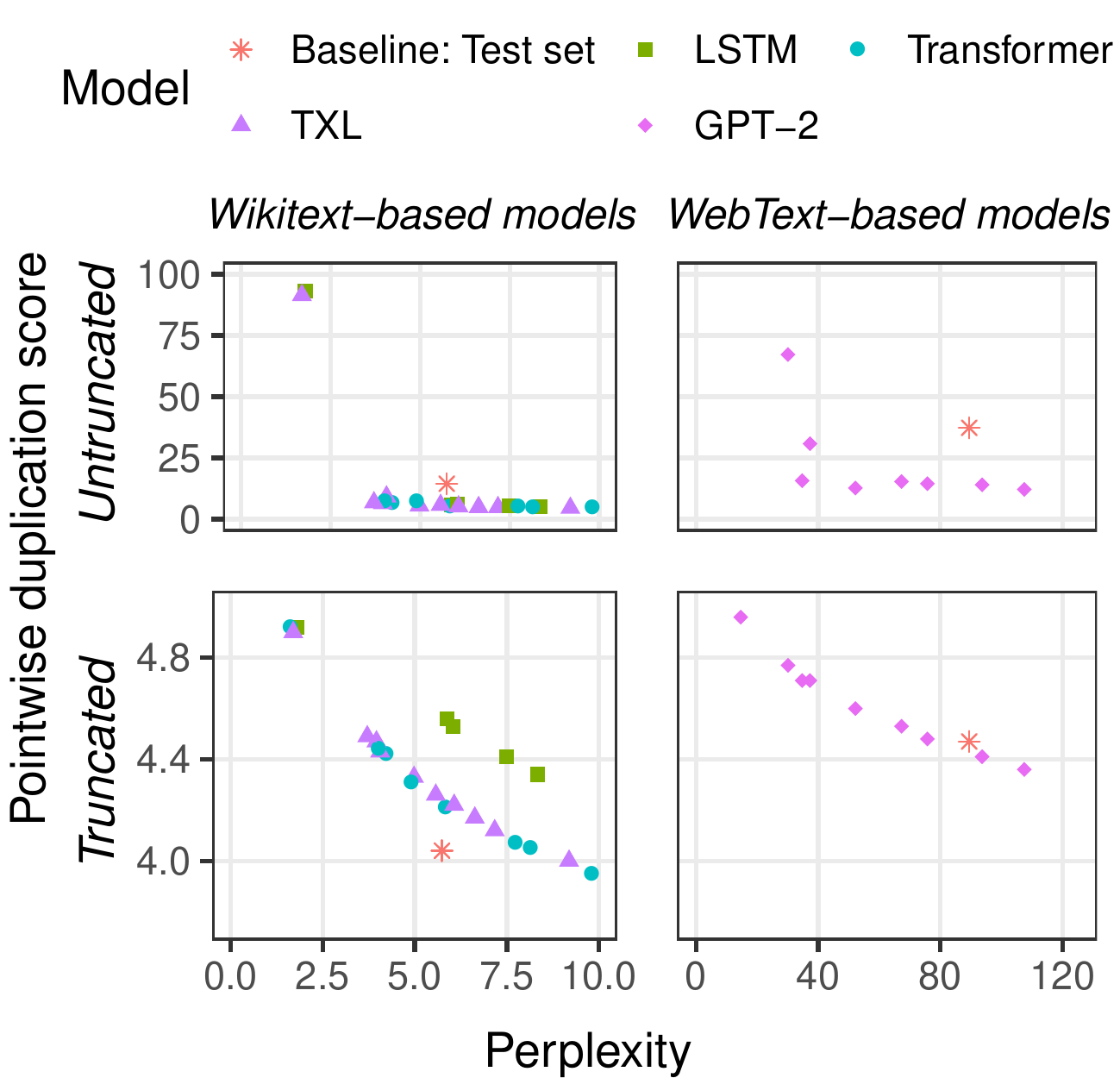}
    \caption{Manipulations to the decoding scheme that result in higher-quality text (i.e., lower perplexity; $x$-axis) also result in decreased novelty (i.e., a greater degree of duplication; $y$-axis). Each point shows a different decoding scheme.}
    \label{fig:quality_app}
\end{figure}

Figure \ref{fig:quality_app} shows the same results as Figure \ref{fig:quality} but with both the truncated and untruncated pointwise duplication scores (whereas Figure \ref{fig:quality} only shows the truncated scores).

\section{Plots showing the effects of the decoding scheme}\label{app:decoding}

\begin{figure}[t]
    \centering
    \includegraphics[width=\columnwidth]{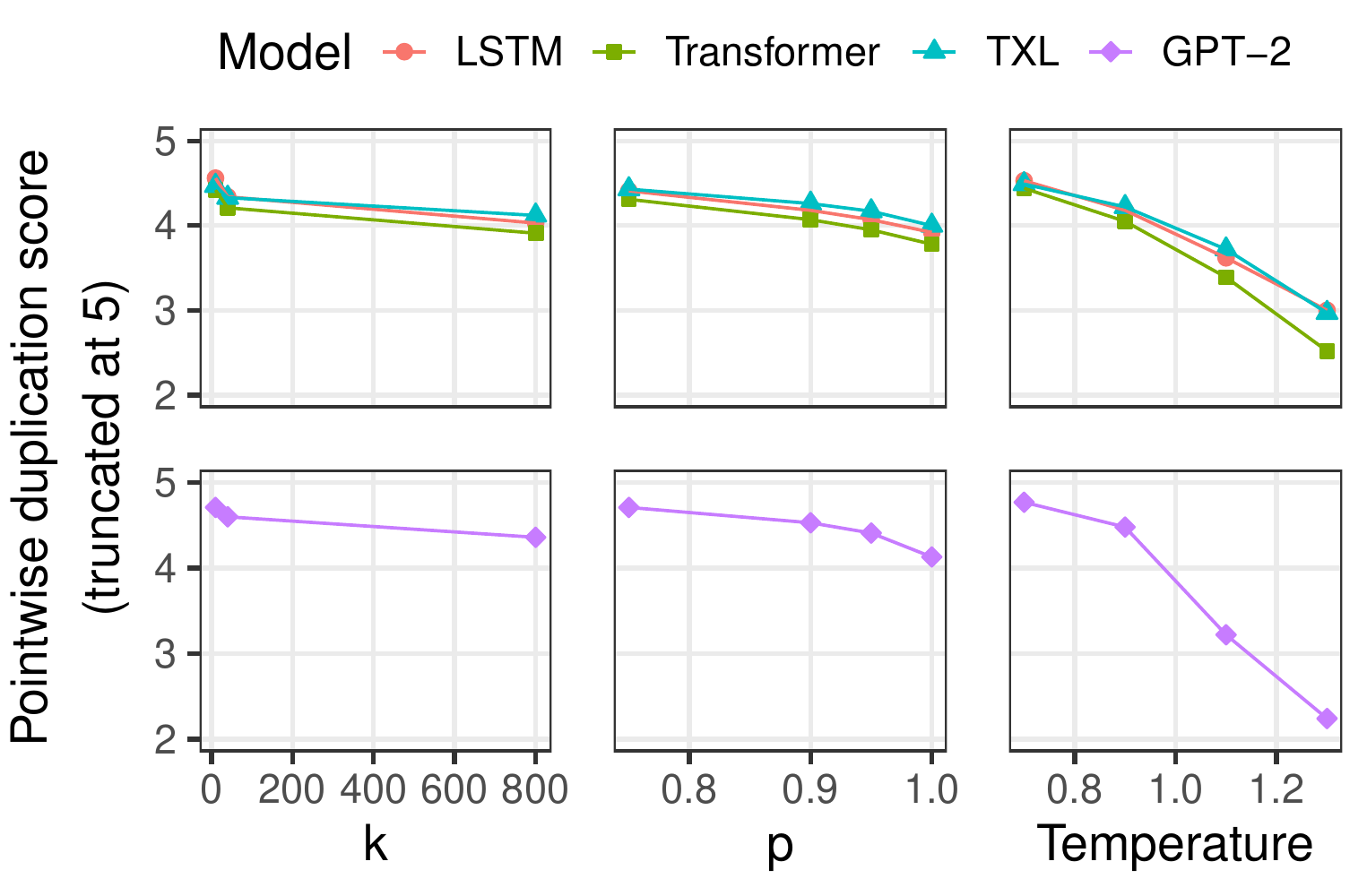}
    \caption{How novelty is affected by 3 decoding parameters: $k$ in top-$k$ sampling, $p$ in top-$p$ sampling, and the temperature. A higher value on the $y$ axis means that the generated text is less novel. As each parameter is increased, novelty also increases (that is, duplication---on the $y$-axis---decreases).}
    \label{fig:decoding}
\end{figure}

Figure \ref{fig:decoding} shows how the pointwise duplication score is affected by varying three decoding parameters.

\section{Evaluating perplexity}\label{app:perplexity}

\subsection{Evaluating overlap between Wikitext-103 and WebText}\label{sec:evaluating_overlap_wikitext_webtext}

To measure the perplexity of generated text, we used GPT-2 (which was trained on WebText) to measure perplexity for models trained on Wikitext-103, and we used TXL (which was trained on Wikitext-103) to measure perplexity for models trained on WebText. We justified this choice based on the fact that Wikitext-103 was constructed entirely from Wikipedia articles, while Wikipedia articles were excluded from WebText, meaning that there should be no overlap between the training sets of Wikitext-trained models and WebText-trained models. However, there is a caveat for making this assumption: the WebText creation process excluded Wikipedia articles, but the \textit{text from} Wikipedia articles could still potentially occur in WebText, because there are many non-Wikipedia websites that copy data from Wikipedia; and because Wikipedia writers could potentially generate Wikipedia content by copying it from other public-domain websites.

Here we test our no-overlap assumption more rigorously. To do so, we randomly selected one thousand 20-grams from the WebText training set and checked whether they appeared in the Wikitext-103 training set. Similarly, we also selected one thousand 20-grams from the Wikitext-103 training set and checked whether they appeared in the WebText training set tokenized with the Moses tokenizer \cite{koehn2007moses}, which was the basis of the tokenization in Wikitext-103. To control for tokenization differences that might persist despite the use of the Moses tokenizer, we lowercased all text and deleted any words containing any characters besides the 26 Roman letters. 

We found that 0 of the 1000 WebText 20-grams appeared in the Wikitext-103 training set, so  it seems very safe to use TXL to evaluate the perplexity of text generated by models trained on WebText. In the other direction, 12 of the 1000 Wikitext 20-grams appeared in the WebText training set. This shows that a small amount of text that appears in Wikipedia text did end up in WebText. Nonetheless, the proportion of overlap is very small (0.012), so we conclude that it is still fair to use GPT-2 to evaluate the perplexity of text generated by models trained on Wikitext-103.

\subsection{Details of perplexity evaluation}

To evaluate the perplexity of a piece of text using Transformer-XL or GPT-2, we adapted code from Hugging Face at \url{https://huggingface.co/transformers/perplexity.html}. For each model, we used a stride of 512 tokens and a maximum length of 1024 tokens. That is, the perplexity was evaluated in segments of 1024 tokens each, with each segment preceded by a further context of 512 words whose perplexity was not evaluated as part of the segment being evaluated. This approach ensures that every token has at least 512 tokens of prior context available; tokens at the end of a 1024-token segment have an even longer context (specifically, a context of 1535 tokens for the last token in each segment).

\section{How does model size affect novelty?}\label{app:size}

It seems possible for model size to affect novelty in either direction: It could be that larger models have a greater capacity to memorize, and would therefore be less novel. On the other hand, larger models are generally stronger \cite{kaplan2020scaling}, which might include greater strength in their ability to be novel.

\begin{figure}
    \centering
    \includegraphics[width=\columnwidth]{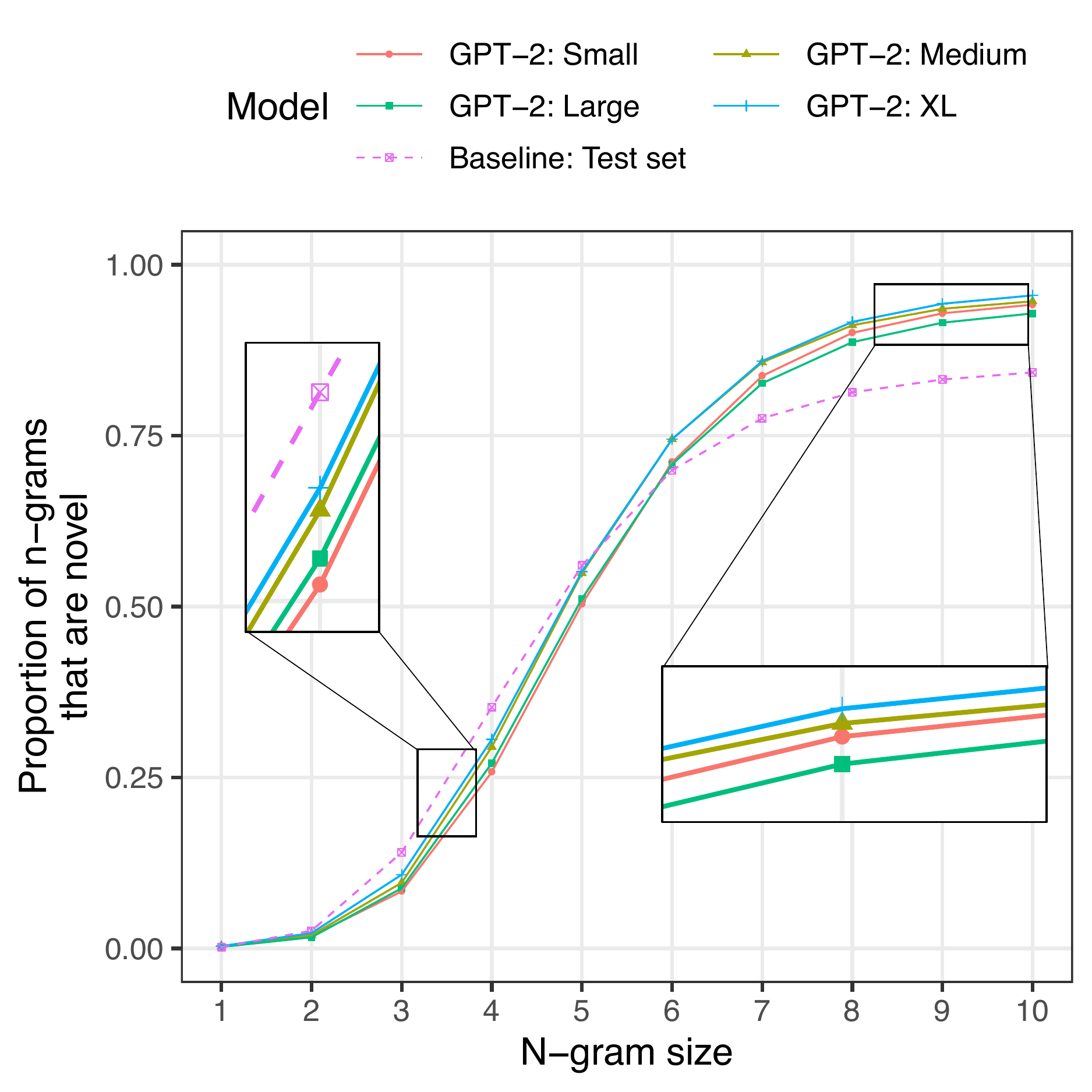}
    \caption{Effect of model size.}
    \label{fig:gpt2size}
\end{figure}

Figure \ref{fig:gpt2size} shows the level of duplication observed for the 4 different sizes of GPT-2 (all using top-40 samping). There is not a clear, consistent effect of size. Across the various $n$-gram sizes, the most novel model is GPT-2 XL; however, GPT-2 Medium is more novel than GPT-2 Large, so it is not the case that larger models are always more novel than smaller models (or vice versa).

\section{How does prompt length affect novelty?}\label{app:prompt_length}

\begin{figure*}
    \centering
    \begin{subfigure}{0.48\textwidth}
        \centering
        \includegraphics[width=\textwidth]{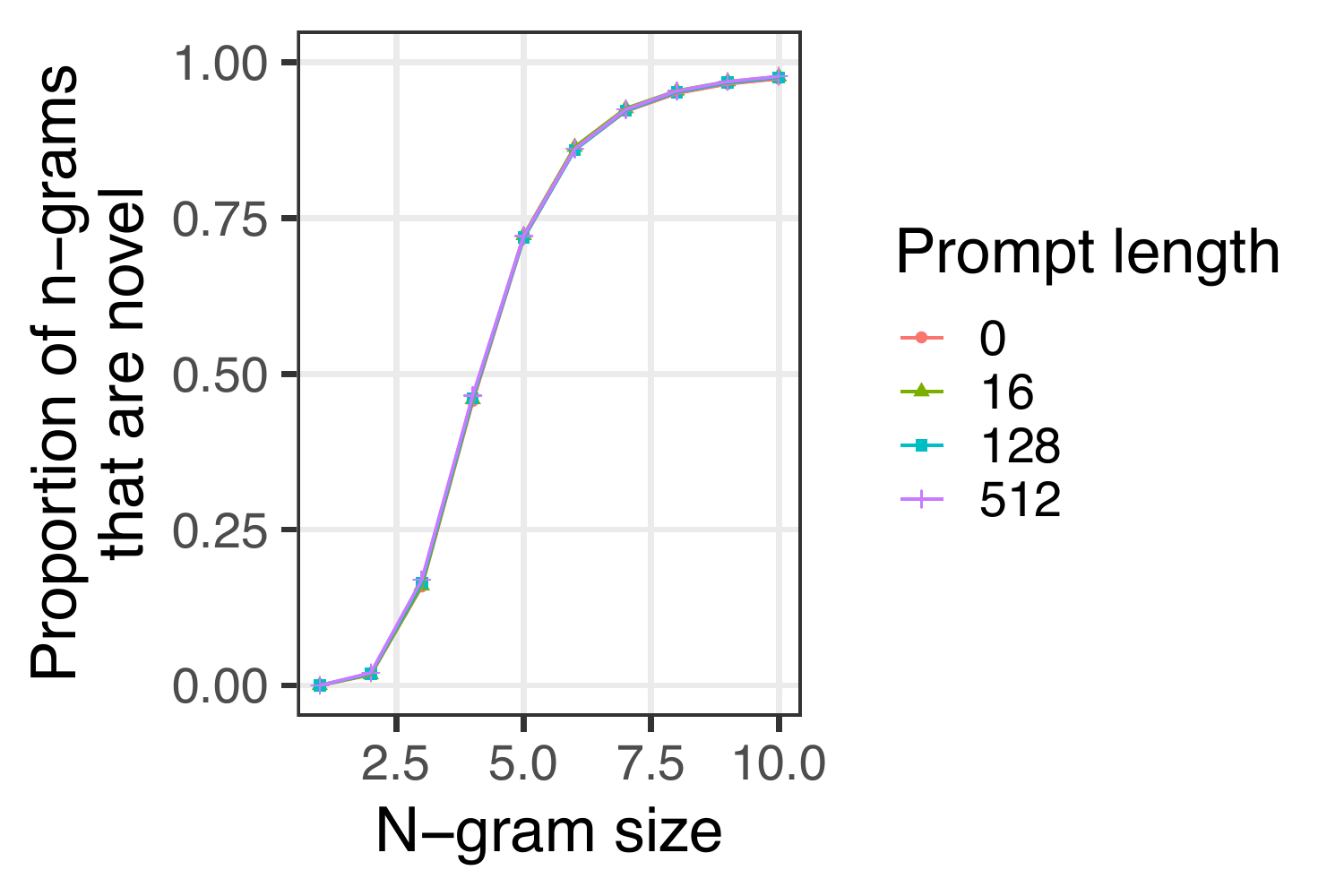}
        \subcaption{LSTM}\label{fig:promptlength_lstm}
    \end{subfigure}%
    \begin{subfigure}{0.48\textwidth}
        \centering
        \includegraphics[width=\textwidth]{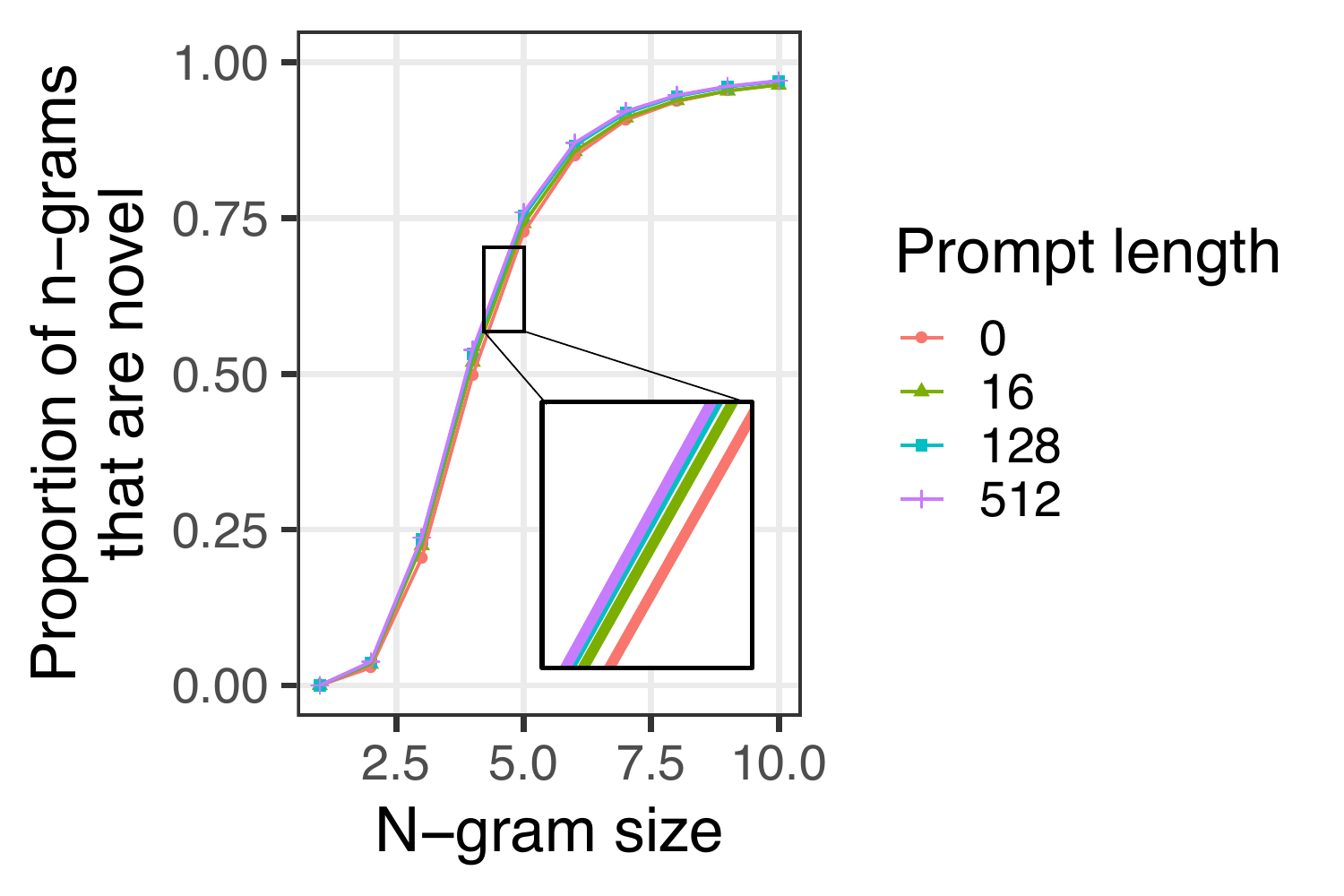}
        \subcaption{Transformer}\label{fig:promptlength_transformer}
    \end{subfigure}%
    
    \begin{subfigure}{0.48\textwidth}
        \centering
        \includegraphics[width=\textwidth]{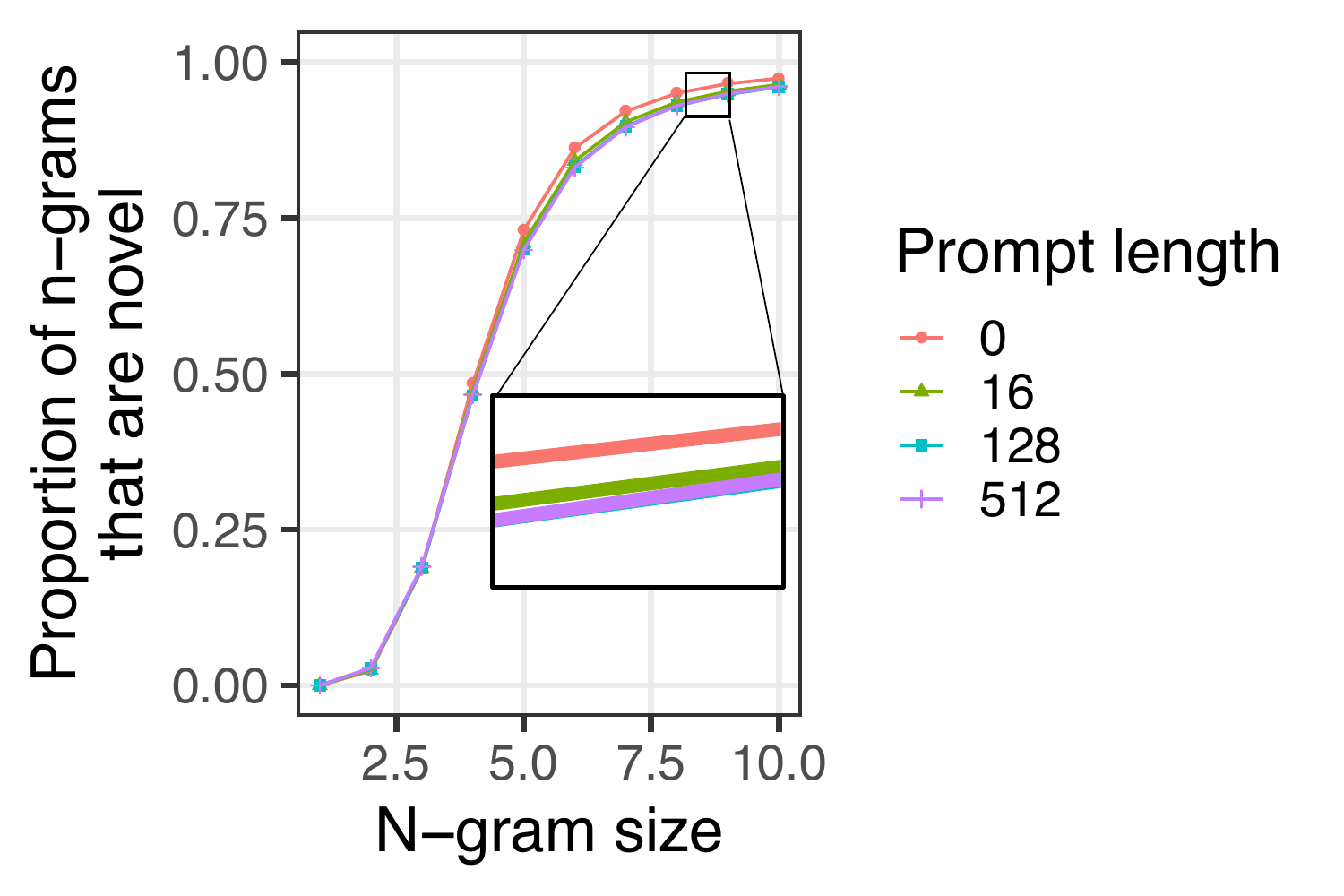}
        \subcaption{TXL}\label{fig:promptlength_txl}
    \end{subfigure}%
    \begin{subfigure}{0.48\textwidth}
        \centering
        \includegraphics[width=\textwidth]{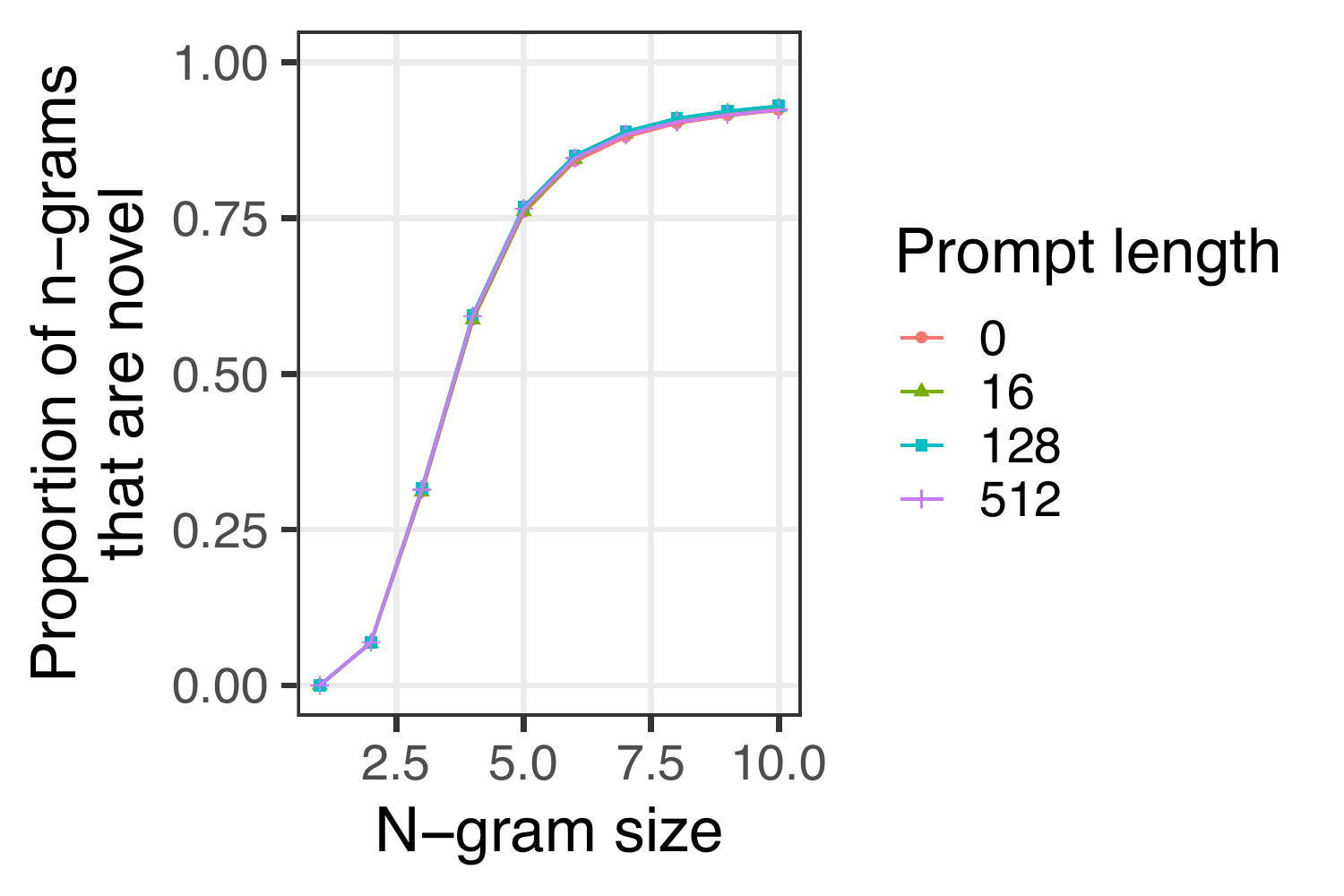}
        \subcaption{Wikitext baseline text}\label{fig:promptlength_wikitext}
    \end{subfigure}%
    
    \begin{subfigure}{0.48\textwidth}
        \centering
        \includegraphics[width=\textwidth]{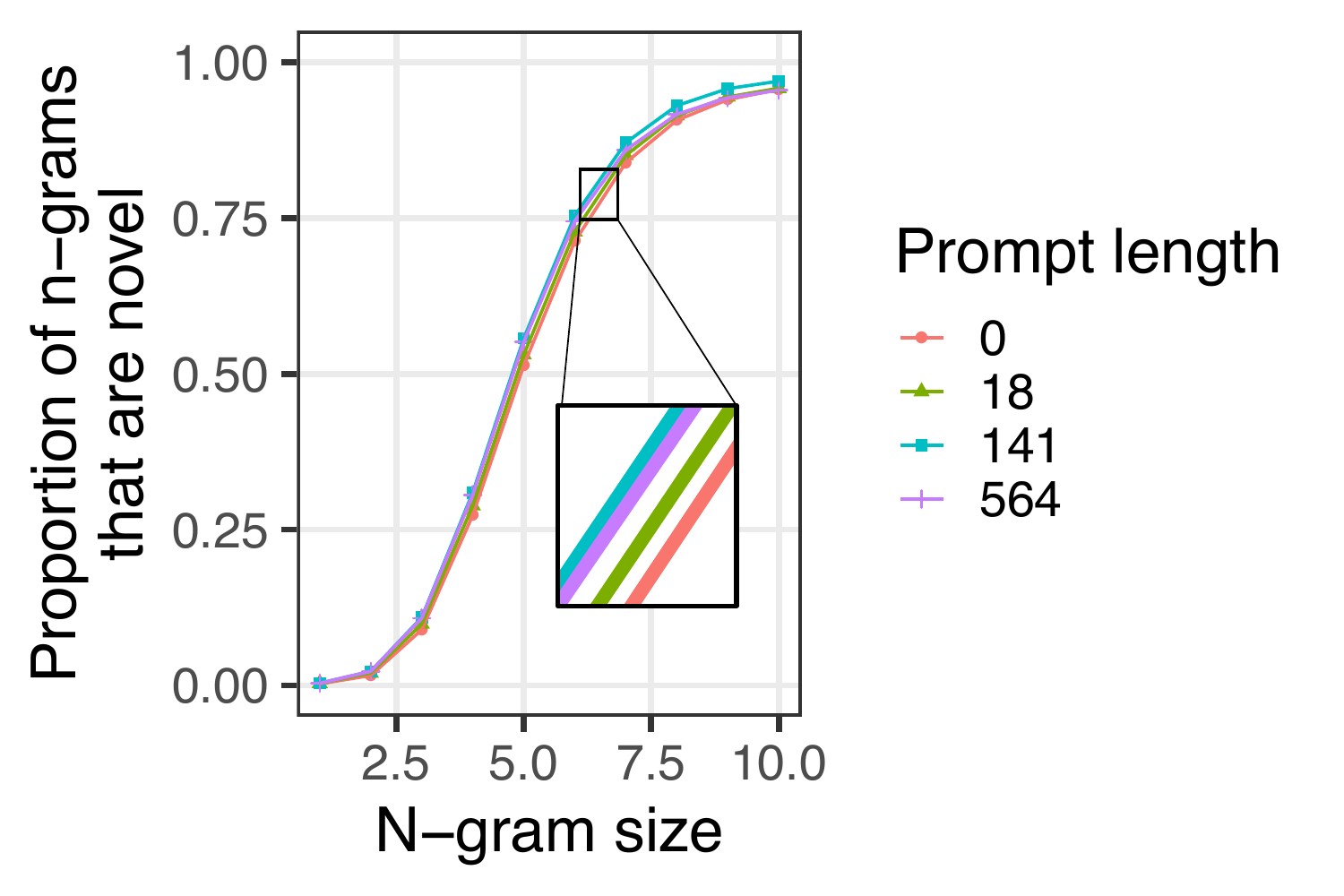}
        \subcaption{GPT-2}\label{fig:promptlength_gpt2}
    \end{subfigure}%
    \begin{subfigure}{0.48\textwidth}
        \centering
        \includegraphics[width=\textwidth]{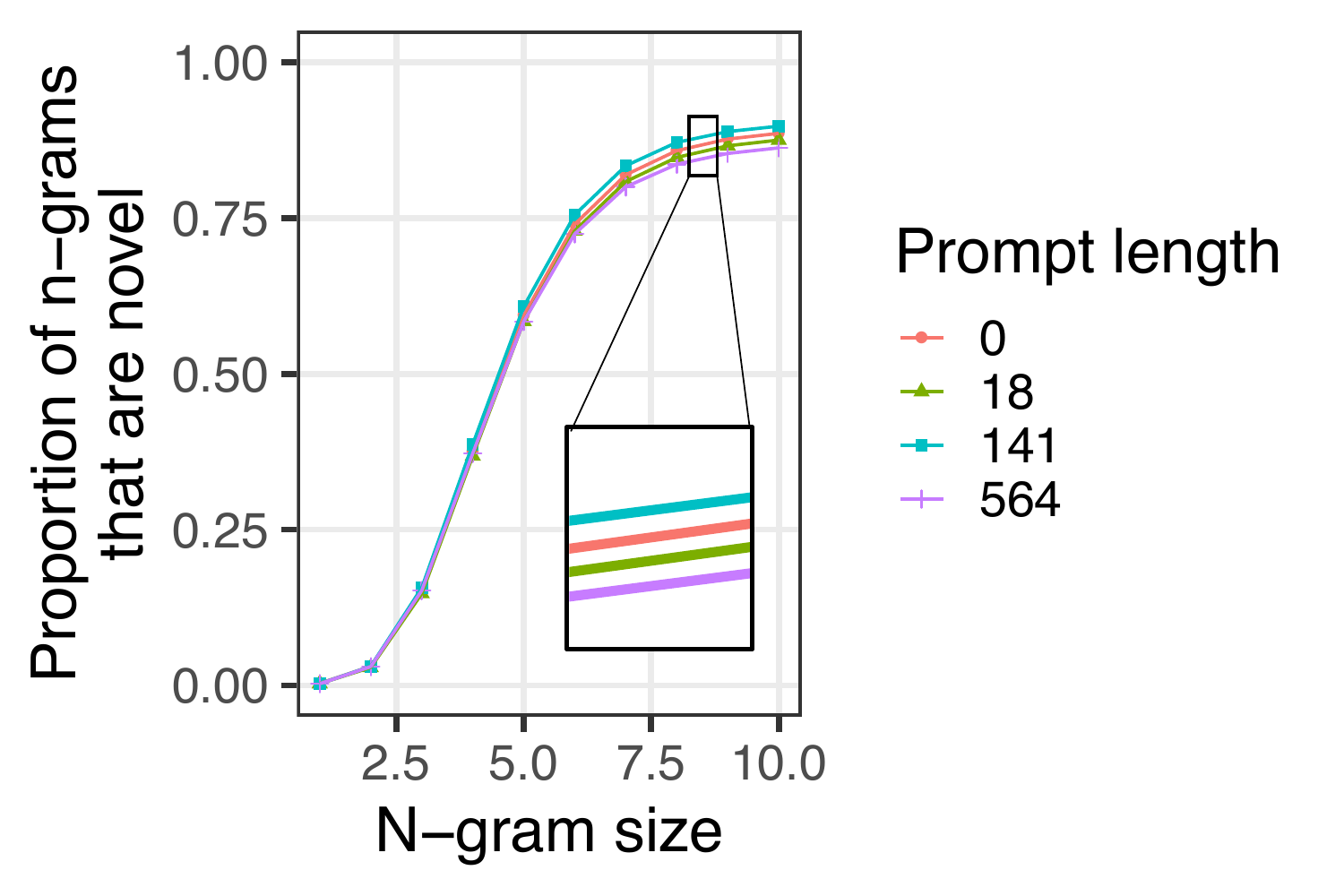}
        \subcaption{Webtext baseline text}\label{fig:promptlength_webtext}
    \end{subfigure}%
    \caption{Effect of prompt length.}
    \label{fig:promptlength}
\end{figure*}

Prompt length could reasonably be expected to increase or decrease novelty. On one hand, shorter prompts might not give the model much context to build from, which could lead the model to fall back on what it has seen during training, making it less novel. On the other hand, we observe from the baselines in Figure \ref{fig:ngram_overlap} that there is a reasonably high overlap between models' training sets and their test sets. Since our prompts are drawn from the test set, a long prompt might contain long portions that also appear from the training set, which could encourage the model to further copy from that part of the training set, in which case longer prompts would lead to lower novelty than shorter prompts. 

To assess how novelty is affected by prompt length, we consider only duplication from the training set, not from the context, because a longer prompt trivially provides more opportunities for copying from the context.
In general, the length of the prompt does not appear to affect novelty much (Figure \ref{fig:promptlength}). For the LSTM and the baseline of text drawn from the Wikitext-103 test set, we do not discern any effect of prompt length. For the Transformer and GPT-2, longer prompts lead to slightly more novelty than shorter prompts, while Transformer-XL shows the opposite effect, with shorter prompts leading to slightly more novelty than longer prompts. The WebText baseline shows some differences between prompt lengths, but we do not see a clear generalization there as novelty is not affected consistently by length: for example, length 0 is more novel than length 18 but less novel than length 141.

\section{Position in generated text}\label{app:position}

\begin{figure}
    \centering
    \begin{subfigure}{0.48\textwidth}
        \centering
        \includegraphics[width=\textwidth]{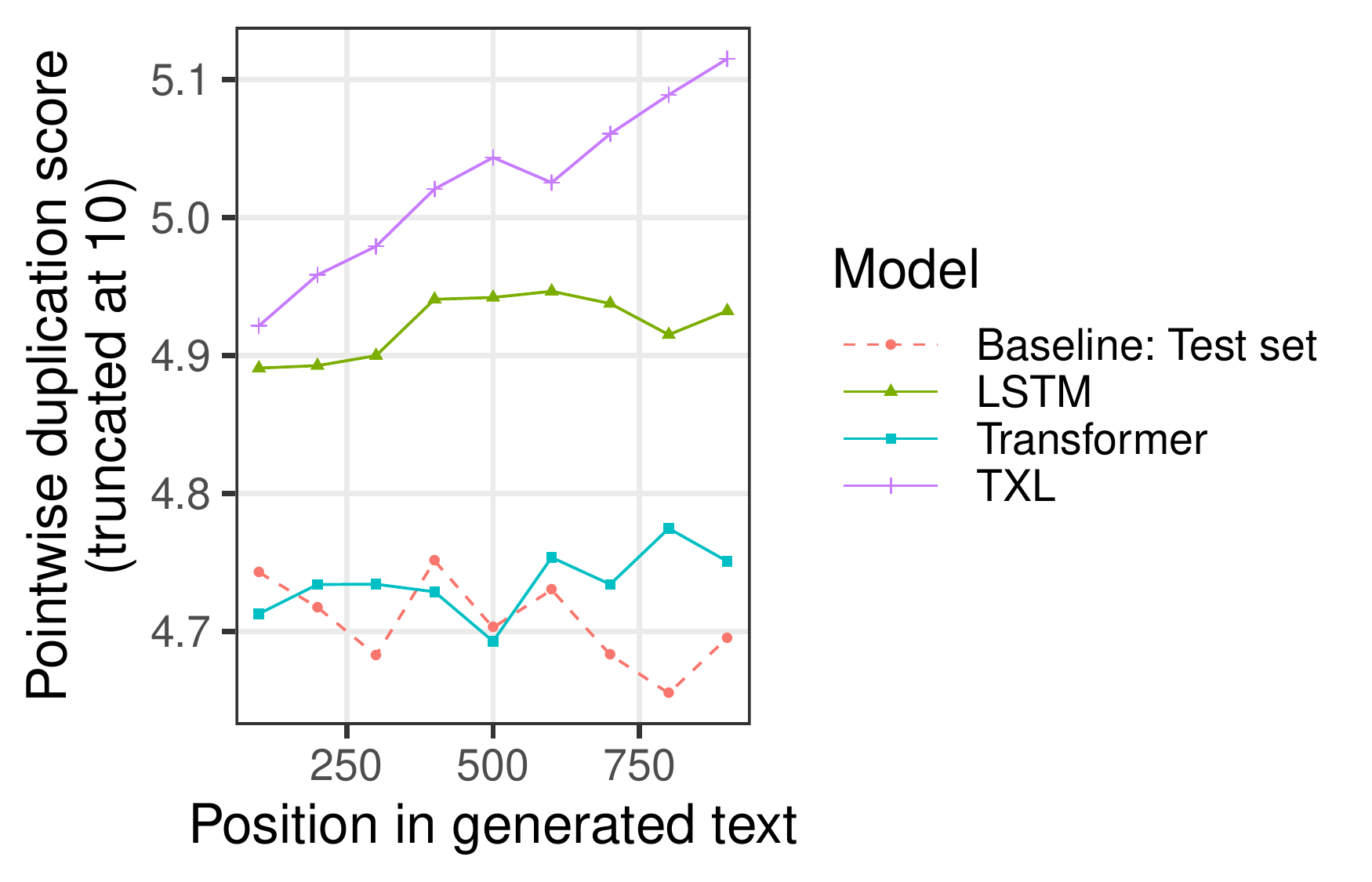}
        \subcaption{Wikitext-based models.}\label{fig:wiki_position}
    \end{subfigure}%
    
    \begin{subfigure}{0.48\textwidth}
        \centering
        \includegraphics[width=\textwidth]{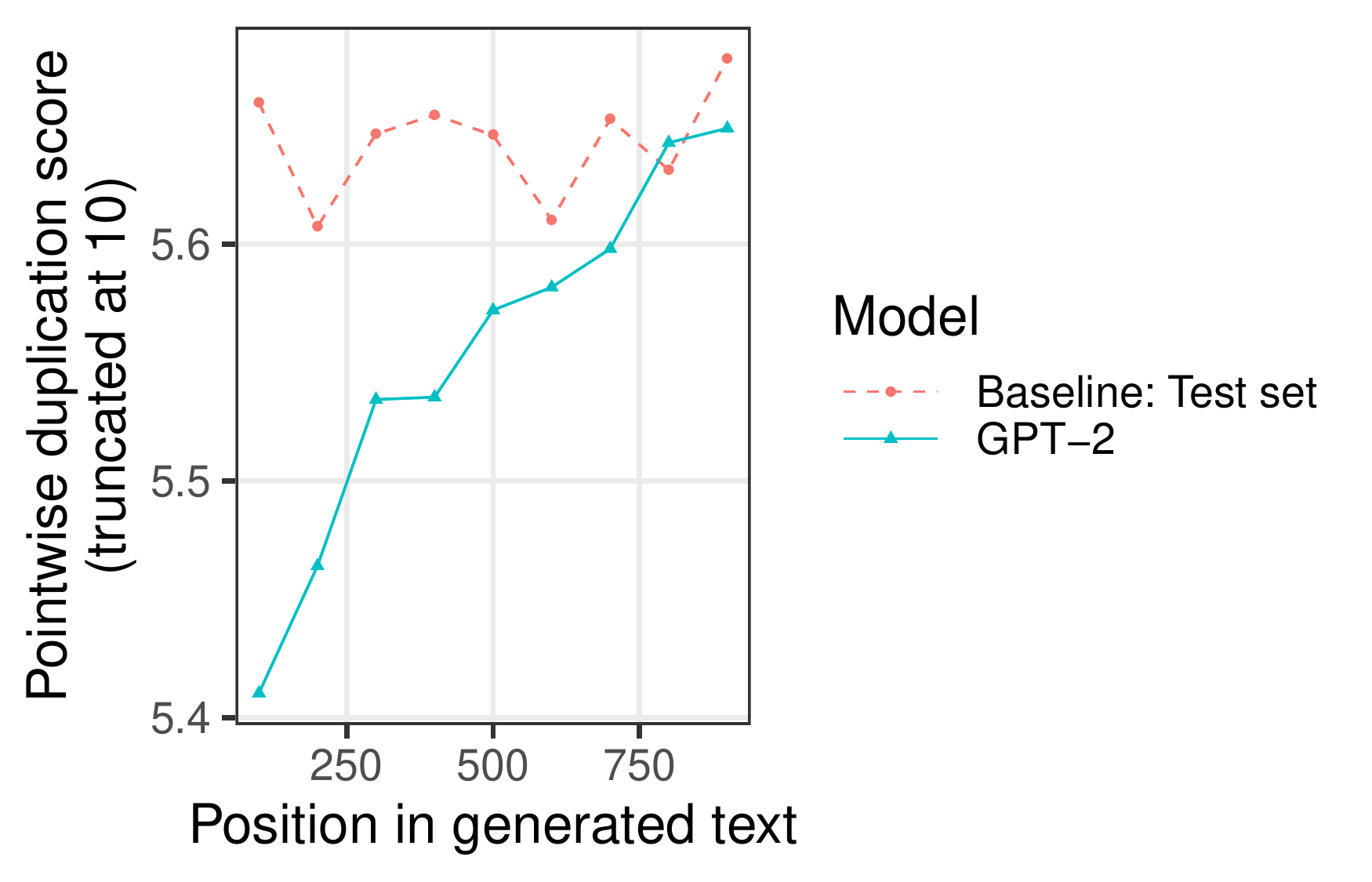}
        \subcaption{WebText-based models.}\label{fig:web_position}
    \end{subfigure}%
    
    \caption{Effect of output position on novelty.}
    \label{fig:position}
\end{figure}

Figure \ref{fig:position} illustrates how novelty is related to position in the output text. For these analyses, we only consider duplication from the training set, not from the context, because positions later in the generation have more context to copy from. We use pointwise duplication scores truncated at 10 to control for the fact that later positions can have higher untruncated pointwise scores than earlier positions can have; by using truncated scores, all positions that we consider have the same possible range of values (1 to 10 inclusive). We group generated text into bins of 100 words (positions 0 to 99, positions 100 to 199, etc.) and then compute the mean truncated pointwise duplication score for each bin, discarding the first bin because its first 10 positions have a different range of possible scores than the rest of the positions in the generation. 

There is little effect of position in the baselines, the LSTM, and the Transformer, but in GPT-2 and Transformer-XL, there is greater duplication at later positions in the generated text. Though the effect is consistent for these two models, the effect size is small, with the pointwise duplication score increasing by only about 0.2 from the start of the generation to the end of the generation.

\section{Duplication from the training set vs. the context}\label{app:duplication_training_context}

\begin{figure*}
    \centering
    \begin{subfigure}{0.45\textwidth}
        \centering
        \includegraphics[width=\textwidth]{figures/nlg_novelty_wiki_dashed.pdf}
        \subcaption{Wikitext-based models: Duplication from the training set and/or context.}\label{fig:wiki_ngram_both}
    \end{subfigure}%
    \hfill
    \begin{subfigure}{0.45\textwidth}
        \centering
        \includegraphics[width=\textwidth]{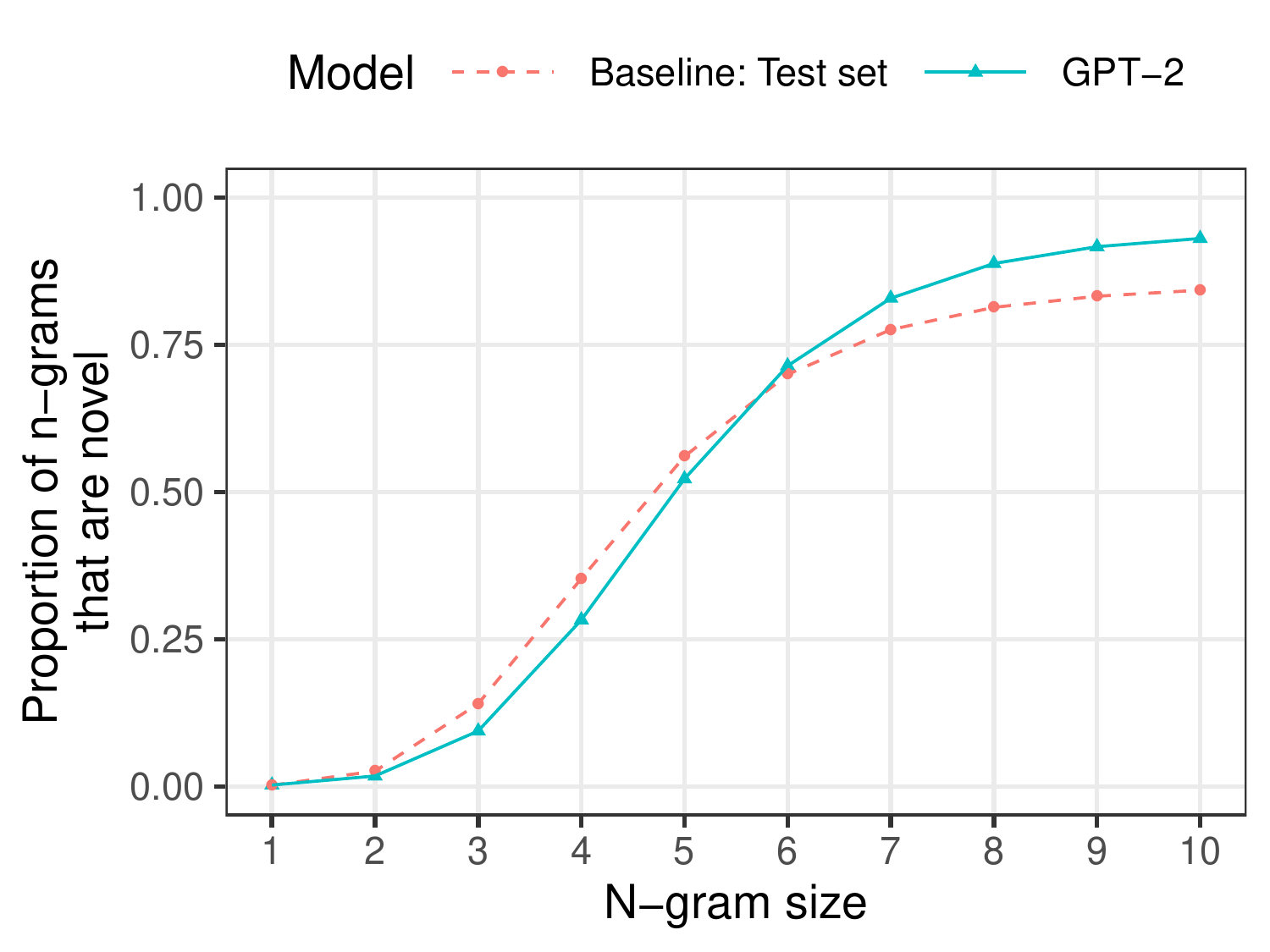}
        \subcaption{WebText-based models: Duplication from the training set and/or context.}\label{fig:web_ngram_both}
    \end{subfigure}%
    
    \begin{subfigure}{0.45\textwidth}
        \centering
        \includegraphics[width=\textwidth]{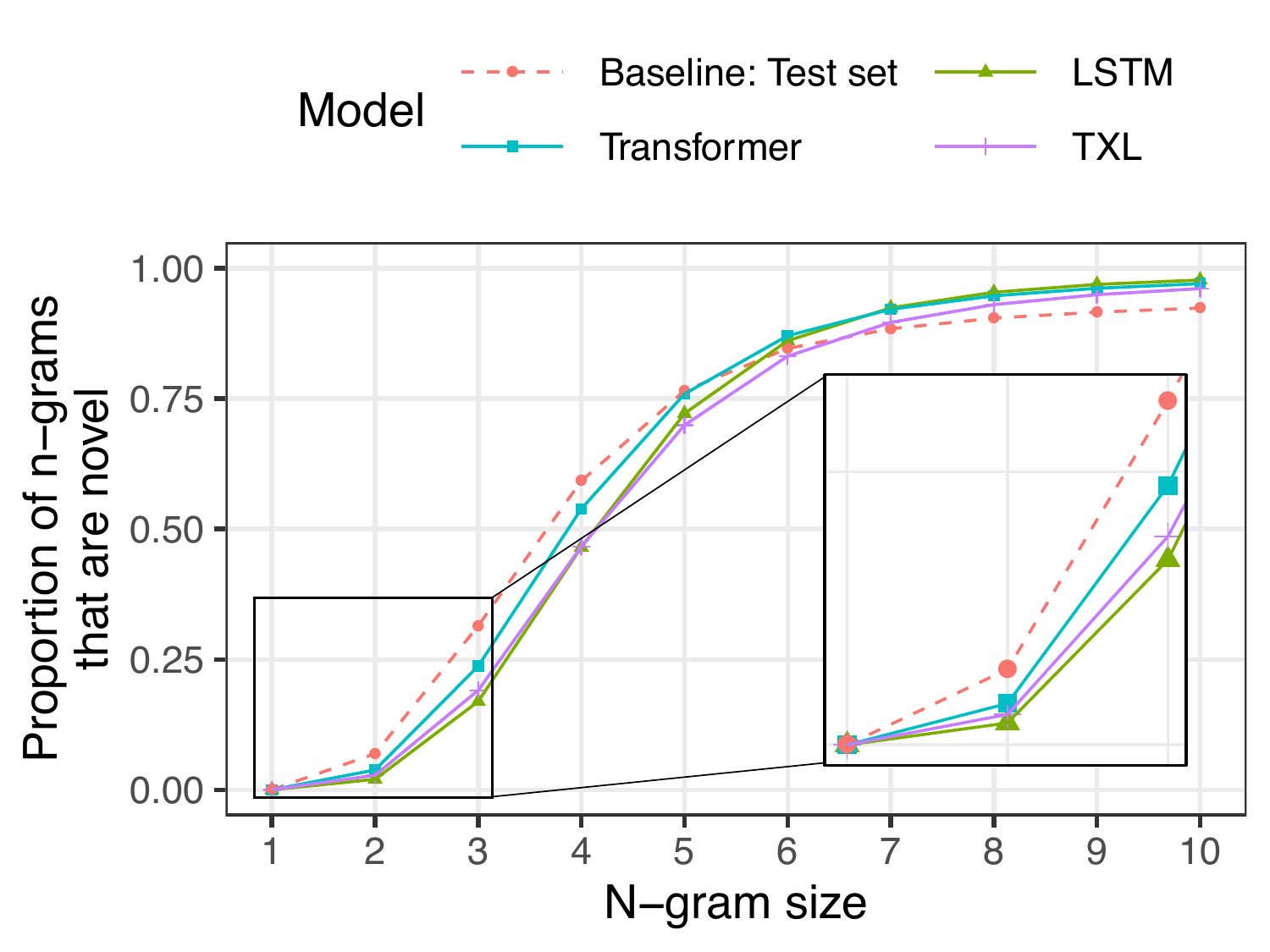}
        \subcaption{Wikitext-based models: Duplication from the training set.}\label{fig:wiki_ngram_training}
    \end{subfigure} %
    \hfill
    \begin{subfigure}{0.45\textwidth}
        \centering
        \includegraphics[width=\textwidth]{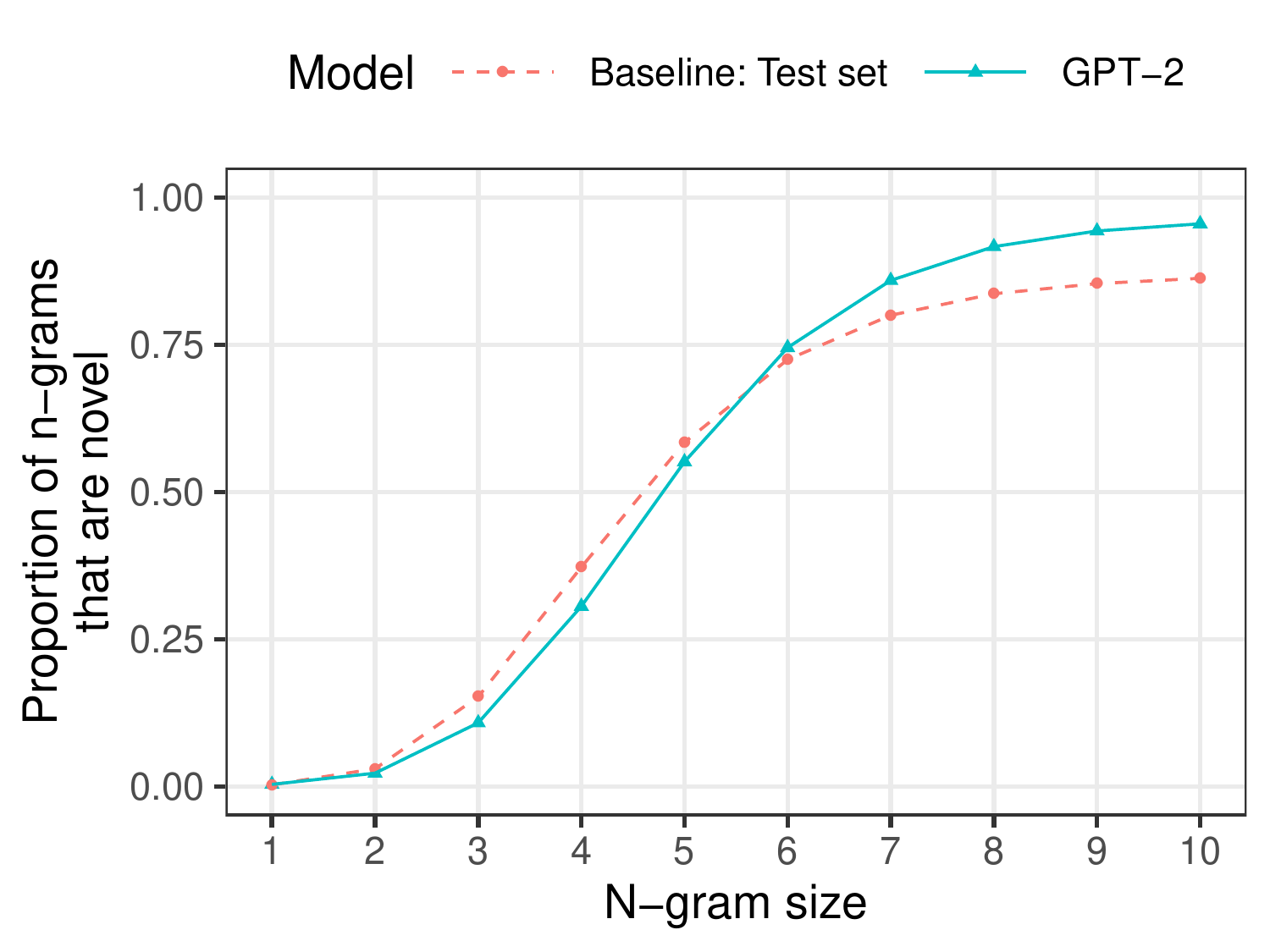}
        \subcaption{WebText-based models: Duplication from the training set.}\label{fig:web_ngram_training}
    \end{subfigure}%
    
    \begin{subfigure}{0.45\textwidth}
        \centering
        \includegraphics[width=\textwidth]{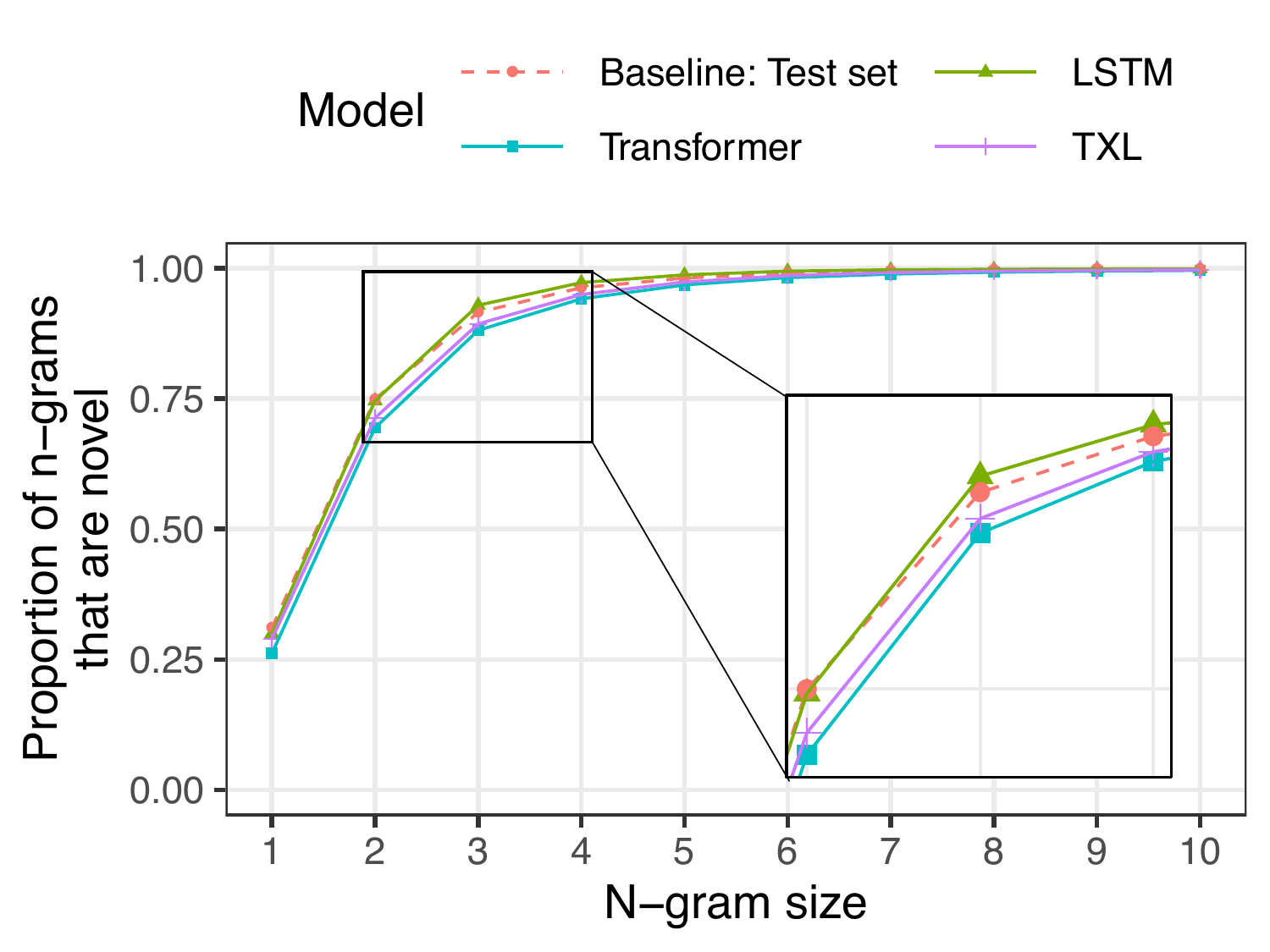}
        \subcaption{Wikitext-based models: Duplication from the context.}\label{fig:wiki_ngram_context}
    \end{subfigure}%
    \hfill
    \begin{subfigure}{0.45\textwidth}
        \centering
        \includegraphics[width=\textwidth]{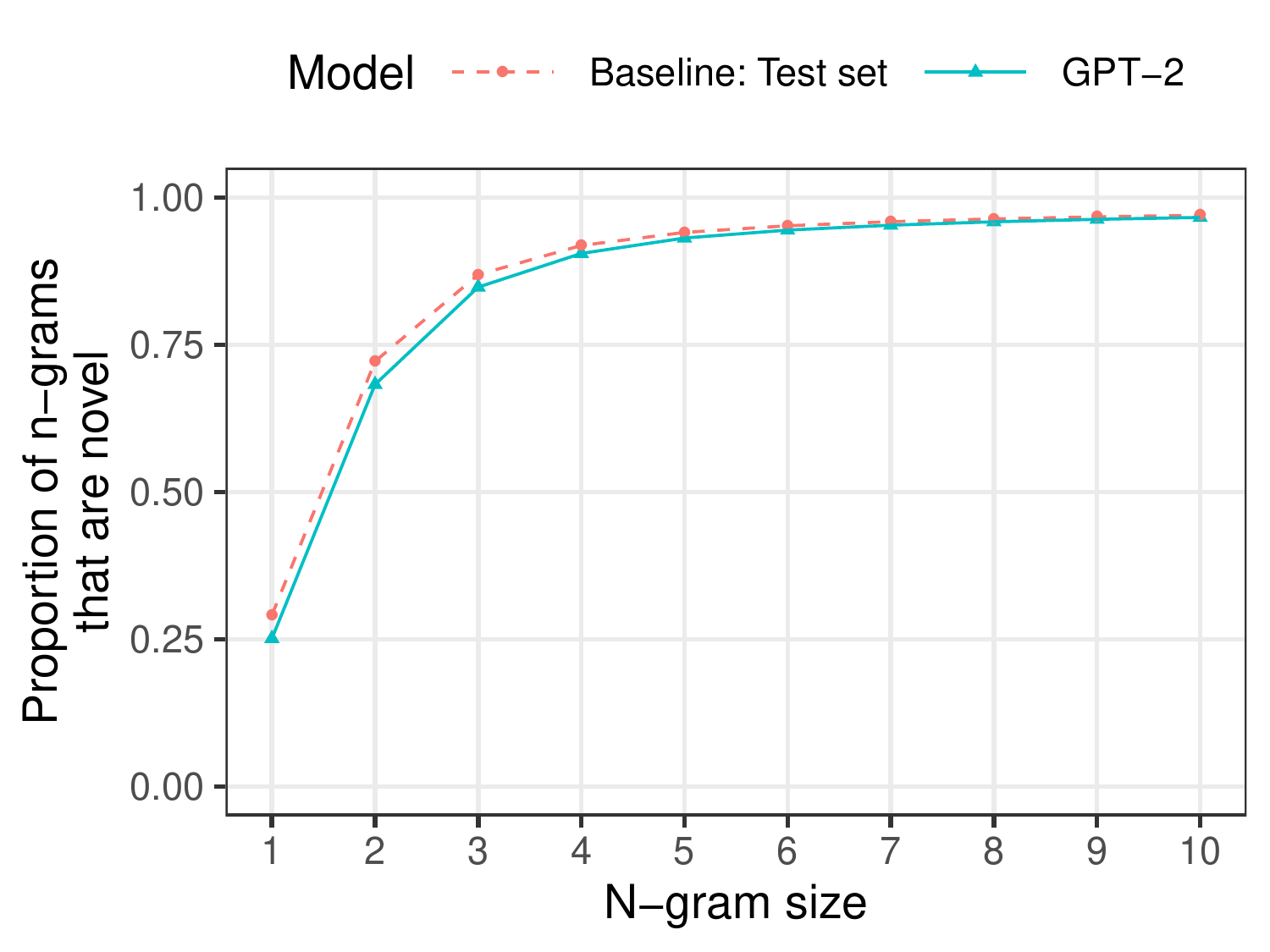}
        \subcaption{WebText-based models: Duplication from the context.}\label{fig:web_ngram_context}
    \end{subfigure}%
    
    \caption{Duplication from the training set, context, or both.}
    \label{fig:ngram_breakdown}
\end{figure*}

In the main text, we only report results that collapse together duplication from the training set and duplication from the context (the prompt and the previously-generated text). In Figure \ref{fig:ngram_breakdown}, we separate these two sources of duplication. The plots showing duplication from the training set alone are almost identical to the plots showing both the training set and the context, showing that models almost never copied content from the context that was not also in the training set. Models showed much less duplication from the context than from the training set, which is unsurprising because the training sets are much larger than the contexts, meaning that there are far more pieces of text that would count as duplicated from the training set than duplicated from the context.

\section{Vetting syntax}\label{app:syntax_vetting}

We considered evaluating the novelty of 7 aspects of syntax. For each of these 7 aspects, we conducted manual analyses to determine whether the numerical results gained from our parses were roughly accurate. Below we describe these manual analyses. 

\paragraph{POS tags:} We considered a novel part-of-speech (POS) tag to be an instance where a word was generated with a POS tag that it had never appeared with in training (but where the word had appeared in training with a different POS tag). From an initial inspection of the generated words that the parser identified as having a novel part of speech, we found that most of them were correctly labeled, but that the training set actually did contain an instance of the word with that part of speech (just with that training instance mislabeled). Thus, we concluded that we could not trust the quantitative results for this factor, because most generated words identified as having a novel POS tag had actually appeared as that POS in training.

\paragraph{CFG rules:} The CFG rules are the context-free rules present in the parses from the constituency parser (ignoring rules that include a terminal symbol---i.e., only considering ones composed entirely of nonterminals). From an initial inspection of the CFG rules identified as novel, every example that we looked at was the result of a parser error (where the parser assigned an overly flat rule, e.g. \textit{NP $\rightarrow$ -LRB- CD -RRB- , NP CD CD CD , NP , NP}). Thus, we concluded that we would be unable to trust numerical results covering CFG rules.

\begin{figure}[t]
    \centering
    \begin{tabular}{lc} \toprule
         &  Correctly tagged  \\ \midrule
        Wikitext baseline & 0.83 \\
        LSTM & 0.81 \\
        Transformer & 0.84 \\
        TXL & 0.86  \\ \midrule
        WebText baseline & 0.77 \\
        GPT-2 & 0.76 \\ \bottomrule
    \end{tabular}
    \caption{POS sequence vetting}
    \label{tab:pos_seq_vetting}
\end{figure}

\paragraph{POS sequence:} The POS sequence of a sentence is the sequence of POS tags assigned to its words by the constituency parser. For these, we looked at 100 generated sentences identified as having a novel POS sequence, and for each manually checked whether the POS sequence assigned by the parser was accurate. These generally were accurate (Figure \ref{tab:pos_seq_vetting}), and the POS sequences generally had high scores for novelty, so we conclude that the quantitative results are approximately the correct order of magnitude for the POS sequences.

\paragraph{Parse structure:} We define a sentence's parse structure as its constituency parse minus the leaves (i.e., the terminal nodes in the tree). If a sentence has a novel POS sequence, it is guaranteed to also have a novel parse structure. Therefore, we did not conduct an additional inspection of the parse structures on top of the POS sequences, because the parse numbers are close to the POS number.

\newcolumntype{C}[1]{>{\centering\let\newline\\\arraybackslash\hspace{0pt}}m{#1}}

\begin{figure}[t]
    \centering
    \begin{tabular}{lC{1.5cm}c} \toprule
         &  Correctly tagged & Truly novel \\ \midrule
        Wikitext baseline & 0.87 & 0.93 (74/80) \\
        LSTM & 0.90 & 0.85 (58/68) \\
        Transformer & 0.84 & 0.89 (62/70) \\
        TXL & 0.86 & 0.85 (63/74) \\ \midrule
        WebText baseline & 0.78 & 0.95 (71/75) \\
        GPT-2 & 0.75 & 0.90 (54/60) \\ \bottomrule
    \end{tabular}
    \caption{Dependency arc vetting}
    \label{tab:dep_arc_vetting}
\end{figure}

\paragraph{Dependency arcs:} For dependency arcs, we first checked 100 dependency arcs identified as novel for each model to confirm that they were not parser errors. We then looked at the subset of each of these 100-arc sets which were correctly labeled and for which there were no more than 100 training sentences that could possibly contain that arc (i.e., training sentences containing both relevant words). We then checked those training sentences to confirm that the dependency arc in question did not appear in that sentence. These results were generally strong (Figure \ref{tab:dep_arc_vetting}), so we conclude that the numerical results for dependency arcs are approximately correct.

\begin{figure}[t]
    \centering
    \begin{tabular}{lC{1.5cm}c} \toprule
         &  Correctly tagged & Truly novel \\ \midrule
        Wikitext baseline & 0.82 & 0.82 (50/61) \\
        LSTM & 0.85 & 0.81 (35/43)  \\
        Transformer & 0.79 & 0.80 (48/56) \\
        TXL & 0.88 & 0.90 (56/62) \\ \midrule
        WebText baseline & 0.71 & 0.85 (56/66) \\
        GPT-2 & 0.74 & 0.79 (52/66) \\ \bottomrule
    \end{tabular}
    \caption{Dependency role vetting}
    \label{tab:dep_role_vetting}
\end{figure}

\paragraph{Dependency roles:} Similarly to the dependency arcs, we checked 100 dependency roles per model that were identified as novel to check if the roles were correctly identified. We then looked at the subset of those 100-role sets which were correctly labeled and for which there were no more than 100 training sentences that could possibly contain that role (i.e., training sentences containing the relevant word). We then checked those training sentences to see if the dependency role truly was novel. These results were also successful enough for us to conclude that the numerical results were approximately correct and could be reported in the paper (Figure \ref{tab:dep_role_vetting}).

\paragraph{Dependency argument structure:} Dependency argument structure is the list of argument types that a verb has (e.g., \textit{subject, direct object, indirect object}). We did not manually analyze this due to how labor-intensive it would be (requiring analysis of a large number of entire training sentences). Thus, we do not provide numerical results for this, since we do not have an estimate of how reliable such numbers would be.

\section{Example of a mismatch between local and global novelty}\label{app:mismatch}

In both our $n$-gram analyses and syntactic analyses, we found that LMs were less novel than the baseline for local structure (e.g., small $n$-grams or individual dependency arcs) but were more novel than the baseline for larger-scale structure (e.g., large $n$-grams or overall sentence structure).
As an example of how such a local/global mismatch is possible, suppose that the training set contained only (\ref{ex:mismatch_training}):

\ex. alligators resemble crocodiles, and dolphins resemble sharks\label{ex:mismatch_training}

Generated sentences (\ref{ex:mismatch_gen_model}) and (\ref{ex:mismatch_gen_baseline}) are both novel trigrams, but (\ref{ex:mismatch_gen_baseline}) also contains novel bigrams whereas (\ref{ex:mismatch_gen_model}) does not. Thus, these two generations have the same trigram novelty despite differing in their smaller-scale, bigram novelty.

\ex. \a. alligators resemble sharks\label{ex:mismatch_gen_model}
\b. crocodiles resemble dolphins\label{ex:mismatch_gen_baseline}

In terms of syntactic structure, both generated sentences have a novel overall parse structure (since the training set contains only a sentence with two clauses in it, while both generated sentences are only a single clause). Despite having the same novelty for global syntactic structure, these sentences differ in their proportion of individual dependency arcs that are novel: In sentence (\ref{ex:mismatch_gen_baseline}), both noun-verb dependency arcs are novel (\textit{crocodiles} as the subject of \textit{resemble} and \textit{dolphins} as the direct object of \textit{resembles}), while in sentence (\ref{ex:mismatch_gen_baseline}) both of the noun-verb dependency arcs appear in the training set. Thus, as with $n$-gram novelty, (\ref{ex:mismatch_gen_baseline}) and (\ref{ex:mismatch_gen_model}) have different levels of small-scale novelty despite having the same level of global novelty.

\section{Examples of syntactic novelty}\label{app:syntactic_novelty_examples}

Although we cannot report reliable numerical results for several of the aspects of syntax that we considered in Appendix \ref{app:syntax_vetting} (due to the infeasibility of manually checking all examples and the high error rates of automatic methods), it may still be worthwhile to see if there are any individual examples that we can identify to provide an existence proof that models do, at least sometimes, perform the types of syntactic generalization in question.

To that end, we identified several types of syntactic generalization that have received attention in prior literature in language acquisition and/or analysis of LMs, and manually looked for examples that we could verify were novel. For example, given a dependency role (e.g., ``\textit{watch} as the head of an nsubj relation"), we identified all training sentences containing the relevant word (here, \textit{watch}) and then manually checked whether that word ever has the relevant dependency role, in case the parser missed it. This method ensures perfect precision (all examples that we identify are certain to be true instances of syntactic novelty), though there are no guarantees about recall (we may miss some, or even many, examples). 
To decrease the amount of manual inspection required, we only considered candidate examples that required inspecting 500 training sentences or fewer. For each example from a Wikitext-based models, we have also provided a relevant training example, to highlight the type of generalization that is being illustrated (but note that, in all cases, there are many more relevant training examples that we do not include in the paper). We have not included training examples from WebText because it is not a publicly-released dataset.

\paragraph{Novel part-of-speech tags:} Here we looked for generated words which are used as a noun in the generation and which appear in the training set, but never as a noun (for this purpose, we consider all noun POS tags to be identical), and similarly for verbs. We find a few examples; note that some of these might be instances of overgeneralization (that is, English speakers might judge some of them to be ungrammatical): 

\ex. Novel as a noun
\a. 
\a. \textit{Transformer:} They declared two days of \textbf{secede} from Yugoslavia.
\b. \textit{Training example}: Led by a former Jedi named Count Dooku , thousands of planetary systems threaten to \textbf{secede} from the Galactic Republic .
\z.
\b. 
\a. \textit{Transformer:} While Vespasian did not have any public office beyond \textbf{tribunician} as emperor, he was probably appointed as a general of the Praetorians in 63 AD.
\b. \textit{Training example:} As the eldest and most experienced of Vespasian 's sons , Titus shared \textbf{tribunician} power with his father , received seven consulships , the censorship , and was given command of the Praetorian Guard
\z.
\b. 
\a. \textit{Transformer:} Apollodorus of Damascus interpreted the \textbf{colossal} as a gigantic head with the crown of Horus on the left hand side.
\b. \textit{Training example:} This \textbf{colossal} project involved around 600 people to restore the monument and cost a total of US \$ 6,901,243 .
\z.
\b. 
\a. \textit{Transformer:} The \textbf{exploratory}, called California California California California-California-Mission (unk), was originally intended to help rebuild damaged homes and buildings in rural New Mexico in search of new ways to improve access to water and housing to help meet the needs of displaced persons who need to work for their communities.
\b. \textit{Training example:} The \textbf{exploratory} parts of the game feature a series of arcane codes and glyphs , treasure maps and chests , and secret rooms .

\ex. Novel as a verb
\a. 
\a. \textit{TXL:} The team used a series of diodes to \textbf{diode} the electrical component of the circuit board...
\b. \textit{Training example:} Devices with this type of negative resistance include the tunnel \textbf{diode} , resonant tunneling \textbf{diode} , lambda \textbf{diode} , Gunn \textbf{diode} , and dynatron oscillators .

\paragraph{Novel determiner:} In the language acquisition literature, one question that has received focus is about determiner-noun pairs \cite{pine1997slot,yang2013ontogeny}: do children have a productive rule that allows them to combine any determiner with any noun, or have they only memorized particular determiner-noun pairs (e.g., \textit{the dog}, \textit{a cat})? We investigate this general topic by turning to the \textit{det} dependency arc to see if there are any nouns appearing with a determiner they have not appeared with during training. We see examples from all models, showing that models have not simply memorized particular determiner-noun pairs. 

\ex. \textit{the} $\rightarrow$ \textit{a}
\a. 
\a. \textit{LSTM:} The upper part of the tooth, from the upper part of the tooth, is covered with \textbf{a} coarse \textbf{keratin}, usually to the base of the teeth.
\b. \textit{Training example:} The scales of birds are composed of \textbf{the} same \textbf{keratin} as beaks , claws , and spurs
\z.
\b. 
\a. \textit{Transformer:} called The unk featured \textbf{a} "very clever" female \textbf{tilefish} who played a role that she had previously played in the pilot episode.
\b. \textit{Training example:} \textbf{the} \textbf{tilefish} will continue to expand its burrow in the sediment throughout its life .
\z.
\b. 
\a. \textit{Transformer-XL:} \textbf{A} medium-sized \textbf{zebrafish}, this small tadpole has two openings in its snout: a small opening on the top of its head and a large opening at the bottom of its head.
\b. \textit{Training example:} The main species used is \textbf{the zebrafish}
\z.
\b. \textit{GPT-2:} A traditional Tread is a "V"shaped surface that has \textbf{a} solid \textbf{groundplate} in both the outboard sidewalls.

\ex. \textit{a} $\rightarrow$ \textit{the}
\a. 
\a. \textit{LSTM:} The pawn base (sometimes called a unk) is in a position called unk, where most of \textbf{the castling} can be compared to the middlegame.
\b. \textit{Training example:} It is critical to keep the king safe from dangerous possibilities . \textbf{A} correctly timed \textbf{castling} can often enhance this .
\z. 
\b. 
\a. \textit{Transformer:} Porcini have a mild taste, and are edible when cooked, although there is confusion over their culinary value from other edible mushrooms, such as the dried and fermented edulis edulis, the ground pepper, or \textbf{the bouillon} with which they are cooked.
\b. \textit{Training example:} Polish cuisine offers \textbf{a} ruby @-@ colored beetroot \textbf{bouillon} known as barszcz <unk> <unk> , or clear red borscht .
\z.
\b. 
\a. \textit{Transformer-XL:} \textbf{The selfie} featuring Jackson, the character he portrayed in the game, is widely considered a popular culture phenomenon.
\b. \textit{Training example:} \textbf{A selfie} featuring Venkatesh and his son Arjun was released as well .
\z.
\b. \textit{GPT-2:} In my case I just changed my coverband (no case is needed) because \textbf{the} new \textbf{coverband} covers the headphone jacks on my iPod touch!

\paragraph{Novel dependency roles:} We focus on 2 dependency roles analyzed in past work \cite{fodor1988connectionism,kim2020cogs}: do we ever see a word used as a subject that has never appeared as a subject before, or a word used as a direct object that has never appeared as a direct object before? Across models, the answer is yes.

\ex. \textbf{Novel as a subject}
\a. 
\a. \textit{LSTM:} By the first week on opening weekend, domestic distribution had accumulated over \$21.7 million, and its first Thursday \textbf{matinee} had already reached a lower \$5.8 million mark.
\b. \textit{Training example:} Columnist Leonard Pitts of The Miami Herald described his difficulty sitting down to read a book , in which he felt like he " was getting away with something , like when you slip out of the office to catch a \textbf{matinee} " .
\z.
\b. 
\a. \textit{Transformer:} Except for the tail, some members of unk are relatively large, and the \textbf{hindlimb} has large finger bones (unk) on each foot.
\b. \textit{Training example:} Coelurus had a relatively long neck and torso due to its long vertebrae , a long slender \textbf{hindlimb} due to its long metatarsus , and potentially a small slender skull .
\z.
\b. 
\a. \textit{Transformer-XL:} With the loss of Mike Hargrove to a knee injury and the retirement of Charlie Haas as player-manager, Dodgers broadcaster Dave \textbf{Kingman} had to seek some sort of replacement.
\b. \textit{Training example:} The interstate continues to head north until it reaches \textbf{Kingman} .
\z.
\b. \textit{GPT-2:} \textbf{Makiza} is a junior researcher at the Institute for Race Relations and an independent journalist.

\ex. \textbf{Novel as a direct object}
\a. 
\a. \textit{LSTM:} The Chicago Tribune (although it was also the first in a magazine to read The Urantia Book) was the first to feature George \textbf{Saintsbury}, a 19th-century art historian; he felt that it was based on the original text of the Bible.
\b. \textit{Training example:} \textbf{Saintsbury} insists it is " beyond all question one of the very greatest of [ Balzac 's ] works " .
\z. 
\b. 
\a. \textit{Transformer:} A pot of the great Celtic goddess Mithras at Ribeira da Ribeira dates to the 6th or 7th century AD, and finds a copy of a manuscript showing \textbf{Mithras} and about thirty other deities bound around him or her, which included gods such as the goddess Venus and the goddess of the earth and water.
\b. \textit{Training example:} Eldred fled to the world that he and \textbf{Mithras} stand on .
\z.
\b. 
\a. \textit{Transformer-XL:} The Popular Republican Party, led by James Whitcomb Riley, denounced the \textbf{NLRB} and argued that they have no right to interfere in the decisions of the National Labor Relations Board. 
\b. \textit{Training example:} The \textbf{NLRB} appealed to the Supreme Court of the United States 
\z.
\b. \textit{GPT-2:} This is what I wanted, so I used the original Bomber King storyline as the foundation for creating \textbf{RoboWarrior}.

\paragraph{Novel argument structures:} Finally we look at cases where verbs have an argument structure they never have in the training set, where the argument structure defines the set of arguments that the verb has (e.g., \textit{laughed} in \textit{Alex laughed} has the argument structure \textit{active subject}; \textit{explained} in \textit{The rules were explained} has the argument structure \textit{passive subject}; and \textit{saw} in \textit{The doctor saw the lawyer} has the argument structure \textit{subject,direct object}). 
We first checked if a verb that had only ever appeared as active was used as passive or vice versa, and found a small number of these. We also found one example of a shift from transitive to intransitive, where TXL used the verb \textit{suffuses} as an intransitive verb when it had always appeared as transitive in the training set.

\ex. Active to passive
\a. 
\a. \textit{LSTM:} By the afternoon of September 9, the large cyclone system had been \textbf{re-strengthened}, with an increase in the surface circulation of the storm. 
\b. \textit{Training example:} The storm quickly \textbf{re-strengthened} early on September 20 , but transitioned into an extratropical cyclone on September 21 .
\z.
\b. 
\a. \textit{LSTM:} However, the Enterprise crew in sickbay was allowed the craft to be \textbf{re-docked}.
\b. \textit{Training example:} According to mission rules , Orion would have then \textbf{re-docked} with Casper
\z.
\b. 
\a. \textit{Transformer-XL:} In 1864, a large part of the state of Georgia was \textbf{seceded} from the United States to form the Confederate States of America.
\b. \textit{Training example:} Pakistan withdrew in 1972 after East Pakistan \textbf{seceded} and became Bangladesh in 26th March 1971 .
\z.

\ex. Passive to active
\a. 
\a. \textit{Transformer:} They then \textbf{drydocked} at Sasebo on 22 January 1916 to be fitted with an additional 4.5 cm / 40 anti-aircraft (AA) guns.
\b. \textit{Training example:} Ostfriesland was \textbf{drydocked} in Wilhelmshaven for repairs , which lasted until 26 July

\ex. Transitive to intransitive
\a. 
\a. \textit{Transformer-XL:} On March 4, 2009, The New York Times published its own review from journalist Mike Brennan, who thought the novel's structure was too similar to several of his best known works, but praised the "rich and evocative language" that \textbf{suffuses}, as well as the plot as "vivid and convincing." 
\b. \textit{Training example:} Genuine poetry \textbf{suffuses} them , and they are scored with brilliance and resource .
\z.

We observed no confirmed instances of a verb being generated with a transitive usage when it had only been intransitive in training. We also observed no instances of novel dative alternations: using an indirect object (e.g., \textit{I gave \textbf{them} a book}) with a verb that, in training, always used a prepositional object (e.g., \textit{I gave a book \textbf{to them}}), or vice versa. Note that the baselines also contained no confirmed novel transitivity or dative alternations, meaning that the lack of such examples in the models should not be interpreted as evidence that they are incapable of making such generalizations, since these types of novelty are rare even in human-generated text.

\section{Morphology categorization}\label{app:morphology_categorization}

\begin{figure}
    \centering
    \includegraphics[width=\columnwidth]{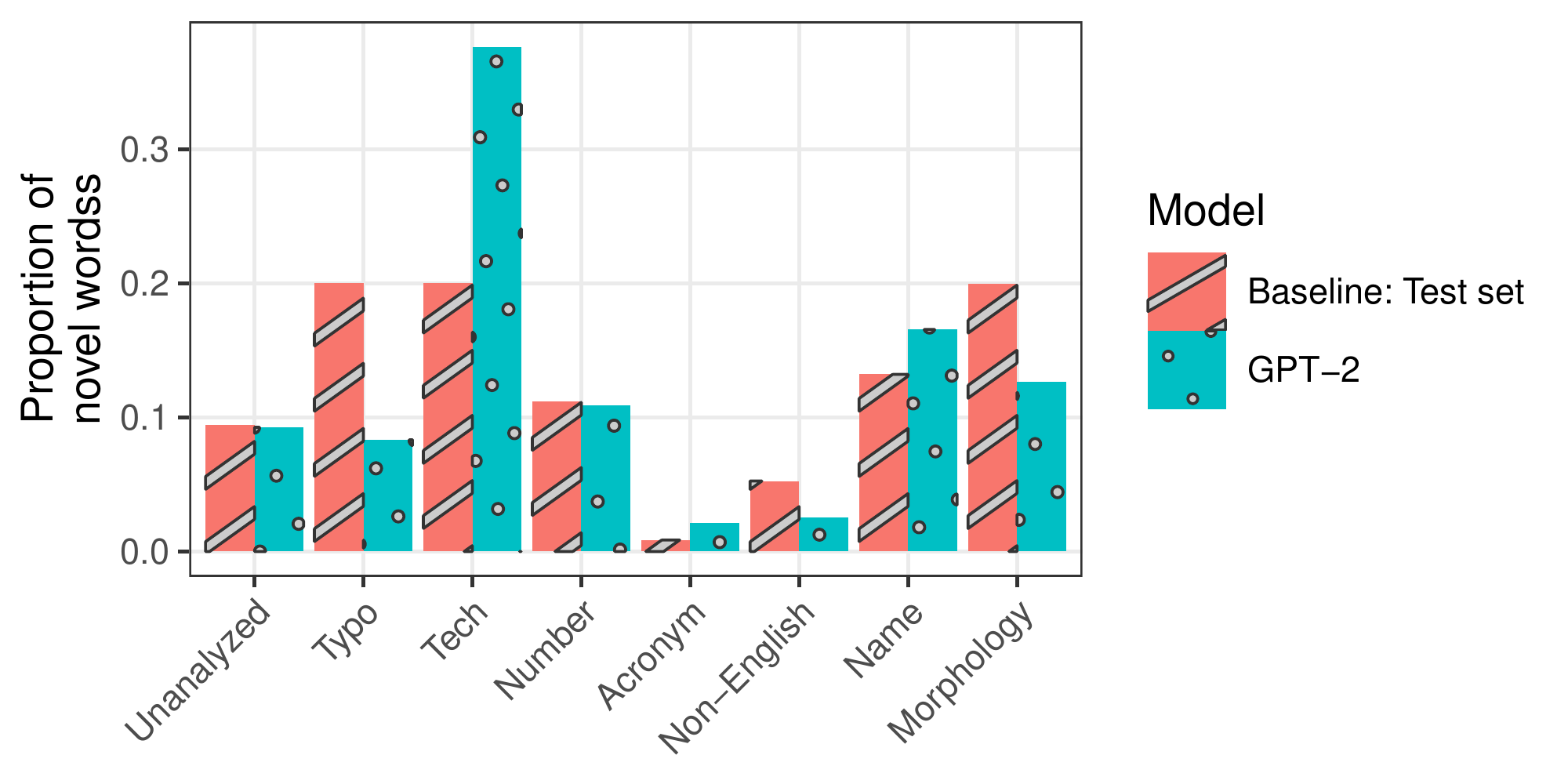}
    \caption{Categorization of the novel words generated by GPT-2}
    \label{fig:novel_words_broad}
\end{figure}

In this section, we give an overview of all the novel unigrams generated by GPT-2. Specifically, we analyze our samples of text generated by the largest size of GPT-2 using top-40 sampling, pooling together the samples from all 4 of our prompt lengths. We then tokenized the generated text using the Moses tokenizer \cite{koehn2007moses} and analyzed all of the resulting tokens that never appeared in GPT-2's training set. We also performed the same procedure for our baseline WebText text. All of the analyses described here were performed manually by one of the authors who is a native English speaker with training in linguistics; we used manual annotation because automatic methods are unlikely to be trustworthy for the rare, long-tail phenomena that give rise to novel unigrams in the generated text.

We define a word as a space-delimited token (after the text has been tokenized with the Moses tokenizer). This means that a word counts as novel as long as this exact form of the word has never appeared during training, but in some cases it may not deviate that much from words in training; for example, it is possible that the word appears during training but with a different capitalization, or with different punctuation (e.g., as two words in a row instead of being hyphenated). We consider all of these to be instances of novelty because, from the model's perspective, all of them are encoded differently. 

In case some readers would prefer a stricter definition of novelty, we searched GPT-2's training data for all of the morphologically-novel words discussed in the main text to see whether any form of that word appeared in the training set. Specifically, we lowercased the morphologically-novel word and removed all characters other than the 26 Roman letters (i.e., we removed all spaces, punctuation, and non-Roman characters), and then searched for the resulting string in the training set after formatting it in the same way. This method was chosen to be extremely thorough; in addition to returning all reasonable ways of formatting words that we could think of (e.g., capitalized/uncapitalized, with punctuation/without punctuation), it also returns many false positives (e.g., a word being split across multiple unrelated words, such as \textit{Welsh} appearing inside \textit{vowel shift}), so we manually inspected all portions of the training set that were identified as candidate matches for the morphologically novel words. Figure \ref{tab:morphology_variants} shows the results, using the following categorization:
\begin{itemize}
    \item \textit{None:} We verified that the word never occurs in the training set in any form.
    \item \textit{Unknown:} Checking whether the word occurs was infeasible because there were too many candidates to check
    \item \textit{Spaced:} The word is a compound word that occurs but with its components separated by a space
    \item \textit{Within larger word:} The word occurs within a larger word in the training set
    \item \textit{Capitalized:} The word occurs but with a different capitalization than it was generated with
\end{itemize}

\begin{figure}
    \centering
    \begin{tabular}{lp{3.5cm}} \toprule
    Word & Related appearances in the training set \\ \midrule
1099es & none \\
752th & unknown\\
anti-Tunisian & none\\
bondbreaking & spaced (\textit{bond breaking})\\
Brazilianism & within larger word (\textit{anti-Brazilianism})\\
Brazilianisms & none \\
Clevelandans & unknown\\
Clevelandians & none\\
co-causation & none\\
cookying & none\\
epineopterygoid & none\\
FOIA-requesters & spaced (\textit{FOIA requesters})\\
Fourteenthly & none\\
Fowleses & none \\
front-floor & spaced (\textit{front floor})\\
genoshaans & none\\
genoshans & capitalized (\textit{Genoshans})\\
Hamiltonans & none\\
headswaters & none\\
Huamangas & none\\
Klymits & unknown\\
LHAWs & unknown\\
load-samples & unknown\\
M-Sinks & unknown\\
mushrooms-related & none\\
no-knockout & spaced (\textit{no knockout})\\
proto-poetry & different punctuation (\textit{protopoetry})\\
re-nitrification & none\\
ridiculousities & none\\
Sarrats & none\\
SQLes & unknown\\
Sub-epineopterygoid & none\\
Summission & none\\
tearingros & none\\
tearro & unknown\\
tearsros & unknown\\
Thirteenthly & none\\
tornro & unknown\\
Torpexes & none\\
townites & none\\
un-competition & unknown\\
\bottomrule
    \end{tabular}
    \caption{Variant forms of words discussed in the main text that display morphological novelty.}
    \label{tab:morphology_variants}
\end{figure}

\noindent
Returning to the full set of novel unigrams (not those just discussed in the main paper), we divided all of the novel unigrams into 8 broad categories. Figure \ref{fig:novel_words_broad} gives an overview of how common each category is in the GPT-2-generated text and in the baseline; in the rest of this section, we describe these categories and give more details about the types of words that make them up. 

\subsection{Unanalyzed}

This category is composed of examples for which we could not discern any internal structure. Of the 838 such examples, most of them (658) were strings of random characters:

\ex. \a. src = `` \textbf{jE4B9BpL9KWv} "
\b. fvw vnvh \textbf{qvwq} wvqw , kqw .

This category also contains 180 examples, like the following, which are pronounceable words but for which we could not find any apparent meaning or internal structure.

\ex. \a. \textbf{Narcow} . Mike Tyson says he 's `` on track...
\b. \textbf{Aumulule} may have been constructed of wool or coarse flax .

\subsection{Typo}

The generated text contains 754 typographical errors. Most of these errors (706 of them) are caused by improper spacing and/or punctuation, sometimes as an error of the language model (\ref{ex:punct_lm}) and other times as an error of the tokenizer that we used to post-process the text (\ref{ex:punct_tokenizer}).

\ex. \a. I 'm going to have to use a different material besides \textbf{plastic.Here} 's the front and back \label{ex:punct_lm}
\b. This material may not be published , broadcast , rewritten , or \textbf{redistributed.} \label{ex:punct_tokenizer}

The remaining 48 typos are ones that involve misspelled words; e.g., (\ref{ex:efforteless}) has \textit{efforteless} instead of \textit{effortless}, and (\ref{ex:oceansic}) has \textit{oceansic} instead of \textit{oceanic}. 

\ex. \a. I use these words to give these thoughts a form in such an \textbf{efforteless} and natural way that I may never really know where the person is coming from \label{ex:efforteless}
\b. Because of the potential influence of atmospheric and \textbf{oceansic} carbon dioxide on the global carbon cycle \label{ex:oceansic}

The generated text contains far fewer typos than the baseline, human-generated text does. We conjecture that this difference arises because humans generate text character-by-character, which might create more opportunities for typos than the subword-based generation process used by GPT-2.

\subsection{Tech}

The most common category of novel words generated by GPT-2 is technology-related terms including both URLs or parts of URLs\footnote{We provide no examples of URLs in case the URLs currently, or in the future, lead to websites containing viruses or harmful content.} and words used in computer code (e.g., variable names) (\ref{ex:coding}). The baseline text also contains a fair number of novel words in this category, but they are only about half as frequent as in the generated text.

\ex. \label{ex:coding} \a.  may now return the string `` $<$ template name = ` \textbf{views.template.name} ' $>$ " .
\b. \$ success = \$ auth - $>$ \textbf{successMessageText} ( `Logged In ') ;
\c. row-reverse ( columns ) \{ \textbf{--widths} 2 --columns 2 ;

We do not make any systematic analysis of how such words are structured, but constructing them likely does require several types of string manipulation, including:

\begin{itemize}
    \item ``Affixation" such as adding \textit{www} or \textit{com} in URLs, or adding the double dash used to introduce an argument in computer code
    \item Concatenation of multiple words, sometimes with a delimiting token such as a period, dash, or underscore
    \item Manipulation of case, particularly converting words entirely to lowercase for URLs, or the use of camelcase within code.
\end{itemize}

\subsection{Number}

GPT-2 generates several types of novel words that we classify as numbers. First are real numbers (672 of them; \ref{ex:numbers}); these are generally well-formed, but note the one example that improperly starts with a 0 (\ref{ex:number_zero_initial}).

\ex. \label{ex:numbers} \a. to \$ 113.48 in trading while the S \& P 500 rose 0.6 percent to \textbf{2,022.64} points .
\b. from July 2008 through March 2015 , to make approximately \$ \textbf{2,569,913} worth of unauthorized loans from military-owned financial institutions .
\c. For the census year 2011 , there were 3,878,000 male-female households ( \textbf{0,936,000} female households in same-sex couples ) . \label{ex:number_zero_initial}

There are also 49 novel dates (\ref{ex:dates}),\footnote{Note that we are only analyzing novel unigrams; there are likely more novel dates that include spaces, along the lines of \textit{June 28, 2021}.} all of which form possible dates (i.e., there are no impossible dates such as November 31 or July 33). Some of these dates use the form month-date-year (\ref{ex:monthday1}--\ref{ex:monthday2}), while others use the form date-month-year (\ref{ex:daymonth1}--\ref{ex:daymonth2}).

\ex. \label{ex:dates} \a. \textbf{Jun-14-16}
\b. \textbf{2013-11-12T18}
\c. \textbf{27-April-2018}
\d. \textbf{11-28-1998}\label{ex:monthday1}
\e. \textbf{04-24-1987}\label{ex:monthday2}
\e. \textbf{13-08-2025}\label{ex:daymonth1}
\e. \textbf{15-08-2027}\label{ex:daymonth2}

There are also 75 instances of a number plus a unit, such as the following: 

\ex. \a. Minimum Water Pressure : \textbf{2.5atm}
\b. Tags : ch.5 , ch.5.1 , \textbf{ch.5.2} , ch.6 , \textbf{ch.6.1} , ch.6.2
\c. Available OS Memory : \textbf{8147MB} RAM

Finally, GPT-2 generates many novel phone numbers (190 of them), mostly following the 10-digit (or 11-digit) format used for North American phone numbers. These phone numbers appear with multiple formats: XXX-XXX-XXXX, 1-XXX-XXX-XXXX, XXX.XXX.XXXX, or (XXX) XXX-XXXX.\footnote{We do not provide examples of generated phone numbers: even though the generated phone numbers are not in the training set, they may still belong to a person who would not want their phone number published in this paper.}

North American phone numbers can also be expressed with some letters instead of numbers, and we see GPT-2 formatting 4 phone numbers in this way too.
However, in all 4 cases, the correspondences between the letters and the numbers provided afterward are incorrect.

\subsection{Acronym}
In our generated text, there are 195 examples of acronyms. Of these, 75 appear along with the full version of what the acronym stands for; some examples are below. 

\ex. \label{ex:acronyms_good_app} 
\a. The Money Funders International Group ( \textbf{MFIG} )
\b. the Cathedral Development Strategy Review Group ( \textbf{CDSRG} )
\c. The National Census and Statistics Bureau ( \textbf{NCBSB} )
\d. The Nigerian Institute for Demographic and Social Research ( \textbf{NIDRS} )\label{ex:acronyms_other_c_app}

\subsection{Non-English}

Even though GPT-2 is primarily an English model, sometimes it generates words that are clearly meant to be interpreted as words in another language. In many cases the intended language is specified: these languages include Arabic, Aramaic, Bulgarian, Czech, Dutch, Esperanto, French, German, Greek, Hebrew, Hungarian, Icelandic, Inuktitut, Japanese, Latin, Mandarin, Middle Dutch, Old English, Old Frisian, Old Norse, Proto-Germanic, Quenya, Russian, Spanish, Swedish, and Swiss German. We do not attempt any investigation of whether these words are actually valid words in the languages they are purported to be in, or whether they have the meanings that they have been attributed.

Within this category we also include the 6 instances of GPT-2 providing pronunciation hints. These are generally somewhat similar to the true pronunciations of the words they are meant to correspond to, but are far from perfect:

\ex. \a. Voltaire ( pronounced \textbf{VORE-ey} )
\b. Tom Paine ( pronounced \textbf{TOHR-in} )

\subsection{Name}

We see 1,499 novel names in the generated text. 7 are names of languages or dialects (\ref{ex:names_language}). 655 are online usernames.\footnote{We provide no examples of usernames in case the examples are real people's usernames.} 39 are first names (the first word in a multi-word name) (\ref{ex:names_first}), 220 are last names (the last word in a multi-word name) (\ref{ex:names_last}), while 150 are sole names (the only word in a single-word name) (\ref{ex:names_sole}).\footnote{Note that cultures differ in terms of the interpretation of first/last names; for instance, in some cultures, one's first name is one's family name, while in other cultures one's last name is one's family name. Because we only have access to the words generated by the models, but no definitive access to their meanings, we classify names based on their position (first vs. last) rather than their meaning (family vs. given).} 11 are names of groups of people (\ref{ex:names_group}). 120 are names of places (\ref{ex:names_place}). 21 are names for which we were unsure of the type of name (e.g., name of a person or of a place?) (\ref{ex:names_name}). 206 were corporate names: names of products, companies, or organizations (\ref{ex:names_corporate}). 70 are parts of scientific names for species (\ref{ex:names_scientific}).

\ex. \label{ex:names_language}
\a. Some of the dialects are known as Inuit Nunaat , Inuit Yupik , Inuit \textbf{Inupialik}
\b. \textbf{Naahauhau} : One of the most widely used dialects in Northland with many dialects of

\ex. \label{ex:names_first}
\a. \textbf{Abduraziz} Akhmadov
\e. \textbf{Armidio} F. Santos
\e. \textbf{Daryousu} Nyamitwe
\e. \textbf{Deshpur} Singh
\e. \textbf{Elianza} Dina
\e. \textbf{Gerewyn} Davies
\e. \textbf{Ikkuei} Chisato
\e. \textbf{Jeeun} Jang
\e. \textbf{Mari-Reijo} Aho
\e. \textbf{Qi-Juan} Wang
\e. \textbf{Rene-Laurent} Bédard
\e. \textbf{Sapsiso} Nkomo
\e. \textbf{Tsu-Bin} Mehta
\e. \textbf{Xian-Lai} Yuan

\ex. \label{ex:names_last}
\a. Ahmet \textbf{Davutoiu}
\b. Anders \textbf{Kjølveberg}
\e. Celine \textbf{Darpien}
\e. Daniel \textbf{Vosschevsky}
\e. David M. \textbf{Kornkrantz}
\e. Eka \textbf{Khodayeva}
\e. Emily \textbf{Kinshel}
\e. Hasan \textbf{Husefije}
\e. Hooman \textbf{Mashharipour}
\e. Isabel \textbf{Pérez-Reyes}
\e. Jacques \textbf{Prévens}
\e. Jafar \textbf{al-Jubaidi} \label{ex:name_al}
\e. José Luis \textbf{Hernández-Méndez}
\e. Kim \textbf{Ki-nong}
\e. Kurcan \textbf{Çapça}
\e. Liam \textbf{Gwozdan-Morgan}
\e. Linda \textbf{Nix-Hogar}
\e. Linda \textbf{Kanjiyagawa}
\e. Liu \textbf{Wu-Qin}
\e. Matt \textbf{Wijeysinga}
\e. Muhrhar \textbf{Rükeli}
\e. Njoki \textbf{Rakotombe}
\e. Reinhard \textbf{Dansberger}
\e. Tom \textbf{McCallaghan} \label{ex:name_mc}
\e. Vladimir \textbf{Chmovarov}
\e. Zhang \textbf{Gongtian}

\ex. \label{ex:names_sole}
\a. one day he returned to his wife , \textbf{Safwanah}
\b. an extra terrapin named \textbf{Puddleguy}
\c. \textbf{Sutsabaiesan} has been working with the Maple Leafs

\ex. \label{ex:names_group}
\a. There existed a community of people called `` \textbf{Srisleha} " within Jodhpur
\b. served under the leadership of Lord Shadowsun of the Clan \textbf{Esh-Or} 

\ex. \label{ex:names_place}
\a. transferred to a police facility in \textbf{Hohenkreuz} , near Bremen
\b. \textbf{Knettestown} \label{ex:name_town}
\c. the small town of \textbf{Vazir-Dzorbievsk}
\d. travel through the world of `` \textbf{Runerealm} "
\e. on the bridge over the river \textbf{Vosne-et-Oise}
\e. a 30min train ride from \textbf{Kuparupu}
\e. in the town of \textbf{Uraojpuram} in Tamil Nadu
\e. in the town of \textbf{Mikhaylenka} , central Ukraine
\e. many other locations at \textbf{Chikamagaloor} where sandstone rock have been quarried
\e. A \textbf{Dagenhamshire} Police spokesperson
\e. \textbf{Wetah} Besar , a city that is famous for its beach
\e. a short walk to \textbf{Úlfsmörk} , \textbf{Þjórsmörk} and \textbf{ístmörk}

\ex. \label{ex:names_name}
\a. the amulet of \textbf{Peryob}
\b. the ring of \textbf{Beryemuonk}

\ex. \label{ex:names_corporate}
\a. told \textbf{Skyfoto} TV
\b. Photo Credits \textbf{RedStock} / Pixland / Getty Images
\c. Sign up for \textbf{OurEarth} magazine
\d. a fitness track from Adidas called \textbf{Sportstiq} is 
\e. \textbf{Playstone} Entertainment
\e. fintech companies such as \textbf{Crowdfin}

\ex. \label{ex:names_scientific}
\a. Acanthopanax \textbf{achariensis}
\b. Nothofagus \textbf{acantholytica}
\c. The genus \textbf{Sarconyx} contains three species
\d. S. \textbf{schirani}

Beyond classifying the types of names that are present, we do not analyze the structure of these names. However, this would be potentially interesting to look into; it is non-trivial to construct a word that is pronounceable and ``looks like" a name, such as generating sequences of syllables that look like plausible names of humans (\ref{ex:names_first} and \ref{ex:names_last}), or like the Greek- and Latin-based words used for scientific names (\ref{ex:names_scientific}). Further, some of these names require some more structured string manipulations, such as lowercasing and concatenating words to form usernames, or using affixes within names such as \textit{al-} (\ref{ex:name_al}) and \textit{Mc-} (\ref{ex:name_mc}) in names of people, or \textit{-town} in place names (\ref{ex:name_town}).

\subsection{Morphology}

\begin{figure*}
    \centering
    \begin{tabular}{p{2cm}p{1cm}p{2cm}p{1cm}p{1.5cm}p{1.5cm}} \toprule
        & & & & Count in generated text & Count in baseline \\ \midrule
        Inflectional morphology & & & & 243 & 388 \\
        & Nouns & & & 210 & 331 \\
        & & Plurals & & 74 & 116 \\
        & & Possessives & & 136 & 215 \\
        & Adjectives & & & 8 & 21 \\
        & & Comparatives & & 6 & 10 \\
        & & & \textit{-er} & 0 & 0 \\
        & & & \textit{more} & 6 & 10 \\
        & & Superlatives & & 2 & 11 \\
        & & & \textit{-est} & 0 & 4 \\
        & & & \textit{most} & 2 & 7 \\
        & Verbs & & & 25 & 36 \\
        & & \textit{-ed} & & 12 & 27 \\
        & & \textit{-ing} & & 13 & 9 \\
        & & \textit{-s} & & 0 & 0 \\
        \bottomrule
    \end{tabular}
    \caption{Inflectional morphology}
    \label{tab:inflectional_morphology}
\end{figure*}

\begin{figure*}
    \centering
    \begin{tabular}{ccccp{1.5cm}p{1.5cm}} \toprule
        & & & & Count in generated text & Count in baseline \\ \midrule
        Compounds & & & & 781 & 1112 \\
        & Dephrasal & & & 84 & 174 \\
        & & n-p-n & & 15 & 24 \\
        & & Other dephrasal & & 69 & 150 \\
        & Coordinative & & & 115 & 79 \\
        & & List & & 3 & 3 \\
        & & Other coordinative & & 112 & 76 \\
        & Compound nouns & & & 243 & 399 \\
        & & N-centered & & 230 & 372 \\
        & & & N + N & 160 & 259 \\
        & & & Adj + N & 52 & 83 \\
        & & & Adv + N & 1 & 2 \\
        & & & V + N & 1 & 8 \\
        & & & N + postpositive & 8 & 7 \\
        & & & N + Num & 8 & 13 \\
        & & Verb-centered & & 13 & 27 \\
        & & & N + deverbal N & 13 & 27 \\
        & Compound adjectives & & & 281 & 391 \\
        & & Adj-centered & & 65 & 114 \\
        & & & N + Adj & 51 & 43 \\
        & & & Adv + Adj & 14 & 70 \\
        & & & Adj + postpositive & 0 & 1 \\
        & & N-centered & & 45 & 57 \\
        & & & Num + N & 44 & 52 \\
        & & & P + N & 1 & 3 \\
        & & & N + P & 0 & 2 \\
        & & V-centered & & 171 & 220 \\
        & & & Adj + gerund & 5 & 4 \\
        & & & Adj + passive & 10 & 7 \\
        & & & Adj + V & 3 & 2 \\
        & & & N + gerund & 45 & 69 \\
        & & & N + passive  & 105 & 133 \\
        & & & V + P & 2 & 5 \\
        & & & V + postpositive & 1 & 0 \\
        & Compound verbs & & & 1 & 9 \\
        & & V-centered & & 1 & 9 \\
        & & & P + V & 1 & 0 \\
        & & & Adv + V & 0 & 3 \\
        & & & N + V & 0 & 4 \\
        & & & V + V & 0 & 1 \\
        & & & V + Adv & 0 & 1 \\
        & Neoclassical compounds & & & 57 & 60 \\
        \bottomrule
    \end{tabular}
    \caption{Compounds}
    \label{tab:compounds}
\end{figure*}

\begin{figure*}
    \centering
    \begin{tabular}{cp{1.5cm}p{1.5cm}p{1.5cm}} \toprule
        & & Count in generated text & Count in baseline \\ \midrule
        Diminutives & & 6 & 13 \\
        & \textit{-ling} & 1 & 1 \\
        & \textit{-lite} & 2 & 0 \\
        & \textit{-y} & 0 & 1 \\
        & \textit{demi-} & 0 & 1 \\
        & \textit{down-} & 0 & 1 \\
        & \textit{hemi-} & 1 & 0 \\
        & \textit{micro-} & 0 & 3 \\
        & \textit{mini-} & 1 & 0 \\
        & \textit{nano-} & 1 & 4 \\
        & \textit{semi-} & 0 & 1 \\
        & \textit{under-} & 0 & 1 \\
        Augmentatives & & 4 & 15 \\
        & \textit{hyper-} & 0 & 3 \\
        & \textit{mega-} & 0 & 3 \\
        & \textit{over-} & 1 & 1 \\
        & \textit{super-} & 2 & 4 \\
        & \textit{ultra-} & 1 & 2 \\
        & \textit{up-} & 0 & 2 \\
        \bottomrule
    \end{tabular}
    \caption{Derivational affixation, part 1 of 4: Diminutives and augmentatives}
    \label{tab:affixes1of4}
\end{figure*}

\begin{figure*}
    \centering
    \begin{tabular}{cp{1.5cm}p{1.5cm}p{1.5cm}} \toprule
        & & Count in generated text & Count in baseline \\ \midrule
        Location in time and space & & 16 & 39 \\
        & \textit{ante-} & 0 & 1 \\
        & \textit{circum-} & 0 & 1 \\
        & \textit{cis-} & 0 & 1 \\
        & \textit{cross-} & 1 & 1 \\
        & \textit{inter-} & 0 & 1 \\
        & \textit{intra-} & 0 & 1 \\
        & \textit{meta-} & 0 & 1 \\
        & \textit{off-} & 0 & 2 \\
        & \textit{outer-} & 0 & 2 \\
        & \textit{post-} & 6 & 10 \\
        & \textit{pre-} & 2 & 11 \\
        & \textit{proto-} & 1 & 4 \\
        & \textit{sub-} & 3 & 1 \\
        & \textit{supra-} & 0 & 1 \\
        & \textit{trans-} & 3 & 1 \\
        Negatives and reversatives & & 32 & 55 \\
        & \textit{anti-} & 8 & 11 \\
        & \textit{counter-} & 0 & 4 \\
        & \textit{de-} & 1 & 3 \\
        & \textit{dis-} & 2 & 1 \\
        & \textit{dys-} & 1 & 0 \\
        & \textit{in-} & 1 & 1 \\
        & \textit{mis-} & 0 & 1 \\
        & \textit{no-} & 1 & 0 \\
        & \textit{non-} & 15 & 25 \\
        & \textit{un-} & 3 & 9 \\
        Positives and repetitives & & 9 & 12 \\
        & \textit{pro-} & 2 & 4 \\
        & \textit{re-} & 7 & 8 \\
        Residency & & 14 & 11 \\
        & \textit{-an} & 12 & 11 \\
        & \textit{-ite} & 2 & 0 \\
        \bottomrule
    \end{tabular}
    \caption{Derivational affixation, part 2 of 4: Location in time and space; negatives and reversatives; positives and repetitives; and residency morphemes.}
    \label{tab:affixes2of4}
\end{figure*}

\begin{figure*}
    \centering
    \begin{tabular}{ccp{1.5cm}p{1.5cm}} \toprule
        & & Count in generated text & Count in baseline \\ \midrule
        Verb or adjective to noun & & 20 & 21 \\
        & \textit{-ant} & 0 & 1 \\
        & \textit{-ation} & 5 & 1 \\
        & \textit{-er} & 7 & 9 \\
        & \textit{-ion} & 1 & 1 \\
        & \textit{-ity} & 3 & 2 \\
        & \textit{-ment} & 1 & 1 \\
        & \textit{-ness} & 3 & 6 \\
        Noun to noun & & 28 & 34 \\
        & \textit{-dom} & 3 & 1 \\
        & \textit{-ese} & 0 & 1 \\
        & \textit{-fest} & 0 & 2 \\
        & \textit{-gate} & 0 & 1 \\
        & \textit{-ia} & 0 & 2 \\
        & \textit{-iana} & 1 & 0 \\
        & \textit{-ism} & 4 & 9 \\
        & \textit{-ist} & 5 & 3 \\
        & \textit{-ista} & 1 & 0 \\
        & \textit{-land} & 0 & 2 \\
        & \textit{-osphere} & 0 & 1 \\
        & \textit{-ry} & 2 & 0 \\
        & \textit{-scape} & 1 & 0 \\
        & \textit{-ship} & 1 & 0 \\
        & \textit{-thon} & 0 & 2 \\
        & \textit{-verse} & 2 & 2 \\
        & \textit{co-} & 3 & 0 \\
        & \textit{e-} & 3 & 5 \\
        & \textit{ex-} & 2 & 2 \\
        & \textit{mal-} & 0 & 1 \\
        \bottomrule
    \end{tabular}
    \caption{Derivational affixation, part 3 of 4: Noun derivation}
    \label{tab:affixes3of4}
\end{figure*}

\begin{figure*}
    \centering
    \begin{tabular}{lp{1.2cm}p{1.5cm}p{1.5cm}} \toprule
        & & Count in generated text & Count in baseline \\ \midrule
        Adjective creation & & 26 & 51 \\
        & \textit{-able} & 2 & 2 \\
        & \textit{-al} & 4 & 3 \\
        & \textit{-ary} & 0 & 1 \\
        & \textit{-esque} & 0 & 5 \\
        & \textit{-ian} & 2 & 3 \\
        & \textit{-ic} & 3 & 6 \\
        & \textit{-inal} & 0 & 1 \\
        & \textit{-ine} & 2 & 0 \\
        & \textit{-ish} & 2 & 6 \\
        & \textit{-less} & 2 & 1 \\
        & \textit{-like} & 6 & 14 \\
        & \textit{-th} & 2 & 1 \\
        & \textit{-y} & 0 & 8 \\
        & \textit{nigh-} & 1 & 0 \\
        Verb creation & & 7 & 9 \\
        & \textit{-ate} & 0 & 2 \\
        & \textit{-ify} & 3 & 1 \\
        & \textit{-ise} & 1 & 1 \\
        & \textit{-ize} & 2 & 3 \\
        & \textit{a-} & 0 & 1 \\
        & \textit{con-} & 1 & 0 \\
        & \textit{out-} & 0 & 1 \\
        Adverb creation & & 2 & 10 \\
        & \textit{-ly} & 2 & 6 \\
        & \textit{-wise} & 0 & 4 \\
        Other & & 3 & 0 \\
        & \textit{-mon} & 1 & 0 \\
        & \textit{digi-} & 2 & 0 \\
        \bottomrule
    \end{tabular}
    \caption{Derivational affixation, part 4 of 4: Adjective derivation, verb derivation, adverb derivation, and miscellaneous affixes.}
    \label{tab:affixes4of4}
\end{figure*}

\begin{figure*}
    \centering
    \begin{tabular}{p{2cm}p{2cm}p{1.5cm}p{1.5cm}} \toprule
        & & Count in generated text & Count in baseline \\ \midrule
        Character manipulation & & 5 & 38 \\
        & Capitalization & 0 & 10 \\
        & Creative spelling & 0 & 6 \\
        & Letter repetition & 1 & 4 \\
        & Nervousness & 2 & 6 \\
        & Onomatopoeia & 2 & 13 \\
        Portmanteau words & & 3 & 31 \\
        \textit{Schm-} reduplication & & 0 & 1 \\ \bottomrule
    \end{tabular}
    \caption{Non-affix-based morphology.}
    \label{tab:nonaffix}
\end{figure*}

Finally, we analyze the words that use morphology---linguistic derivation of novel words. Our categorization largely follows that of \textit{The Cambridge Grammar of the English Language} \cite{huddleston2001cambridge}, chapters 18 \cite{palmer2001inflectional} and 19 \cite{bauer2001lexical}. 

\subsubsection{Inflectional morphology}

Inflectional morphology is the inflection of a word---not creating a new word, rather simply changing some grammatical feature of the word. Figure \ref{tab:inflectional_morphology} gives an overview of the inflectional morphology found among the novel words in our text samples.

\paragraph{Nouns}

There are two types of inflected forms for English nouns: plurals and possessives. Both occur in the generated text.

The generated text includes the plural forms of common nouns (\ref{ex:plural_common_1}--\ref{ex:plural_common_2}), proper names (\ref{ex:plural_name}), and abbreviations (\ref{ex:plural_abbr}). Most of the plurals are formed using the \textit{-s} form of the plural morpheme, but there are some formed with \textit{-es}, such as (\ref{ex:plural_es}). Though the English language features a few pluralization processes other than the \mbox{\textit{-(e)s}} suffix (e.g., changing \textit{-um} to \textit{-a}, as in \textit{bacterium/bacteria}), none of them are employed in the sample of generated text that we analyzed.

\ex. \textbf{Plurals}
\a. The Commission has taken no action with respect to the issuance of \textbf{nonpublications} in this or any other area . \label{ex:plural_common_1}
\b.  A new generation of leadership , the so-called `` Indonesia \textbf{Dreamists} , " include the candidates of the Movement of New Forces and PKS-Partido Komunist , \label{ex:plural_common_2}
\c. Ann spent her days cycling up and down her city by train . The \textbf{Sarrats} were lucky to have her as part of their lives \label{ex:plural_name}
\d. The U.S.R.A. alleged that the \textbf{C.O.O.s} of approximately 350 horse racing clubs conspired in 1990\label{ex:plural_abbr}
\e. Torpex NEWLINE NEWLINE \textbf{Torpexes} are small hardpoints found on smaller ships \label{ex:plural_es}

There are also possessive forms of both proper nouns (\ref{ex:poss1}--\ref{ex:poss2}) and common nouns (\ref{ex:poss3}--\ref{ex:poss4}).\footnote{While we mostly focus only on novel unigrams in this section, these possessives are bigrams, as our tokenizer separates the \textit{'s} from the rest of the word.} The observed possessives involve both forms of the possessive morpheme: \textit{'s} as in (\ref{ex:poss1}--\ref{ex:poss3}), and \textit{'} as in (\ref{ex:poss4}).

\ex. \textbf{Possessives}
\a. According to a report by UK accounting firm Deloitte , \textbf{Fregoli 's} pizza delivery service achieved \$ 250m of revenues in 2014 \label{ex:poss1}
\b. or from their own personal maps as an alternative to \textbf{MapMyRide 's} online maps .\label{ex:poss2}
\c. The \textbf{shillelagh 's} name comes from the word `` shillelagh , " a military term meaning \label{ex:poss3}
\d. We hope that what we uncover in our \textbf{census-takers '} reports will lead to a better understanding of how to best serve New York City \label{ex:poss4}

\paragraph{Adjectives}

English adjectives can be inflected for grade (also known as degree). This can take the form of a suffix (\textit{-er} or \textit{-est}), or a preceding word (\textit{more} or \textit{most}). The generated text contains no instances of adjectives inflected with the suffixes, but there were a few examples with the separate words \textit{more} (\ref{ex:comparatives}) and \textit{most} (\ref{ex:superlatives}).\footnote{These \textit{more} and \textit{most} examples are the other cases of bigrams that we consider in this section, besides the possessives. All other examples in this section are unigrams.} 

\ex. \textbf{Comparatives} \label{ex:comparatives}
\a. We 've become much \textbf{more androcentric}
\b. and other news reports indicate that this particular incident represents a trend among the \textbf{more socially-diverse} , tolerant areas of Japan

\ex. \textbf{Superlatives} \label{ex:superlatives}
\a. rabbits we adopted , which ate wheat products , are now two of the \textbf{most gluten-intolerant} cats we know .
\b. the `` F-14 class of fighters , the \textbf{most stealth-capable} combat aircraft ever manufactured

\paragraph{Verbs}

There are several examples of generated verbs inflected with the suffix \textit{-ed}; in all cases, these examples are used as passive verbs (\ref{ex:ed_passive}) or past participles (\ref{ex:ed_participle})---that is, there are no instances where it is used to form a novel active past tense.

\ex. \textbf{Passive uses of \textit{-ed}} \label{ex:ed_passive}
\a. to the awesomeness of the Swiss to the point of being \textbf{Swissified}
\b. Most traffic signals are either signified with a sign or \textbf{roadmarked} with either an arrow or flashing yellow lights .

\ex. \textbf{Participial uses of \textit{-ed}} \label{ex:ed_participle}
\a. In addition to the \textbf{self-profiled} and self-labeled population ,
\b. the word in its more \textbf{colloquialised} form is most often associated with the 2000s

There are also examples using \textit{-ing}:

\ex. \textbf{\textit{-ing}}
\a. and GCHQ was spying on this too ( for example , by `` \textbf{cookying} " certain searches on the internet ) .
\b. I hear how you worry that you 're being judgmental or \textbf{judgmentalizing} 
\c.  will revert back to the normal way of updating your OS , which includes completely \textbf{re-restoring} all applications and settings to their default settings

There are no novel verbs with the last remaining major type of inflectional verb morphology, namely the suffix \textit{-s} used to form 3rd-person present-tense verbs.

\subsubsection{Lexical word-formation}

Under this heading, we include all processes that create novel words (as opposed to novel inflections of existing words). 

\paragraph{Compounds}

The text generated by GPT-2 contains many types of compound words, summarized in Figure \ref{tab:compounds}.
First are dephrasal compounds, created by converting an entire phrase into a single word by conjoining its words with hyphens. Many of these are of the form \textit{noun-preposition-noun} (\ref{ex:npn1}--\ref{ex:npn2}), but there are many more with a wide range of other structures (\ref{ex:dephrasalfirst}--\ref{ex:dephrasallast}).

\ex. \textbf{Dephrasal compounds}
\a. It looks as if we are going to continue our \textbf{spending-to-student} ratios going up \label{ex:npn1}
\b. it became , in \textbf{comics-as-genre-wholesale} parlance , a `` reset " of sorts . \label{ex:npn2}
\c. the governor signed a similar law — with a different version of the controversial `` \textbf{we-told-you} " law — into law less than a month ago . \label{ex:dephrasalfirst}
\d. Here 's just a snippet of what Demetrio said he sees in his \textbf{two-year-and-counting} career of working at the MTA
\e. So I feel no obligation to go along with Jimmy in finding the \textbf{good-guy-but-evil-man} , unless he 's a good guy .
\e. that happened Friday was that President Obama and President Xi stopped playing the `` it \textbf{'s-a-civilized-country-and-we-should-just-accept-each-other} " game and started working together .  \label{ex:dephrasallast}

Another common type of compound is coordinative compounds, in which all elements of the compound have an equal status, and the meaning of the whole compound is essentially the meaning arrived by joining all of its elements with the word \textit{and}. A few of these compounds are ones that we categorize as lists because the order of the elements matter (\ref{ex:compound_list}), but most are of the more general type that involves joining the elements with no obvious reason for the ordering (\ref{ex:coordinativefirst}--\ref{ex:coordinativelast}).

\ex. \textbf{Coordinative compounds}
\a. So we go \textbf{left-right-right} and left-right-left . \label{ex:compound_list}
\b. According to the CIA , the \textbf{Soviet-Azerbaijani} relationship had become increasingly `` intense " in recent years . \label{ex:coordinativefirst}
\e. What really happened on Tatooine ? The \textbf{Hutt-Kling-Mandalorian} trade dispute ?
\e. Fox 's \textbf{Titans-Cowboys} game at 12 : 00 p.m. on September 9 averaged 8.5 million viewers
\e. the moons of Jupiter provided the foundation for what came to be called the `` \textbf{Newton-Laplace-Einstein} model of the solar system .
\e. It has its roots in the `` \textbf{anarchism-capitalism} " debate
\e. the Northern \textbf{Norway-Iceland} border
\e. proton beam producing roughly the same number of gluon plasma as is observed for the \textbf{proton-photon} collisions ;
\e. Pa had a little iron \textbf{grill-barbecue} outside
\e. And it was very easy for him to be a \textbf{painter-scientist} , because he used nature and art .
\e. the most successful form of handstand execution known to man : the \textbf{Krzyzewski-Frohwirth-Kacmar-Cunningham} Combination \label{ex:coordinativelast}

Most of the rest of the compounds we categorize based on the part of speech of the compound that is created. First are compound nouns (which can, in some circumstances be used as adjectives), which can be created in a variety of ways:

\ex. \textbf{Compound nouns: Noun + noun}
\a. He tries his \textbf{dad-voice} again
\b. Healthy Sesame \textbf{Almond-Fluff} Pasta
\c. in order to save trillions of precious \textbf{life-months} and lives .
\d. Not all carriers offer their own \textbf{bagshare} program
\e. The best times to use cooking oil on your grill / \textbf{grilltop} include
\e. This plant is very similar to \textbf{Flowerpotweed} . 
\e. using calipers and then used while the lens was still fresh to obtain samples for \textbf{x-ray-analysis} ,
\e. Some bears have more of a \textbf{densize} capacity than others .
\e. The Pongo pygmaeus vocal repertoire is highly complex and reflects the complexity of the \textbf{callsong} behaviour of this species ,
\e. Upon hitting a \textbf{ball-cage} , the player becomes unable to move and must jump 
\e. He refers to them as `` \textbf{hill-elves} " , or Eistlaes , in Quenya ;

\ex. \textbf{Compound nouns: Adjective + noun}
\a. So what happens to a \textbf{raw-fruit} diet with no dairy ?
\b. The paper concludes by proposing possible further extensions toward an actual `` \textbf{computational-self} theory "
\c. and the front and the back shall be of \textbf{bluework} with a gold frame for the robe .

\ex. \textbf{Compound nouns: Adverb + noun:} a press release on the topic from \textbf{then-CMA} president Tom McGinnis .

\ex. \textbf{Compound nouns: Verb + noun:} only the latter can actually be seen with a \textbf{sputterball}

\ex. \textbf{Compound nouns: Noun + postpositive modifier}
\a. at least 1 in 3 employed adults reported having a \textbf{job-to-be-held} .
\b. Tags : \textbf{Apprentice-in-training} , Magic , Apprentice-in-training , Magic Academy , \textbf{mentor-in-training}

\ex. \textbf{Compound nouns: Noun + number}
\a. At the beginning of the Siege of Eayn , \textbf{Eylan-3} was attacked by the Republic .
\b. A test flight at the end of 2017 will conduct \textbf{Thaad-AMM-3}

\ex. \textbf{Compound nouns: Noun + deverbal noun}
\a. SoundFont is available on apt for installation , by replacing the \textbf{font-encoder} .
\b. \textbf{Poster-maker} David Hill has now stepped forward
\c. but in recent years the practice has taken on a more popular reputation as the \textbf{fuel-extraction} method of choice for many small businesses hoping to make a profit

Next are compound adjectives, which also can be constructed in a variety of ways:

\ex. \textbf{Compound adjectives: Noun + adjective}
\a. Today , I have another record to share with my \textbf{wing-weary} students and colleagues :
\b. Last year , Rangers hitters batted.243 against a league-leading 1,200 or more innings of \textbf{relief-friendly} MLB pitching .
\c. Mexico seemed to be having a bit more trouble with its \textbf{backpass-heavy} tactics than in previous World Cups .
\d. and Dan Dennett as they talk about their work together and the world of`` \textbf{science-neutral} " philosophy .

\ex. \textbf{Compound adjectives: Adverb + adjective}
\a. Both wheelsets feature a \textbf{slightly-slender} offset .
\b. a non-distorted copy may be displayed in at least a \textbf{partially-circular} area 

\ex. \textbf{Compound adjectives: Number + Noun}
\a. On-highway travel is permitted between campus and off-highway travel only within a \textbf{two-quarter-mile} radius of any of the following :
\b. and the presence of militants in many areas along its 2,500-kilometre ( \textbf{1,460-mile} ) perimeter .
\c. out of Carver-Hawkeye Arena to the new \$ 70 million , \textbf{7,400-seat} Assembly Hall ,
\d. It boasts a \textbf{4-stereo} sound system with a pair of subwoofers mounted a few inches

\ex. \textbf{Compound adjective: Preposition + Noun}
Therefore , in the present study we performed non-invasive ( \textbf{in-participants} )

\ex. \textbf{Compound adjectives: Adjective/Adverb + gerund}
\a. something that is false or \textbf{false-nurturing}
\b. to create clear-cut attacking situations and a heavy , \textbf{sometimes-lurching} attack

\ex. \textbf{Compound adjectives: Adjective/adverb + passive}
\a. The very people who should know better , are making \textbf{scientifically-skewed} claims .
\b. Eligible for Broadcasting \textbf{Governmental-Funded} Eligibility

\ex. \textbf{Compound adjectives: Adjective + verb}
\a. The hotend does not need any special hardware and has \textbf{easy-set} screws .

\ex. \textbf{Compound adjectives: Noun + gerund/participle}
\a. Extremely powerful ultrasonic cleaner that cleans debris and \textbf{debris-containing} debris out of every inch of surface it hits
\b. teachers are given incentives to target a `` \textbf{school-chasing} " strategy .
\c. characters themselves , but of who those characters mean to a different segment of the \textbf{superhero-watching} public .
\d. One of the principal problems with \textbf{carbohydrate-eating} is how it stimulates blood sugar and fat stores ,
\e. island of Oahu , Hawaii , where they began a life of mushroom picking and \textbf{mushroom-making} , later known as the `` Mycological Arts " . 
\e. Kenow hopes to set up a \textbf{loon-viewing} deck in a few years to better identify the small , mostly nocturnal birds .

\ex. \textbf{Compound adjectives: Noun + past participle}
\a. browser features also come as Microsoft has added Internet Explorer 11 to the list of \textbf{IE10-based} browsers .
\b. and we 'll be navigating to the folder where your project and \textbf{composer-managed} libraries are in .
\c. NASA had to move closer to developing the systems necessary : a \textbf{propellant-fed} rocket , an ascent engine , an emergency escape system and a crew evacuation system
\d. \textbf{Flag-shaped} flag sold in Alameda County
\e. Soy is often used in place of dairy-containing dairy products in \textbf{dairy-reduced} diets and is being studied as a possible supplement for osteoporosis prevention .

\ex. \textbf{Compound adjectives: Verb + preposition}
\a. If you aren 't using \textbf{nail-on} nails and tape , you also can glue them to the wood . 

\ex. \textbf{Compound adjectives: Verb + postpositive modifier} \\ Torpexes are typically of a weapon \textbf{mount-only} .

\noindent
Finally, there is exactly one example of a compound verb:

\ex. \textbf{Compound verb: Preposition + verb} One side of the farm has been taken \textbf{off-grazing} over the winter

\noindent
The last major class of compounds is neoclassical compounds, which are formed by adding one or more Greek- or Latin-based affixes to a stem. Occasionally these involve stacking several affixes; e.g., (\ref{ex:neo_stack}) uses the prefixes \textit{epi-}, \textit{neo-}, and \textit{ptery-}.

\ex. \textbf{Neoclassical compounds}
\a. These reservoirs supply a special type of rock called \textbf{granophyllite} that was originally made up of the granites from which the Svalbard region was originally
\b. Hydrochlorothiazide contains hydroxychlorothiazoxide which , together with \textbf{hydroxychlorothiazide} and hydroxypyridazine hydrochloride , form a yellow crystalline powder .
\c. \textbf{Thermolithotrophic} bacteria are Gram- negative bacteria that produce a wide range of enzymes and are more
\d. the author of the books `` Ethical and \textbf{Religio-Economic} Consequences of American Wars Since 1898 : A Primer for Political Policy Makers "
\e. Pelagic \textbf{epineopterygoid} \label{ex:neo_stack}

\paragraph{Derivational affixes}

Derivational affixes are prefixes or suffixes that are added to a word to create a new word. The derivational affixes that we observe are summarized in Figures \ref{tab:affixes1of4} through \ref{tab:affixes4of4}.

The first two categories of such affixes are diminutives such as \textit{-ling}, \textit{-lite}, and \textit{mini-} (\ref{ex:diminutives}), which make the meaning smaller along some dimension, and augmentatives such as \textit{super-} and \textit{ultra-} (\ref{ex:augmentatives}) which make the meaning larger. (Note that the diminishing/augmenting effects of these affixes are not very apparent in these examples).

\ex. \textbf{Diminutives} \label{ex:diminutives}
\a. THE REAL \textbf{FISHLING} POND
\b. This social democracy will be a `` \textbf{state-lite} " , in the sense that the king will not be in a situation where
\c. the three \textbf{Mini-Compounders} eventually defeated the pair

\ex. \textbf{Augmentatives} \label{ex:augmentatives}
\a. The Professor went on to design new \textbf{super-fabricated} paper products
\b. It was revealed in \textbf{Ultra-Sized} G ! that there was originally a card named `` Berserker " in the set

Several derivational affixes in the generated text indicate location in space or time, specifically \textit{cross-}, \textit{post-}, \textit{pre-}, \textit{proto-}, \textit{sub-}, and \textit{trans-}:

\ex. \textbf{Location in time and space}
\a. A large drawbridge ( with a \textbf{cross-reinforcement} mechanism to support it )
\b. Bach 's music can be classified , along with Beethoven , as `` \textbf{post-Johann} Sebastian styles "
\c. However , if the GC CPUs are slower than they need to be due to emulation-based issues , there is often the option of going back to a \textbf{pre-Emulation} model
\d. The `` \textbf{proto-poetry} " of modern times is the `` hyperbole " spoken by Shakespeare
\e. Other companions , enemies and \textbf{sub-Bosses} 
\f. the \textbf{Trans-Dniestria} railway

Observed morphemes that negate or reverse the meaning of the stem include \textit{anti-}, \textit{de-}, \textit{dis-}, \textit{in-}, \textit{no-}, \textit{non-}, and \textit{un-}:

\ex. \textbf{Negatives and reversatives}
\a. the most prominent member of the \textbf{anti-St-Pierre} camp
\b. He calls this phenomenon the `` Great \textbf{De-concentration}
\c. Chaos is seen today not as an all-encompassing disorder and \textbf{disorderlessness} but as a complex interplay of extremes
\d. He eventually found one of the Inhuman \textbf{Inveterans} in an abandoned building .
\e. adding an optional `` \textbf{no-knockout} " version that also removes the knockout effect
\e. they immediately pointed out my \textbf{non-Arabic-sounding} pronunciation of Arabic words .
\e. 15 Proposed 8A NEWLINE NEWLINE 16 \textbf{Unproposed} 9A

In contrast to the negatives and reversatives, there is also one prefix that expresses positive sentiment (\textit{pro-}) and one that expresses repetition (\textit{re-}):

\ex. \textbf{Positive or repetitive}
\a. In this study , we investigated the anxiogenic-like and \textbf{pro-immobility} and anxiolytic-like effects
\b. They go through the process of \textbf{re-nitrification} that gives them a new supply of nitrogen

Two observed morphemes, \textit{-an} and \textit{-ite}, attach to a place name or description to create a word meaning a resident of that place:

\ex. \textbf{Residency morphemes}
\a. The Riot ' at the Cleveland Institute of Art — a time when \textbf{Clevelandians} riot for a variety of reasons
\b. In the meantime , the \textbf{Aquallans} were forced to contend with the Supermen .
\c. made them dwell in the land of Nod ( \textbf{Nodites} )
\d. Inspired by the successes of social movements such as the Paris Commune and the American labor movement in the 1890s , these small \textbf{townites} were inspired to take action to create a new economic alternative

We categorize most of the rest of the affixes by the part of speech of the word they create (and potentially also the part of speech of the stem, though this categorization is imprecise because some morphemes, such as \textit{-ism}, can take multiple different parts of speech as a stem). We observe affixes that create a noun from a verb or an adjective (\ref{ex:verb_to_noun}); affixes that create a noun from another noun (\ref{ex:noun_to_noun}); affixes that create an adjective from a different part of speech (\ref{ex:adjectivalization}); affixes that create a verb from a different part of speech such as a noun or adjective (\ref{ex:verbalization}); and an affix that generates an adverb from an adjective (\ref{ex:adverbalization}).

\ex. \textbf{Verb or adjective to noun} \label{ex:verb_to_noun}
\a. \textit{-ation}: Now , there 's also a difference between an \textbf{IKEA-ification} and a \textbf{Vogue-ification} .
\b. \textit{-er/-or}: This is what it means to be a \textbf{data-aggregator}
\c. \textit{-ity}: \textbf{Behaviourality} was evaluated during the first and first and second consecutive day in groups of 10
\d. \textit{-ment}: The most anticipated part of the redevelopment will be the `` \textbf{Prapliftment} Zone " - the area where the residences are put up .
\e. \textit{-ness}: Chaos is seen today not as an all-encompassing disorder and \textbf{disorderlessness} but as a complex interplay of extremes

\ex. \textbf{Noun to noun}\label{ex:noun_to_noun}
\a. \textit{-dom}: a new version of a zombie meme began sprouting from \textbf{quackdom} .
\b. \textit{-ism}: \textbf{YIMBYISM} is all about supporting and increasing the density ( and the number of units )
\c. \textit{-ist}: it might be worth looking at the responses from Pythonistas or the \textbf{JavaScriptists}
\d. \textit{-ry}: and has an opportunity to become one of the Scouts who upholds the values of \textbf{scoutry} at any age
\e.  \textit{-scape}: When we compiled this list of \textbf{famescapes} , it wasn 't about proving the validity of this view .
\e. \textit{-ship}: Charlie is a good horse : good at pulling a plow and excellent at \textbf{foalship} training .
\e. \textit{-verse}: The Smurfs vs. The \textbf{Smurfverse} 
\e. \textit{co-}: we are witnessing a major movement away from a capitalist workplace toward a `` \textbf{co-workplace} " , where people are working together to solve problems and create goods and services
\e. \textit{e-}: Scooters with electronic systems include the eGo-T and \textbf{e-Scooter} .

\ex. \textbf{Adjective-forming affixes}\label{ex:adjectivalization}
\a. \textit{-able}: \textbf{Gemcraftable}
\b. \textit{-ian}: is truly related to the mass of the \textbf{bosonian} Higgs-like particles
\c. \textit{-ic}: isolinear memory processor ; isolation field ; \textbf{isonetic} wave
\d. \textit{-ish}: the \textbf{Moldbug-ish} strategy to end `` civic nationalism " .
\e. \textit{-less}: The dual injection systems on this engine feature \textbf{injector-less} , single-port fuel injection systems .
\e. \textit{-like}: the console lacks any form of \textbf{Blu-ray-like} functionality
\e. \textit{-th}: the \textbf{752th} year of the Hebrew Calendar

\ex. \textbf{Verb-forming affixes}\label{ex:verbalization}
\a. \textit{-ify}: In this case , that 's `` vogue " for `` \textbf{vogue-ification} "
\b. \textit{-ize}: I hear how you worry that you 're being judgmental or \textbf{judgmentalizing}

\ex. \textbf{Adverb formation} \label{ex:adverbalization}
\a. \textit{-ly}: \textbf{Thirteenthly} : NEWLINE NEWLINE It was narrated that al-Tirmidhi and ' Atiya narrated

Finally, there are 2 morphemes that are used in the context of a specific television franchise, the Digimon franchise, to coin words relating to that franchise:

\ex. \textbf{Television-franchise-specific morphemes}
\a. \textit{digi-}:  with 50 \textbf{digifications}
\b.  \textit{-mon}: Black Ogremon can digivolve to Gabumon with \textbf{Shoujoumon}

\paragraph{Character manipulation}

Some novel words involve manipulations at the level of individual characters (summarized in Figure \ref{tab:nonaffix}). First is the use of letter repetition, either to elongate a word for emphasis (\ref{ex:elongate}) or to indicate nervousness (\ref{ex:nervous}). Second is the creation of onomatopoeias, in which each letter is meant to represent a sound in the real world (\ref{ex:onomatopoeia}).

\ex. \textbf{Letter repetition}
\a. \textbf{Youuuuuuuuu} ! ! \label{ex:elongate}
\b. `` W-what are y-you \textbf{m-meant-} " \label{ex:nervous}

\ex. \textbf{Onomatopoeia}\\ make a humming noise that sounds like `` tch-tch " or as low a `` \textbf{ka-a-la} " and a `` \textbf{hwa-hwa} " while searching for food \label{ex:onomatopoeia}

\paragraph{Portmanteau}

Finally, there are three generated words that could potentially be viewed as portmanteau words: (\ref{ex:portmanteau1}) is a blend of \textit{Disqus} and \textit{etiquette}, (\ref{ex:portmanteau2}) is a blend of \textit{pizza} and \textit{apocalypse}, and (\ref{ex:portmanteau3}) is a blend of \textit{gel} and \textit{popsicle}. However, it is possible that the model does not view these as blends but rather as compounds: it may have learned a morpheme \textit{-iquette} that means ``etiquette," a morpheme \textit{-pocalypse} that means ``apocalypse," and a morpheme \textit{-sicle} that means ``popsicle." Indeed, \citet{pinter2020blend} found that LMs perform poorly at handling portmanteau words, which would support the hypothesis that, in the cases we have observed, GPT-2 is not handling these words as portmanteau words.

\ex. \textbf{Portmanteau words}
\a. Please make sure to read the \textbf{Disqusiquette} before leaving comments . \label{ex:portmanteau1}
\b. What 's in the future for the ` \textbf{Pizza-Pocalypse} ' ?  \label{ex:portmanteau2}
\c. and took two gels ( Molly and a \textbf{gelsicle} ) .  \label{ex:portmanteau3}

\section{Additional examples for the analyses}\label{app:additional_examples}

Here we provide additional examples from the manual analyses discussed in Section \ref{sec:analysis}.

\subsection{Plurals}

To form English plurals, it is necessary to choose between \textit{-s} and \textit{-es}. Of the 74 plurals in our analyzed sample, the model made the correct choice for 72 of them, only getting wrong the two shown in (\ref{ex:incorrect_plural}) (Figure \ref{tab:pluralconfusion}). The 5 cases where it correctly predicted \textit{-es} instead of the more common \textit{-s} are in (\ref{ex:correct_es}).

\begin{figure}
    \centering
    \begin{tabular}{ccc} \toprule
        Predicted form & \textit{-s} is correct & \textit{-es} is correct \\ \midrule
        \textit{-s} & 67 & 0 \\
        \textit{-es} & 2 & 5 \\ \bottomrule
    \end{tabular}
    \caption{Confusion matrix for plurals}
    \label{tab:pluralconfusion}
\end{figure}

\ex. Incorrect plurals\label{ex:incorrect_plural}
\a. in the same way as regular \textbf{1099es} .\label{ex:es1}
\b. Why do \textbf{SQLes} have to change\label{ex:es2}

\ex. Correct usage of \textit{-es}\label{ex:correct_es}
\a. more than two million `` \textbf{metches} " 
\b. when Mr. Fowles asked him about it . The \textbf{Fowleses} ' lawyer 
\c. \textbf{Electories} by Jason Ditz
\d. the \textbf{ridiculousities} of war
\e.  \textbf{Torpexes} are small hardpoints found on smaller ships

\subsection{Possessives}

Forming possessives in English requires a choice between two possible forms, \textit{'s} and \textit{'}. All but one of the generated possessives had a correct form (Figure \ref{tab:possessiveconfusion}); the incorrect example is in (\ref{ex:watchmakers}). Some examples of the model correctly using the apostrophe-only form of the possessive are in (\ref{ex:correct_plural_possessive}).\footnote{Note that, when the possessor ended with \textit{-s} but was not plural, we allowed either form of the possessive, following variation in common usage \cite{huddleston2001cambridge}. Thus we count as correct both \textit{This model is based on Adam \textbf{Ondrus 's} classifier} and \textit{content offered on \textbf{NEXUS '} Website}.}

\ex. Incorrect possessive
\a. known as \textbf{watchmakers 's} timepieces\label{ex:watchmakers}

\ex. Correct plural possessives\label{ex:correct_plural_possessive}
\a. our \textbf{census-takers '} reports 
\b. The \textbf{Fowleses '} lawyer
\c. the \textbf{genoshans '} fear of the Andromeda Initiative
\d. The \textbf{Flexagons '} unique patterns

\begin{figure}
    \centering
    \begin{tabular}{ccc} \toprule
        Predicted form & \textit{-'s} is correct & \textit{-'} is correct \\ \midrule
        \textit{-'s} & 125 & 1 \\
        \textit{-'} & 0 & 10 \\ \bottomrule
    \end{tabular}
    \caption{Confusion matrix for possessives}
    \label{tab:possessiveconfusion}
\end{figure}

\subsection{Acronyms}

Of the 195 novel acronyms in our generated text, 75 appear with the full version of what the acronym stands for. In 21 of the 75 cases, the acronym is a suitable abbreviation for the shortened form (\ref{ex:acronyms_good}); in the remaining 54 cases, the acronym is not a suitable abbreviation. Often, the errors involve having extra letters in the acronym (\ref{ex:acronyms_extra_letters}), often repeats of letters that appear elsewhere in the acronym (\ref{ex:acronyms_extra_letters_a} through \ref{ex:acronyms_extra_letters_d}), but not always (\ref{ex:acronyms_extra_letters_e} through \ref{ex:acronyms_extra_letters_g}). Other types of errors include the omission of a letter (\ref{ex:acronyms_other_a} and \ref{ex:acronyms_other_b}), having letters out of order (\ref{ex:acronyms_other_b} and \ref{ex:acronyms_other_c}), and replacing a letter with a different, incorrect letter (\ref{ex:acronyms_other_d} and \ref{ex:acronyms_other_e}). In a few cases, the generated acronym differs from its expansion to a more substantial degree (\ref{ex:acronyms_other_f}).

\ex. \label{ex:acronyms_good} 
\a. The Money Funders International Group ( \textbf{MFIG} )
\b. the Cathedral Development Strategy Review Group ( \textbf{CDSRG} )
\c. the US-China Economic and Security Review Commission ( \textbf{US-CESRC} )
\d. the West of England Cricket and Athletics Club ( \textbf{WECAC} )

\ex. \label{ex:acronyms_extra_letters}
\a. The National Census and Statistics Bureau ( \textbf{NCBSB} ) \label{ex:acronyms_extra_letters_a}
\b. the Parliamentary Joint Committee on Human Rights ( \textbf{PJCHRC} ) \label{ex:acronyms_extra_letters_b}
\c. Koch Companies Public Sector ( \textbf{KCPSP} )\label{ex:acronyms_extra_letters_c}
\d. the American Academy of Pain Medicine ( \textbf{AAAPM} ) \label{ex:acronyms_extra_letters_d}
\e. the Tennessee River Gorge ( \textbf{TNWRG} ) \label{ex:acronyms_extra_letters_e}
\e. Health Resources and Services Administration ( \textbf{HRSDA} ) \label{ex:acronyms_extra_letters_f}
\e. the International Bank of Settlements ( \textbf{IBNSA} ) \label{ex:acronyms_extra_letters_g}

\ex. \a. Ruby Interpreters and Ruby Users Committee ( \textbf{RIRC} )\label{ex:acronyms_other_a}
\b. the Gulf Coast Disaster Recovery Task Force ( \textbf{GCDFT} )\label{ex:acronyms_other_b}
\c. The Nigerian Institute for Demographic and Social Research ( \textbf{NIDRS} )\label{ex:acronyms_other_c}
\d. the National Coalition of Latino Elected Officials ( \textbf{NCLEP} )\label{ex:acronyms_other_d}
\e. Extremely Large Interactive Neutrino Experiment ( \textbf{ELIGO} )\label{ex:acronyms_other_e}
\e. Angola Democratic League ( \textbf{ULANL} )\label{ex:acronyms_other_f}

\subsection{Examples of incorrect morphology}

Here we review some common sources of morphological errors among novel words generated by GPT-2.

\paragraph{Incorrect stem changes:} Some of the errors arise from GPT-2 making unwarranted changes to the stem. In (\ref{ex:stemchange3}), a name that was consistently spelled as \textit{Shuutou} earlier in the passage has been given an extra \textit{u} in its possessive form. In (\ref{ex:stemchange1}) and (\ref{ex:stemchange2}), GPT-2 has generated two different words that are most likely intended to be the adjectival form of the word \textit{Pentagon} (the headquarters of the US Department of Defense). Neither of these terms (\textit{Pentagorean} and \textit{Pentagran}) are plausible ways to turn \textit{Pentagon} into an adjective; the most plausible correct form would be \textit{Pentagonian}, which has in fact appeared in the training set. Thus, by using \textit{Pentagorean} and \textit{Pentagran}, GPT-2 is being inconsistent both with itself and with its training data.

\setlength{\Exlabelwidth}{1em}

\ex. Uncalled-for changes to the stem:
\a. \textbf{Shuuutou 's} flight speed \label{ex:stemchange3}
\b. Pentagon and Cybersecurity ... \textbf{Pentagorean} nuclear arms \label{ex:stemchange1}
\b. \textbf{Pentagran} war on the way \label{ex:stemchange2}

\paragraph{Inconsistent morphology:}  Beyond the Pentagon-based examples above, the suffix \textit{-an} appears to be a common source of inconsistency for GPT-2. In (\ref{ex:hamiltonan}), the generated word for people from the town Hamilton is \textit{Hamiltonan}, but later in the same generation it is formed as \textit{Hamiltonian}. (\ref{ex:genoshan}) shows inconsistency in how to refer to residents of genosha, and (\ref{ex:clevelandan}) and (\ref{ex:clevelandian}) show 2 different ways from 2 different generations to refer to someone from Cleveland. None of these (except for \textit{genoshaans}) count as ill-formed in Figure \ref{tab:syntactic_well_formedness} because there is variability in how the \textit{-(i)an} suffix can be applied, but the inconsistency is an issue. These examples display a different type of inconsistency from the inconsistency observed in prior work, namely inconsistency in how to apply morphology, as opposed to factual inconsistency \cite{welleck2019dialogue,li2020dont}.

\ex. Inconsistent demonyms
\a. as the percentage of \textbf{Hamiltonans} in the GTA increases , so does the number of people leaving the city ... almost a quarter of all \underline{Hamiltonians} \label{ex:hamiltonan}
\d. the \textbf{genoshans} were the first of a number of species...the \textbf{genoshaans} managed to regain their homeworld\label{ex:genoshan}
\b. that \textbf{Clevelandans} are forced to endure\label{ex:clevelandan}
\c. a time when \textbf{Clevelandians} riot\label{ex:clevelandian}

\paragraph{Missing sound changes:} As mentioned above, the word \textit{genoshaans} in (\ref{ex:genoshan}) is considered ill-formed. The reason is that it lacks the proper phonological change to the \textit{-an} suffix that occurs when the stem ends with \textit{-a}, namely of deleting one of the instances of \textit{-a}. Another example of failing to make a sound change is in (\ref{ex:tenaounce}) in which \textit{a} should instead be \textit{an}.

\ex. the \$ 6 , \$ 8 or \$ \textbf{10-a-ounce} china cup cake \label{ex:tenaounce}

\paragraph{Plurals in compounds:} A few ill-formed examples arise from using the plural form of a noun as the first part of a noun-noun compound; generally, the first noun in a noun-noun compound is the singular form of the noun, though there are exceptions in standard usage, so it is unclear if these should actually be viewed as errors.

\ex. Plural in compound
\a. `` common-sense " \textbf{guns-control} policies
\b. The...rivers had their \textbf{headswaters} in a larger basin
\c. \textbf{mushrooms-related} products

\paragraph{Overregularization:} Some of the errors can be classified as overregularization: applying a linguistic process in an overly broad way. In (\ref{ex:syllogist}), the word \textit{syllogist} has been created, presumably by changing the \textit{-ism} ending from \textit{syllogism} into \textit{-ist}. In many cases it is valid to swap \textit{-ism} and \textit{-ist} (e.g., \textit{tourism/tourist}, \textit{optimism/optimist}), but not in this case. In (\ref{ex:752th_app}), the suffix \textit{-th} has been applied to the number 752, even though numbers ending with 2 should instead get a different suffix, \textit{-nd}. 

\ex. Overregularization
\a. objections aimed against the \textbf{syllogist}\label{ex:syllogist}
\b. the \textbf{752th} year\label{ex:752th_app}

\subsection{Syntactic errors}

Most of the novel words that GPT-2 generates fit properly into their syntactic context, but it does make some mistakes, a few of which are below. 
In (\ref{ex:syntax_mistake_1}), \textit{bat-washer} is used as a verb when its structure suggests it should be a noun (though this could potentially be valid given English's flexibility about parts of speech). In (\ref{ex:syntax_mistake_2}), there is a noun-noun compound (\textit{cyber-missiles shortfall}) with a plural noun as the first noun; typically, such compounds start with singular nouns, though there are some exceptions. In (\ref{ex:syntax_mistake_3}), it should most likely either say \textit{look anti-Tunisian} or \textit{look like anti-Tunisians}. Finally, in (\ref{ex:syntax_mistake_4_app}), \textit{load-samples} is plural but is given a singular verb, \textit{provides}.

\ex.
\a. if I \textbf{bat-washer} it \label{ex:syntax_mistake_1}
\b. its massive \textbf{cyber-missiles} shortfall \label{ex:syntax_mistake_2}
\c. just to make our community look like \textbf{anti-Tunisian} \label{ex:syntax_mistake_3}
\d. Slicex \textbf{load-samples} provides a single button\label{ex:syntax_mistake_4_app}

\subsection{Agreement} 
Here we look at the novel plural nouns that GPT-2 generates to see whether the rest of the generated sentence observes the correct consequences of the word's plurality. First, (despite the one mistake in \ref{ex:syntax_mistake_4_app}), GPT-2 generally does well at providing plural verbs (underlined) to agree with novel plural nouns, whether the verb appears after the noun (\ref{ex:subjverb_basic_app}) or before the noun in the context of a question (\ref{ex:subjverb_inversion_app}). In (\ref{ex:subjverb_rc_app}), it correctly uses a plural verb for both verbs that agree with the novel plural subject---a verb within the relative clause, and a verb after it. The correct agreement with the verb after the relative clause is especially impressive because, in both sentences, there are 3 singular ``distractors" (italicized) between the subject and the verb.

\ex. \label{ex:subjverb_basic_app}
\a. We know that \textbf{M-Sinks} \underline{need} a target
\b. when \textbf{Clevelandians} \underline{riot} 
\c. the \textbf{Aquallans} \underline{were} forced to contend with the Supermen
\d. \textbf{Torpexes} \underline{are} small hardpoints
\d. Another indicator of the poverty that \textbf{Clevelandans} \underline{are} forced to endure :
\e. The \textbf{YR-2s} \underline{were} designated simply as YR . 2 and were designated simply as YR . 2
\e. \textbf{Hustlings} \underline{work} when those people can work the job market for the required number of hours
\e. It was revealed that the \textbf{genoshans} \underline{were} the first of a number of species that the Andromeda Initiative had already studied
\e. For some reason my old 1 / \textbf{1-01s} \underline{do} not turn and drive with the shift lever down like my new ones do

\ex. \label{ex:subjverb_inversion_app}
Why \underline{do} \textbf{SQLes} have to change

\ex. \label{ex:subjverb_rc_app}
\a. The \textbf{Huamangas} , who \underline{are} descendants of indigenous people who lived on the \textit{Isthmus} of \textit{Tehuantepec} before it was covered by \textit{farmland} , \underline{have} been demanding that the federal government address the issue of climate change .
\b. \textbf{FOIA-requesters} who \underline{think} an \textit{agency} has a good \textit{reason} for withholding \textit{information} \underline{are} not always given a second opportunity to press their case .

\subsection{Other plural-relevant syntax} 

Beyond agreement, syntactic consequences of plurality are observed in a few other places as well: in using the plural possessive form that is just an apostrophe instead of the singular form of \textit{-'s} (\ref{ex:plural_possessive_app}); in having the pronouns that are coreferential with the noun be plural as well (\ref{ex:plural_pronoun_app}); and in following determiners that require a plural noun (\ref{ex:plural_determiner_app}).

\ex. \label{ex:plural_possessive_app}
when Mr. Fowles asked him about it . The \textbf{Fowleses} \underline{'} lawyer , James F. Kelly ,

\ex. \label{ex:plural_pronoun_app}
\a. I love \textbf{Klymits} , but it has been nearly impossible for us to find \underline{them} in stores .
\b. The \textbf{Sarrats} were lucky to have her as part of \underline{their} lives
\c. The \textbf{color-coats} are far more black \& white than \underline{their} predecessors 

\ex. \label{ex:plural_determiner_app}
\a. as the Paris Commune and the American labor movement in the 1890s , \underline{these} small \textbf{townites} were inspired to take action to create a new economic alternative
\b. to help you understand why there are so \underline{many} \textbf{Brazilianisms} in the English language as opposed to the Portuguese one

\subsection{Incrementing/ordering} 

Here we provide the examples mentioned in the main text where GPT-2 successfully increments. In (\ref{ex:increment_numbers}), it increments numbers from \textit{Firstly} to \textit{Fourteenthly}, with the last two (\textit{Thirteenthly} and \textit{Fourteenthly}) being novel. In (\ref{ex:increment_variables}), it increments the letters at the ends of variable names in computer code, going from \textit{multiplyx} to \textit{multiplyy} to \textit{multiplyz}. Finally, in (\ref{ex:increment_alphabet}), the prompt ends with an alphabetical list of companies, and GPT-2 continues this list, largely (though not entirely) staying in alphabetical order, including many novel words along the way (all in bold).

\ex. 
\a. Firstly : NEWLINE NEWLINE It is not permissible...Secondly : NEWLINE NEWLINE It was narrated that...Twelfthly : NEWLINE NEWLINE It was narrated from...\textbf{Thirteenthly} : NEWLINE NEWLINE It was narrated that al-Tirmidhi and ' Atiya narrated that... \textbf{Fourteenthly} : NEWLINE NEWLINE It was narrated that al-Shawkaari narrated that... \label{ex:increment_numbers}
\b. \textbf{multiplyx} = math. ceil ( self. multiplier [ 0 ] * self. multiplier * 2.0 ) self. honey \_ hive [ 0 ] . \textbf{multiplyy} = math. ceil ( self. multiplier [ 1 ] * self. multiplier * 2.0 ) self. honey \_ hive [ 1 ] . \textbf{multiplyz} = math. ceil ( self. \label{ex:increment_variables}
\c.
BWS Buffalo C-Tech Can-Am Carre Revero Chameleon \textbf{Chintsoft} Claris \textbf{ClouDio} \textbf{CO-IoC} Cisco Connex Computer Comply Coopers \textbf{ConsalCO} Computer \textbf{CoreGear} Crestron Corel Dell D \& H Digital Storm Dell Eizo Epson Epoch Exabyte Exponent Falcon Formica FreeNet G-Technology Gigaset Gionis Gigabyte \textbf{GigaForce} \textbf{Glance-on} \textbf{Glass-einhard} Hauppauge Hauppauge \textbf{HPC-Link} HP i-mate IBM IBM PPro \textbf{ICON-S} ICOM \textbf{ICOM-ITIC} ICOM World \textbf{iBuyInte.Com} \textbf{iBuyMall} \textbf{Icom-U2} \textbf{iGK} Computers \textbf{IISkills} Inet IPC \textbf{iPlayIiMac} Iguana iRobot IceCool iPort \textbf{iSoftimage-iMac} Jackson JBL JEDI Kingston KeySmart Kinesis \textbf{Konica-Tek} Konika Logitec LSI Lenovo Maxtor Marvell Matrox Maxix \textbf{Maxixx} Microvision Microsoft \textbf{MoboModem} MSN MobileNet MyNet MXM \textbf{MyTouchNokia} \textbf{Nandmark} Neopan Tandem NoSQL Netside Netgear \textbf{NetVu} NetXtreme Network Technology Novell NTG Nyko \textbf{Noxel} Octave ODU Okidata Omnipath Online Data Osram Panasonic Patek Pivotal Prodigy \textbf{PQ-Link} Qorvo Quanta Quad-Core \textbf{Quadmark} RCA Rapoo \textbf{Rega-Link} \textbf{Regal-Link} Regelation \textbf{Rapidra} Regulus Redstone Redbox Reuleaux \textbf{RTI-Siemens} Semtec \textbf{Simlogic} Sintek Silicon Integrated Systems \textbf{Simvit} Simulink Smartcard \textbf{Sonic-com} South River Soundstar SPARC SuperComputer Systems SRT Systema TATA The Best Technology TLC Toshiba \textbf{Tungst-Sang-Tzu} TriQuint Travelstar \textbf{TSSTech} US Robotics \textbf{ViaL} Vision Vision VisionX WebPro Technologies WEIT Vantec VirnetX \textbf{VideoFusion} Vipnet \textbf{VSCare} VSP VTech Vortex Works Zebra Zeta ZXZ ZTE \label{ex:increment_alphabet}

\subsection{Quotation marks} 

In GPT-2's text, as in human-written text, novel words are more likely to be enclosed in quotation marks than non-novel words.
Some examples of novel words being in quotations are below:

\ex. 
\a. the wave function in the `` \textbf{meganiverse} "
\b. a `` \textbf{co-workplace} ", where people are working together
\c. `` \textbf{Active-Passive-Inactive} " investing
\d. the `` \textbf{anarchism-capitalism} " debate
\e. The `` \textbf{proto-poetry} " of modern times
\e. the `` \textbf{un-competition} " that is happening as a result of rapid technological advances

\subsection{Novel words with meanings that are suitable for their context}

Below are some examples of novel words that are used in ways that are particularly well-suited for their semantic contexts. 

\ex.
\a. And it was very easy for him to be a \textbf{painter-scientist} , because he used nature and art .
\b. The process is pretty simple : a cyclist is followed ( suspect or \textbf{not-so-suspect} ) from a safe distance
\c. They go through the process of \textbf{re-nitrification} that gives them a new supply of nitrogen
\d. These include the concept of ` \textbf{co-causation} ' , in which effects are thought to be caused by causes that act in parallel
\e. The other thing of course is that the companies will be allowed to sell and to sell this information to the government in real time . This is what it means to be a \textbf{data-aggregator} and it is an interesting way to think about all of this .
\e. Thirdly , a new unique feature , the `` \textbf{bondbreaking} enchantment " , which renders any item cursed by the `` Cursed item " bug inadmissible to the user and permanently breaks any binding .
\e. we are witnessing a major movement away from a capitalist workplace toward a `` \textbf{co-workplace} " , where people are working together to solve problems and create goods and services on a much smaller scale .

\subsection{Novel words with meanings that are not suitable for their context}

Below are some examples of novel words where there is clear evidence that the word is not used in a semantically-sensible way. In (\ref{ex:judgmentalizing}), \textit{judgmentalizing} is used in a way that suggests it should mean ``being judgmental," but the word's structure should yield the meaning of ``making someone judgmental." In (\ref{ex:brazilianism}), \textit{Brazilianism} is used to refer to an English term, not a Brazilian term. In (\ref{ex:bittrex}), Bittrex is referred to as Bittrex-like, but it is not standard to refer to something as being ``like" itself. (\ref{ex:disorderlessness}) is contradictory because \textit{disorderlessness} should be the opposite of disorder. (\ref{ex:anticatatonia}) is also contradictory because anti-catatonia effects are said to decrease motor activity, even though a decrease would be consistent with catatonia, not anti-catatonia.
(\ref{ex:frontfloor}) refers to the front floor, even though the floors of buildings are arranged vertically, so a building cannot have a front floor. In (\ref{ex:nitratedeficient}), nitrification is referred to as a nitrate-deficient state, even though it most likely should be a nitrate-rich state. (\ref{ex:markdown}) refers to a markdown-to-HTML converter having markdown output even though its output would actually be HTML. (\ref{ex:ie10}) says that Internet Explorer 11 is built on Internet Explorer 10, even though most likely it would be viewed as a new browser, not a version of Internet Explorer 10. Finally, (\ref{ex:noknockout}) says that a no-knockout effect would enable people to be knocked out, which is contradictory.

\ex. 
\a. I hear how you worry that you 're being judgmental or \textbf{judgmentalizing} if you talk about how you 're leaving\label{ex:judgmentalizing}
\b. An old school English term is a \textbf{Brazilianism} .\label{ex:brazilianism}
\c. Blockstream 's shares were traded on Bittrex , a \textbf{Bittrex-like} cryptocurrency exchange .\label{ex:bittrex}
\d. Chaos is seen today not as an all-encompassing disorder and \textbf{disorderlessness} but as a complex interplay of extremes ,\label{ex:disorderlessness}
\e. Moreover , the main aim of this study was to investigate whether an anxiolytic effect of Vitex by increasing OAT \% and OAE \% is accompanied by anti-immobility and \textbf{anti-catatonia} effects by decreasing motor activity\label{ex:anticatatonia}
\e. wandering around the \textbf{front-floor} lobby of the Hotel Del Coronado \label{ex:frontfloor}
\e. This \textbf{nitrate-deficient} state is called nitrification .\label{ex:nitratedeficient}
\e. so you can use its \textbf{markdown-to-HTML} convertor with the markdown output format you prefer\label{ex:markdown}
\e. Microsoft has added Internet Explorer 11 to the list of \textbf{IE10-based} browsers .\label{ex:ie10}
\e. The only thing I 've done with my mod since then ( well , maybe a little bit before ) is adding an optional `` \textbf{no-knockout} " version that also removes the knockout effect , so you can actually be knocked out again if you take enough damage .\label{ex:noknockout}

\subsection{Numbers} 

The analyzed sample of text includes 75 instances of a number plus a unit, such as the following: 

\ex. \a. Minimum Water Pressure : \textbf{2.5atm}
\b. Tags : ch.5 , ch.5.1 , \textbf{ch.5.2} , ch.6 , \textbf{ch.6.1} , ch.6.2
\c. Available OS Memory : \textbf{8147MB} RAM

Several of these involve math, which gives us an opportunity to see whether GPT-2 understands the numbers it is using. (For convenience, we also include some that are classified as number-noun compounds, rather than numbers (\ref{ex:number_n_conversions_app})). Generally it appears that GPT-2 does not understand the numbers it generates; all the examples involving math are included below.  (\ref{ex:conversion_difference_app}) includes the computation of a difference between two numbers, where it is said that $1065mhz - 1030mhz = 90.5MHz$. Regardless of whether the MHz on the right hand side is interpreted as the same unit as mhz on the left hand side, or if the capitalization is viewed as meaningful, this computation is incorrect.
(\ref{ex:conversion_subtraction_app}) involves a physical impossibility: a 4-milliliter container cannot hold 10.4 milliliters of juice. Meanwhile, (\ref{ex:40yd_app}) and (\ref{ex:3cone_app}) give quantities that are not strictly impossible but are highly unlikely: according to ESPN,\footnote{\url{https://www.espn.com/nfl/story/\_/id/28774721/nfl-combine-records-best-worst-performances}} the fastest 40-yard dash in history was 4.22 seconds, making the 2.64-second time in (\ref{ex:40yd_app}) implausibly fast; and the fastest three-cone drill in history was 6.28 seconds, making the 4.15-second time in (\ref{ex:3cone_app}) also implausible.

The rest of the examples involve conversions between units, and generally the conversions are not equivalent. (\ref{ex:lbkg_app}) says that 1240 pounds equals 735 kilograms, when in fact it equals 562 kilograms. (\ref{ex:kmmile_app}) says that 2500 kilometers equals 1460 miles when in fact it equals 1553 miles (though this example is close enough to perhaps be reasonable). (\ref{ex:mlgal_app}) says that 975 milliliters equals 2.2 gallons when in fact it equals 0.26 gallons. (\ref{ex:pounds_dollars_app}) says that 1 billion US dollars equals 610.9 million British pounds; this one is reasonable, because at current exchange rates it equals 718 million British pounds, so this is possible given fluctuations in exchange rates. Finally, examples (\ref{ex:kes1_app}) through (\ref{ex:kes4_app}) involve conversions between Kenyan shillings (KES) and British pounds (£). Across these examples (which all come from the same piece of generated text), we observe four different exchange rates: £1 = KES14.3 (\ref{ex:kes1_app}); £1 = KES25 (\ref{ex:kes2_app}); £1 = KES66.7 (\ref{ex:kes3_app}); and £1 = KES200 (\ref{ex:kes4_app}). Given this inconsistency, it appears that the model does not have any consistent meaning stored for these numbers.

\ex. \label{ex:number_conversion_app} \a. The highest speed in the XFX version is \textbf{1065mhz} , which is around \textbf{90.5MHz} higher than the 1030mhz in our testing .\label{ex:conversion_difference_app}
\b. the original 4ml tank holds \textbf{10.4ml} of e juice .\label{ex:conversion_subtraction_app}
\c. Water Tank Capacity : \textbf{975mL} ( 2.2 Gallons ) \label{ex:mlgal_app}
\d. found Mr Mitchell guilty of a combined \$ 1bil ( £ \textbf{610.9m} ) in damages and costs .\label{ex:pounds_dollars_app}
\d. In a town like Kajiado that can cost up to \textbf{KES50} ( £ 3.50 ) to find a taxi .\label{ex:kes1_app}
\d. Prices range up to \textbf{KES100} ( £ 4.00 ) a night\label{ex:kes2_app}
\d. a bed in one of these rooms can cost \textbf{KES300} ( £ 4.50 ) .\label{ex:kes3_app}
\d. be prepared for the ride to cost you \textbf{KES200} ( £ 2.50 ) .\label{ex:kes4_app}

\ex. \label{ex:number_n_conversions_app} \a. He posted a \textbf{2.64-second} 40-yard dash this spring \label{ex:40yd_app}
\b. Mandarich ran a 4.29-second 40-yard dash and \textbf{4.15-second} three-cone at the NFL combine in March .\label{ex:3cone_app}
\c.  by a \textbf{1,240-lb} . ( \textbf{735-kg} ) device \label{ex:lbkg_app}
\d. along its 2,500-kilometre ( \textbf{1,460-mile} ) perimeter . \label{ex:kmmile_app}

\subsection{Phone numbers} 

In North American phone numbers, the first 3 digits of the phone number indicate the area where the phone number is from. In many cases, the context of a generated phone number makes it clear where the phone number is meant to be from. Does GPT-2 generate phone numbers appropriate for the places? Overall, we find 77 cases where the context makes a North American location clear and where the phone number is specific to a North American location. In 54 of these 77 cases (70\%), the phone number and the contextually-specified place match. For instance, (\ref{ex:phone_seattle_good})\footnote{For all of the examples in this section, we redact information that could potentially provide contact information or addresses for real people.} uses the Seattle area code 612 and mentions Seattle; (\ref{ex:phone_nyc_good}) uses the New York City area code 212 for a location in New York City; (\ref{ex:phone_baltimore_good}) uses the Baltimore area code 410 and mentions both Baltimore and the Baltimore-located Johns Hopkins Medical Institute (and its email abbreviation of \textit{jhmi}); and (\ref{ex:phone_pittsburgh_good}) uses the Pittsburgh area code 412 for a location in Pittsburgh. It appears, then, that GPT-2 has learned valid associations between area codes and locations. 

In other cases, the phone numbers are somewhat geographically off: in both cases in (\ref{ex:phone_philly_pittsburgh}), the context refers to a location in Philadelphia (either the city of Philadelphia, or the University of Pennsylvania, which is in Philadelphia), yet the phone number uses the Pittsburgh area code 412 (which GPT-2 had properly associated with Pittsburgh in (\ref{ex:phone_pittsburgh_good})). Philadelphia is in the same state as Pittsburgh, so this is not too far off; but the cities are nonetheless about 300 miles apart, so they are not geographically interchangeable. Finally, some cases are far off: (\ref{ex:phone_wrong_state_colorado}) refers to Colorado but uses the New Hampshire area code 603, and (\ref{ex:phone_wrong_state_lubbock}) refers to Lubbock (a city in Texas) yet uses the area code 914, which is for Westchester County, New York. There are plausible reasons why a phone number might not match a location (e.g., because a person has moved), so it is not necessarily a big problem that only 70\% of the cases matched; however, in the baseline text, 55 of the 56 cases matched (98\%), suggesting that there should be a higher rate of matching than is observed.

\ex. \a. \censor{XXXXXX} : \textbf{612-6\censor{YY}-4\censor{XYX}} or \censor{XXXX} @ \underline{seattle}times.com ; \label{ex:phone_seattle_good}
\b. Call ( 212 ) \textbf{3\censor{YY}-2\censor{XYX}} to receive a customized quote for your own individual investment needs ; or contact us using our online or telephone services , or by mail at the address below : NEWLINE NEWLINE Money Funders International NEWLINE NEWLINE \censor{XXXXXXXXXXX} NEWLINE NEWLINE \underline{New York , NY} 10019 \label{ex:phone_nyc_good}
\c. In fiscal year 2017 , \underline{Johns Hopkins Medicine} provided health care to approximately 6.2 million residents of the \underline{Baltimore} region... MEDIA CONTACT : \censor{XXXXXXX} , \textbf{410-9\censor{YY}-6\censor{XYX}} ; \censor{XXXXXXX} @ \underline{jhmi}.edu \label{ex:phone_baltimore_good}
\d. \censor{XXXXXXXXXXXXXXXXXXXX} Drive NEWLINE NEWLINE \underline{Pittsburgh} , PA 15219-8011 NEWLINE NEWLINE ( 412 ) \textbf{4\censor{YY}-8\censor{XYX}} \label{ex:phone_pittsburgh_good}

\ex. \label{ex:phone_philly_pittsburgh} \a. Editor , \underline{Philadelphia} City Paper NEWLINE NEWLINE e ‐ \&\#91; at \&\#93; phillypagewriter ( \&\#91; at \&\#93; phillycitypaper.com ) ) NEWLINE NEWLINE \textbf{412.3\censor{YY}.5\censor{XYX}}
\b. Editor : \censor{XXXXXXXXXX} , MD , MPH , Chief Medical Officer , University of Pennsylvania Health System . Corresponding contributor : \censor{XXXXXXXXXX} , \censor{XXXXXX} @ phshp.org , \textbf{412-3\censor{YY}-4\censor{XYX}} .

\ex. \a. \underline{Colorado} State Patrol : \textbf{603-
2\censor{YY}-3\censor{XYX}} , \censor{XXXXX} @ denverpost.com \label{ex:phone_wrong_state_colorado} 
\b. Lubbock Police Department NEWLINE NEWLINE \textbf{914-5\censor{YY}-3\censor{XYX}} \label{ex:phone_wrong_state_lubbock}

\subsection{Generalization by composition}

There are a few cases where GPT-2 generates a novel word whose stem never appears in training but does appear in the context (the prompt plus the previously-generated words). Specifically, there are 5 cases of GPT-2 pluralizing a novel word from its context (\ref{ex:gen_from_context_plural_app}), and 3 cases of it adding other affixes to a novel word from its context (\ref{ex:gen_from_context_other_app}). We believe that these examples are best explained by composition: analogy requires some notion of similarity between the two word parts being swapped for each other, and it is unlikely that the model would have such similarity notions for a word stem it has never seen before. Thus, we think these examples are better understood as the model adding a prefix or suffix to a word from its context, without direct reference to another word that has that prefix or suffix---a form of composition.

\ex. \label{ex:gen_from_context_plural_app}
\a. ... a major movement away from a capitalist workplace toward a `` \underline{co-workplace} " ... These types of \textbf{co-workplaces} , if truly integrated and facilitated , could potentially improve and strengthen the worker-worker relationship
\b. both the Overdone and \underline{Overloved} series ... Book No. 4 : The \textbf{Overloveds} are available now on both Amazon and Barnes and Noble , respectively .
\c. during a \underline{fuedo} ... Most tornros however use their bare hands in \textbf{fuedos} . This type of fighting allows the tearo to use his / her own strength
\d. sort of support force . For example , using the \underline{LHAW} to take out other \textbf{LHAWs} , or the machineguns of the new infantry YoRHa suits.YoRHa is currently one of the
\e. An old school English term is a \underline{Brazilianism} ... of these differences and discuss them to help you understand why there are so many \textbf{Brazilianisms} in the English language as opposed to the Portuguese one .

\ex. \label{ex:gen_from_context_other_app}
\a.  They can also be local \underline{pain-dissipating} systems ( V4 ) ( sometimes the V1 and V3 fibers also form V4 ) or \textbf{non-pain-dissipating} central pain systems ( V5 ; the V1 and V3 and V4 fibers can also
\b. m. nesino NEWLINE NEWLINE Pelagic \underline{epineopterygoid} , S. kuramotoi ( L. ) dehayesii NEWLINE NEWLINE \textbf{Sub-epineopterygoid} , N. scapulatus ( M.G. ) alvarezii NEWLINE NEWLINE NEWLINE Subgenus Heteracarina contains three species
\c. ... seen with a \underline{sputterball} ... It is also very hard to escape a Vulcan warrior 's \textbf{sputterballing} . 

\subsection{Generalization by analogy}
On the other hand, there are some cases that we believe are best explained by analogy. A first example is in (\ref{ex:quackdom_app}):

\ex. the same time as a new version of a zombie meme began sprouting from \textbf{quackdom} . \label{ex:quackdom_app}

It is possible that \textit{quackdom} was formed by adding the suffix \textit{-dom} to the stem \textit{quack}. However, \textit{-dom} is a rare and idiosyncratic suffix, which makes it less likely (though, we stress, not impossible) that the model has learned a generic rule for combining it with words. On the other hand, there is a plausible path for the model to have generated it by analogy: the training set contains the word \textit{hackdom}, and it also contains a large number of words (61 of them, to be exact) which contain \textit{hack} and where changing \textit{hack} to \textit{quack} creates a different word that appears in training. For instance, the training set contains all of the following pairs of words: \textit{hackademia} and \textit{quackademia}; \textit{hackfest} and \textit{quackfest}; \textit{hacktivist} and \textit{quacktivist}; \textit{hacktacular} and \textit{quacktacular}; and \textit{hackery} and \textit{quackery}. From the distributional similarity between \textit{hack} and \textit{quack}, it is plausible that GPT-2 has learned similar representations for these two words, which would then allow it to generalize from \textit{hackdom} to \textit{quackdom}. It is still of course possible that \textit{quackdom} was created by a compositional rule that appends \textit{-dom} to a stem, but we believe that analogy from \textit{hackdom} is a more plausible account: there is a clear path for that to happen, whereas the rarity and idiosyncrasy of the \textit{-dom} suffix make it harder for a model to learn a general rule involving it.\footnote{From looking at the words, it is tempting to think that the spelling overlap between \textit{hack} and \textit{quack} would also influence GPT-2 to treat them similarly. However, this spelling cannot play a direct role: \textit{quack} is split into two tokens (\textit{qu} + \textit{ack}), but \textit{hack} is a single token of its own, so \textit{hack} and \textit{quack} have no subword tokens in common. Though the spelling similarity cannot play a direct role, it still likely plays an indirect role: the fact that these two words sound similar means that they are both conducive to the same fanciful coinages like \textit{hackademia} and \textit{quackademia} that would not work as well with a word that does not end with \textit{-ack}, and having so many similar coinages would certainly help push a model toward treating them similarly.}

Another piece of generated text which we believe provides even clearer evidence for analogy is given in full in the next subsection. 
The prompt for this generation contains the real English word \textit{torero} (borrowed from Spanish), which means ``bullfighter." The generation then contains several alternate forms of this word (some of them with plural inflection added): \textit{tearro}, \textit{tornro}, \textit{tearingros}, and \textit{tearsros}. It appears, then, that GPT-2 has taken the word \textit{torero} and then replaced the first 4 letters (\textit{tore}) with other inflections of the verb \textit{tear}: \textit{tear}, \textit{torn}, \textit{tearing}, and \textit{tears} (covering the full inflectional paradigm for this verb). Because there is no morphological process in English that can add \textit{-ro} to verbs in this way, it is unlikely that these words were generated via composition; instead, we find it much more likely that they were generated via analogy.

One final argument for analogy comes from the stem changes shown in the main paper in the examples in (\ref{ex:spellign_change}): based on subword tokenization of the words generated in these examples, we believe it is likely that GPT-2 is swapping out one token for another. For example, \textit{cookying} is tokenized as \textit{cook} + \textit{ying}. It would not make much sense to view this word's generation as starting with \textit{cook} and then adding \textit{-ying} to turn it to the progressive form of the verb, because the way to make \textit{cook} progressive is to add \textit{-ing} (creating \textit{cooking}). Instead, \textit{cookying} seems more likely to be explained by analogy with \textit{cookies} (tokenized as \textit{cook} + \textit{-ies}) and \textit{cookied} (tokenized as \textit{cook} + \textit{-ied}), where GPT-2 has learned that \textit{-ies}, \textit{-ied}, and \textit{-ying} are related suffixes (as they are in many other words such as \textit{copying} and \textit{copies}).

\subsection{Full generation about toreros}\label{app:torero}

\textit{This subsection (after the end of these italicized sentences) gives the full generation about toreros. The prompt from which this generation was generated includes both the word \textbf{torero} (meaning a bullfighter) and its plural, \textbf{toreros}.}

\bigskip
\noindent
over their heads ( this is known as a `` fuego " ) . These are made up of metal bars that can stop the bull from getting back into the ring . NEWLINE NEWLINE 2 . The toreros begin to fight the bull with a knife or stick NEWLINE NEWLINE 3 . When the \textbf{tearro} loses , he / she is killed by the bull . NEWLINE NEWLINE Do \textbf{tornros} wear clothes ? Are they hurt in any way during a fight ? Are \textbf{tornro} costumes the same colors as those of the bull ? What are the odds of winning ? NEWLINE NEWLINE A \textbf{tearros} wears a costume in which they wear a costume of some kind over their head . In Spanish bullfighting , the costume consists of a blue cloth with red trim , the sides of which are covered in leather . A tearro is then trained to become highly skilled in this style of bullfighting . The training takes 5-8 years , depending on the type of training . During this time , a tearro must learn to avoid the bull 's powerful horns . In Spain and other Spanish-speaking countries , \textbf{tearingros} are taught to avoid their horns . However , in recent years bullfighters in this style have been performing stunts to get close to the bull , wearing tornro costumes and injuring themselves . This is known as ` racing ' . The bull , for its part , can be more aggressive and more unpredictable than usual , and it could be easier to get a tearro in a dangerous situation . NEWLINE NEWLINE How do tornro 's defend themselves ? The easiest way for a tearro to protect themselves from injuries is by using a knife or stick to protect themselves . Tornros will punch or kick at their adversaries . Most often , they will also use two heavy wooden sticks to push each other back . It also pays to recognize the type of weapon a tearro uses during a \textbf{fuedo} . Some tearros will use only their hands ; these are known as `` \textbf{handarros} , " and have long , sturdy fingers that can be used for many different tricks . Most tornros however use their bare hands in \textbf{fuedos} . This type of fighting allows the \textbf{tearo} to use his / her own strength . The use of bare hands can make a tearro extremely unpredictable , so it is important to watch carefully when a tearo begins to fight . NEWLINE NEWLINE Are tornros allowed to touch the bull during a fuedo ? NEWLINE NEWLINE When fueding , the bull always wins . Once the bull enters a fuedo , a tearo must let go . This is called el aire en nuevo ( the `` fall in the night " ) . The bull then is ready for the next fuedo battle . NEWLINE NEWLINE So are all \textbf{tearsros} the same ? Tornros are one of the most widely known and documented breeds in the world . Tornros are beautiful in their own way ; these beautiful animals have a different set of needs than those of other breeds of dogs . They can only do so much to make up for the damage they do to people and other animals due to their aggressive nature . NEWLINE NEWLINE Tornros are also considered to be a more difficult breed to own than most . Not only are there many health problems that can arise with your pups , the cost of pups can be fairly expensive , even if you choose to feed and care for them the way the breed is supposed to be fed and cared for . NEWLINE NEWLINE This breed is also generally hard to train , as well as to deal with the other animals in your household , such as your pet rabbit , hamsters , guinea pigs , hamsters , guinea pigs , and rabbits . They can also be a bit on the rough side , especially around large cats . NEWLINE NEWLINE Tornros do make great pets for those who are willing to do their part in keeping the animal healthy and happy . Many owners with other pet breeds can learn how to cope with these issues as well . You don 't need a lot of love , training , and experience , but I do believe that a little bit can go a long way . NEWLINE NEWLINE There are many places that have a Tornros pup , especially in the US , where you can keep a pup . I know of two dog kennels in Texas that do take Tornros . There are also many pet shops in Texas that will take these type dogs . Also , some people love the look of a beautiful Tornros and want to adopt one as their pets . These types of people often contact those businesses to help them look for a Tornros . NEWLINE NEWLINE If you haven 't been able before to have one of these wonderful dogs with you , I 'm sure you will be able to find a home for one in your home . You probably already know many things that you can do to make sure you have a good fit for a Tornros .

\section{Overregularization and nonce CVC forms}\label{app:overregularization_cvc}

An influential observation in cognitive science is that, when a word is irregular, children sometimes produce a form predicted by the relevant regular rule instead of saying the correct irregular exception (e.g., saying \textit{goed} instead of \textit{went}). It is typically assumed that overregularized forms such as \textit{goed} would not appear in the child's input, such that the usage of such forms indicates novelty on the child's part. Would such forms also count as examples of novelty for GPT-2? To answer this question,  we used Wiktionary's list of English verbs with irregular past-tense forms,\footnote{\url{https://en.wiktionary.org/wiki/Appendix:English\_irregular\_verbs}} excluding multi-morphemic words (such as \textit{unbind} or \textit{forbear}), auxiliaries (such as \textit{can} and \textit{will}), verbs which are only irregular in their past participle but not their past tense form (e.g., \textit{shave} and \textit{sew}), archaic verbs (e.g., \textit{clepe} and \textit{nim}), and words for which the regular-rule-produced past tense is a common English word, even if unrelated.\footnote{In some cases, this is an alternate past tense of the verb in question: although some use \textit{besought} and \textit{dove} as the past tense forms of \textit{beseech} and \textit{dive}, others use the regular forms \textit{beseeched} and \textit{dived}. In other cases this is the past tense of another verb: though one meaning of \textit{ring} has \textit{rang} as its past tense, another has \textit{ringed}; and though the past tense of \textit{sing} is \textit{sang}, the regular form \textit{singed} is the past tense of a different verb, \textit{singe}. Finally, in some cases, the regularly-predicted past tense is simply a different word such as \textit{seed}, as in the seed of a plant, and \textit{leaded}, as in leaded gasoline.} This left a list of 92 English verbs with irregular past tenses, and for all 92 of them, the form predicted by the regular rule appears in the WebText training set. Thus, if GPT-2 were to use any of these regular forms, such as \textit{beginned} or \textit{thinked}, it would not be strong evidence for overregularization, as it could instead be the case that it is simply copying something it has seen before. We do not see any such forms in the sample we analyze in Section \ref{sec:analysis} (generated with top-40 sampling), but we do see some in text generated with other decoding methods, a few of which are given below:

\ex. \textbf{Overregularized past-tense forms generated by GPT-2 (decoding method in parentheses)}
\a. We \textit{builded} a bath (temperature=0.9)
\b. The Arsenal manager has admitted that his team should have been more positive in order to exploit Arsenal 's `` \textit{drawed} out " game . (top-0.9)
\c. and their disciples after did already \textit{finded} them out (top-1.0)
\d. Nick led us in rebounding and he \textit{layed} down a lot of offensive rebounds . (top-1.0)
\e. The most egregious example of TGI-18 failure is documented in a recent class action lawsuit filed by a group of soldiers , who say that the failure to distinguish between mental and physical fitness , as \textit{seeked} in the National Physical Fitness Test ( NPFT ) ... (temperature=0.9)
\e. The shadow \textit{shaked} in pain as black energy were sucked out hit it . (top-800)
\e. Two rookies — Antonio Richardson of the Chargers and George Wilson of the Browns — \textit{sitted} out a game this year because of suspensions . (top-0.9)

Other cognitive literature relies on a different type of word being assumed to be novel: words that are phonologically well-formed but that do not happen to be real English words. For instance, in the classic wug test, \citet{berko1958child} tested whether children could form the plural of the made-up word \textit{wug}. The motivation for using made-up words was that children would not have seen these words before, so, if they correctly formed the plural \textit{wugs}, it would show that they knew English's plural formation rule and had not simply memorized plural forms they had seen. Can we similarly assume that such made-up words are novel for GPT-2? 

To answer this question, we considered all possible words of the form consonant-vowel-consonant where the vowel could be \textit{a}, \textit{e}, \textit{i}, \textit{o}, or \textit{u}, and each consonant could be any of the remaining letters except \textit{y}; the two consonants could be the same as each other or different from each other. For each of these words, we also generated what its plural form would be, assuming that the word was a singular noun. For wourds ending with \textit{j}, \textit{s}, \textit{x}, or \textit{z}, we formed the plural by adding \textit{-es};\footnote{In English, some words that end with \textit{s} or \textit{z} can optionally have this letter doubled in the plural; e.g., the plural of \textit{bus} can be \textit{buses} or \textit{busses}, and the plural of \textit{fez} can be \textit{fezes} or \textit{fezzes}. We did not consider such consonant doubling in this analysis.} otherwise, we added \textit{-s}.

From these singular/plural pairs, we excluded all words for which the singular and/or plural form is a real English word, as determined by whether it is present in the NLTK word list.\footnote{\url{https://www.nltk.org/}} We also excluded all pairs for which the plural ends with \textit{-es} and where removing only the final \textit{s} creates a word in the word list. After these exclusions, 1339 nonsense words remained.

Of these 1339 words, there were 765 (57\%) for which both the singular and the plural form occurred in the training set. For the remaining 574 words (43\%), only the singular form occurred in the training set. There were no words for which the singular form did not occur in training (because it turns out that there is a part of the training set that lists all possible sequences of 3 letters).

These nonsense word results are not as extreme as the overregularization results, as there are plenty of nonsense words that would truly be novel for the model. Nonetheless, many of them are not novel, so it is not safe to assume that a word will be novel for GPT-2 solely because it is not listed in an English dictionary: it is necessary to check the training set to confirm that it is novel.

\end{document}